\documentclass[twocolumn,11pt]{article}

\usepackage{latexml}

\usepackage[letterpaper,margin=1in]{geometry}
\usepackage[utf8]{inputenc}
\usepackage[T1]{fontenc}
\usepackage{hyperref}
\usepackage{url}
\usepackage{booktabs}
\usepackage{amsfonts}
\usepackage{microtype}
\usepackage{graphicx}
\usepackage[numbers]{natbib}

\usepackage{lmodern}
\usepackage{amsmath}
\usepackage{amssymb}
\usepackage{booktabs}
\usepackage{array}
\usepackage{colortbl}
\usepackage{fvextra}
\usepackage{enumitem}
\usepackage{xcolor}
\iflatexml
  \usepackage{framed} 
\else
  \usepackage[breakable,skins]{tcolorbox} 
\fi
\usepackage{tikz} 
\usepackage{subcaption}
\usepackage{placeins}
\usepackage{xfp}

\newenvironment{TextPanel}[1]{%
  \begin{quote}\raggedright
  \textbf{#1}\par\smallskip\ignorespaces
}{\par\end{quote}}

\iflatexml
\newenvironment{HtmlPrompt}[1]{%
  \par\medskip\begingroup
  \normalsize
  \setlength{\FrameRule}{0.4pt}%
  \setlength{\FrameSep}{6pt}%
  \setlength{\parskip}{0pt}%
  \begin{framed}
  \noindent\textbf{#1}\par\smallskip
}{%
  \par\end{framed}%
  \par\endgroup\medskip}

\newenvironment{HtmlPromptText}{%
  \par\begingroup\normalfont\normalsize\rmfamily\raggedright
  \setlength{\parindent}{0pt}%
  \setlength{\parskip}{0pt}%
  \noindent\ignorespaces
}{\par\endgroup}

\else
  \newenvironment{PromptVerbatim}[2][]{\VerbatimEnvironment
  
  \begin{tcolorbox}[
    enhanced jigsaw,
    breakable,
    title={#2},
    colback=black!2,
    colbacktitle=black!8,
    colframe=black!35,
    coltitle=black,
    fonttitle=\scriptsize,
    boxrule=0.5pt,
    arc=1mm,
    left=2mm,
    right=2mm,
    top=1.5mm,
    bottom=1.5mm,
    before skip=0.5\baselineskip,
    after skip=0.75\baselineskip]
  \begin{Verbatim}[
    breaklines=true,
    breakanywhere=true,
    breakautoindent=false,
    breakindent=0pt,
    breaksymbolleft={},
    breaksymbolsepleft=0pt,
    breaksymbolindentleft=0pt,
    fontsize=\tiny,#1]%
}{%
  \end{Verbatim}
  \end{tcolorbox}}

\fi

\newcommand{\metriccell}[2]{\shortstack[c]{#1\\{\tiny [#2]}}}
\newcommand{\metriccellstar}[2]{\shortstack[c]{#1\textsuperscript{*}\\{\tiny [#2]}}}
\newcommand{\metricna}{--}
\newcommand{\secondbestmetric}[2]{\textcolor{blue}{\metriccell{#1}{#2}}}

\definecolor{virA}{HTML}{440154}
\definecolor{virB}{HTML}{3B528B}
\definecolor{virC}{HTML}{21918C}
\definecolor{virD}{HTML}{5EC962}
\definecolor{virE}{HTML}{FDE725}

\newcommand{\viridisstrength}{0.5}

\newcommand{\scorecell}[2]{\begingroup
  \edef\s{\fpeval{max(2,min(5,#1))}}\ifdim\s pt<2.75pt
    \def\leftcolor{virA}\def\rightcolor{virB}\edef\t{\fpeval{(\s-2)/0.75}}\else\ifdim\s pt<3.5pt
    \def\leftcolor{virB}\def\rightcolor{virC}\edef\t{\fpeval{(\s-2.75)/0.75}}\else\ifdim\s pt<4.25pt
    \def\leftcolor{virC}\def\rightcolor{virD}\edef\t{\fpeval{(\s-3.5)/0.75}}\else
    \def\leftcolor{virD}\def\rightcolor{virE}\edef\t{\fpeval{(\s-4.25)/0.75}}\fi\fi\fi
  \edef\mix{\fpeval{round(100*\t,4)}}\edef\strength{\fpeval{round(
    100*\viridisstrength,4)}}\edef\paintcell{\noexpand\cellcolor{\leftcolor!\fpeval{100-\mix}!\rightcolor
      !\strength!white}}\paintcell
  \metriccell{#1}{#2}\endgroup
}

\title{Steering Readerly Dynamics: Trajectory-Guided Story Generation with Large Language Models}

\date{}

\author{%
  Danrui Li\\
  Department of Computer Science\\
  Rutgers University, USA\\
  \texttt{danrui.li@rutgers.edu}
  \and
  Yiwei Zhao\\
  Netflix, USA\\
  \texttt{yiweiz@netflix.com}
  \and
  Mubbasir Kapadia\\
  Department of Computer Science\\
  Rutgers University, USA\\
  \texttt{mk1353@cs.rutgers.edu}
  \and
  Lingxiao Li\\
  Netflix, USA\\
  \texttt{lingxiaol@netflix.com}
}

\hypersetup{
pdftitle={Steering Readerly Dynamics: Trajectory-Guided Story Generation with Large Language Models},
pdfauthor={Danrui Li, Yiwei Zhao, Mubbasir Kapadia, Lingxiao Li},
}

\begin{document}

\maketitle

\begin{abstract}
Large language models (LLMs) are increasingly used to generate stories, but these stories struggle to reproduce engaging \emph{readerly dynamics}, the evolving reader responses that shape narrative experience.
To address this challenge, we propose a framework that learns readerly dynamics from high-quality human-written stories and uses them to guide story generation. 
Instead of directly modeling high-dimensional story content,
we train a small LLM to predict future reader responses as low-dimensional trajectories. 
We then leverage the strong instruction-following capability of a larger, off-the-shelf writer LLM to generate story segments that follow these predicted trajectories within a given story setting.
Our framework improves trajectory-based fidelity to human-written stories by 12\%--35\% while largely preserving story coherence.
\end{abstract}

\addtocontents{toc}{\protect\setcounter{tocdepth}{-1}}

\section{Introduction}

As a story unfolds, readers experience evolving reader responses, encompassing cognitive, affective, and evaluative aspects~\cite{phelan1996narrative}. The time series of reader responses forms \emph{readerly dynamics}, which are central to human entertainment, from football matches to detective novels~\cite{ely2015}. In storytelling, these dynamics are particularly important for creative industries because they are positively associated with readers' immersion and subsequent social-sharing intentions~\cite{Winkler04032023}.

Despite the wide use of LLMs for story generation, producing high-quality readerly dynamics remains difficult.
Prior work has identified differences between LLM-generated and human-written stories in higher-level narrative properties, such as plot structures~\cite{russell2026storyscope} and affective dynamics~\cite{li2026narrativeflattening}. 
Moreover, conventional post-training such as RLVR/DPO typically optimizes for coherent, readable, and conventionally well-formed text, which suppresses dramatic dynamics in creative writing~\cite{li2026narrativeflattening}.
These findings highlight a gap between generating logically consistent narrative text and reproducing the evolving reader experiences characteristic of compelling stories.

\begin{figure}[htbp]
    \centering
    \includegraphics[width=\linewidth]{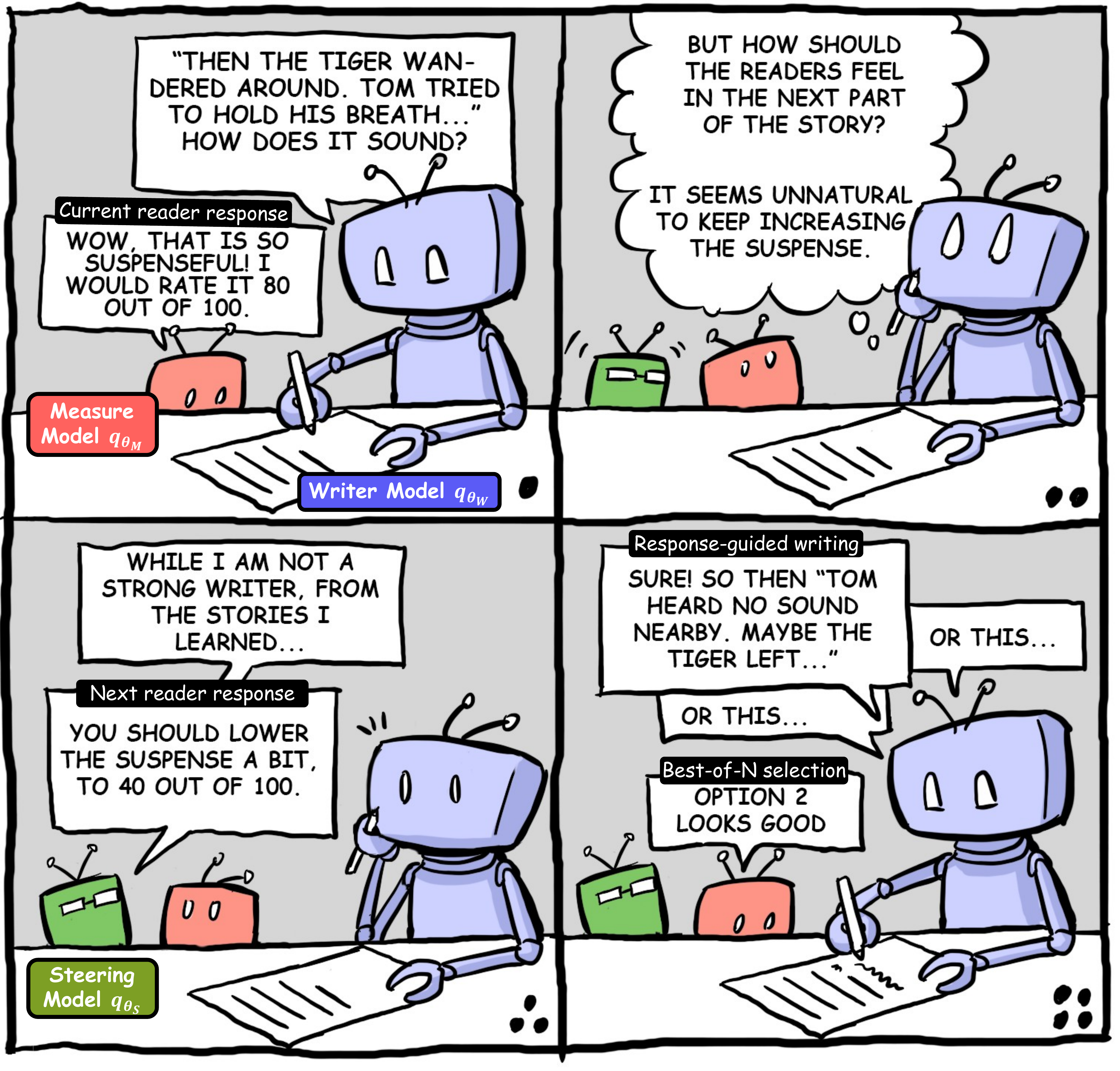}
    \caption{Our work finetunes a small LLM to steer off-the-shelf larger LLM in writing stories segment by segment, enabling more natural readerly dynamics in the narration.
    }
    \label{fig:teaser}
\end{figure}

While there is no single ideal trajectory of reader responses, high-quality human-written narratives provide examples of how such responses unfold over time. This motivates learning readerly dynamics directly from these narratives and using them as guidance during generation. However, our studies (see \S~\ref{sec:results_main}) show that in-context learning is insufficient to reliably induce them, while direct supervised fine-tuning on story text improve dynamics fidelity at the expense of story coherence. These findings suggest that implicitly learning readerly dynamics through story generation alone is insufficient to achieve both desirable reader experiences and coherent narratives.

We therefore explore a different approach: rather than directly learning how to write story text, we explicitly learn the distribution of how the readers will response \emph{in the future}, based on readerly dynamics extracted from high-quality story dataset. Then we use this knowledge to guide story generation.
Our work unfolds as three steps: first, we extract readerly dynamics data from human-written stories, using off-the-shelf LLMs with calibrated system prompts as a proxy to human readers. Next, we use the extracted data to finetune a small LLM that predicts future reader responses from an unfinished story.
Finally, we used the predicted responses as a steering signal to guide a frontier LLM, which has strong instruction following capabilities, to generate the story content for the next segment. This process repeats in an auto-regressive manner, as shown in Fig~\ref{fig:teaser}.  

Experiments across two reader response domains (suspense and humor) with two writer LLMs (GPT-5.5 and Qwen) demonstrate that our framework improves trajectory-based fidelity to high-quality human-written stories by 12\%--35\% while largely preserving story coherence (see Fig.~\ref{fig:quality-fidelity-tradeoff}). 
In contrast, in-context learning fails to improve readerly dynamics fidelity in most evaluated settings, with gains below 5\% even in the few cases where improvements occur.
Direct supervised fine-tuning can achieve competitive dynamics fidelity in some settings, but incurs coherence degradation of up to 30\%.
On the contrary, our modular approach separates readerly dynamics learning from story generation, enabling plug-and-play integration with future frontier writer LLMs.
Our framework also enables generalization to different readerly dynamics, cross-topic transfer, and interactive generation shown in \S~\ref{app:flexibility}, highlighting its potential for broader reader-response-aware storytelling applications.

\section{Related Work}

\paragraph{Common topics in LLM-based story generation.} There is rich literature in using LLM agentic systems to create long stories that is coherent and in close compliance to user instructions. Top-down approaches typically starts with a story plan and implements it with hierarchical generations~\cite{fan-etal-2018-hierarchical,yao-etal-2019-plan,yang-etal-2022-re3,yang-etal-2023-doc,mirowski-etal-2023-dramatron,li2024word2world,huot2025agentsroom}. Other studies also integrate bottom-up agentic designs that maintain modular story contexts. When the context is modularized by story characters, the frameworks enable more natural role-playing~\cite{shao-etal-2023-character,han-etal-2024-ibsen,chen-etal-2024-hollmwood,ran-etal-2025-bookworld}. 

\paragraph{Measuring readerly dynamics.} Prior studies have modeled reader responses through computational representations of narrative structure, such as protagonists' action spaces~\cite{xie-riedl-2024-suspenseful}, predicted story futures~\cite{wilmot-keller-2020-modelling,fong2024storyexpectations,sui2026spoiler}, and symbolic story events~\cite{cheong-young-2015-suspenser,doust-piwek-2017-model}. However, they are domain-specific and hard to generalize. 
Recent work instead uses LLMs to directly model readers' mental states~\cite{bissell-etal-2025-theoretical}. While prior studies have aligned LLMs with holistic reader responses~\cite{fein2025litbench,teleki-etal-2025-survey}, few capture how these responses evolve as stories unfold~\cite{matlin2025suspense,romanowski-etal-2025-punchlines}. 
We go beyond post hoc measurement by predicting future reader responses to guide story generation, enabling readerly-dynamics-in-the-loop.

\paragraph{Controlling story dynamics.}
Prior work has explored segment-wise control of story content and emotional arcs through agentic systems~\cite{wang2026ecscript} or learned guidance models~\cite{gurung2025learning}. Most closely related, SWAG~\cite{pei2024swag} generates editing instructions during segment-wise story generation. However, these general-purpose actions fail to explicitly target natural readerly dynamics, and their steering effectiveness remains underexplored. These approaches largely follow director-actor paradigms~\cite{kim2025codi,han-etal-2024-ibsen}, whereas our work grounds generation control in narrative theory by explicitly modeling readerly dynamics.

Prior work has applied reinforcement learning (RL) for better control over plot goals~\cite{tambwekar-etal-2019-controllable,alabdulkarim-etal-2021-goal}, narrative coherence~\cite{lu-etal-2023-narrative}, protagonist emotion arcs~\cite{brahman-chaturvedi-2020-modeling}, and holistic story quality~\cite{castricato-etal-2022-robust}. Unlike these approaches, which optimize stories toward \emph{predefined} targets, our work focuses on \emph{predicting better targets}: how reader responses evolve, and using these predictions to guide generation.

\section{Method}
\label{sec:method}

Following \citet{lehne2015reading}, we view reader responses as evolving as a story unfolds, such that a reader response can be obtained from a story truncated at any position. To approximate this process discretely, we divide a story into contiguous text segments, $(x_0,\ldots,x_T)$, and represent its readerly dynamics as $(r_0,\ldots,r_T)$, where $r_t$ reflects the reader response after reading the prefix $x_{0:t}$. 
Our framework is flexible to segmentation methods, which can be tailored for various domains (see \S~\ref{sec:data_process}). Here $r_t$ can be either numeric values (throughout the main paper) or structured text (see \S~\ref{app:generality}). 

Then an autoregressive story continuation task with explicit readerly dynamics modeling is defined as: given an unfinished story $x_{0:t}$ and its readerly dynamics ${r}_{0:t}$,
jointly predict its future reader response $r_{t+1}$  and story segment $x_{t+1}$, framed as 
$p(x_{t+1}, {r}_{t+1} \mid x_{0:t},  r_{0:t})$.

\begin{figure*}[htbp]
\begin{subfigure}[t]{.49\linewidth}
\centering
    \includegraphics[width=\linewidth]{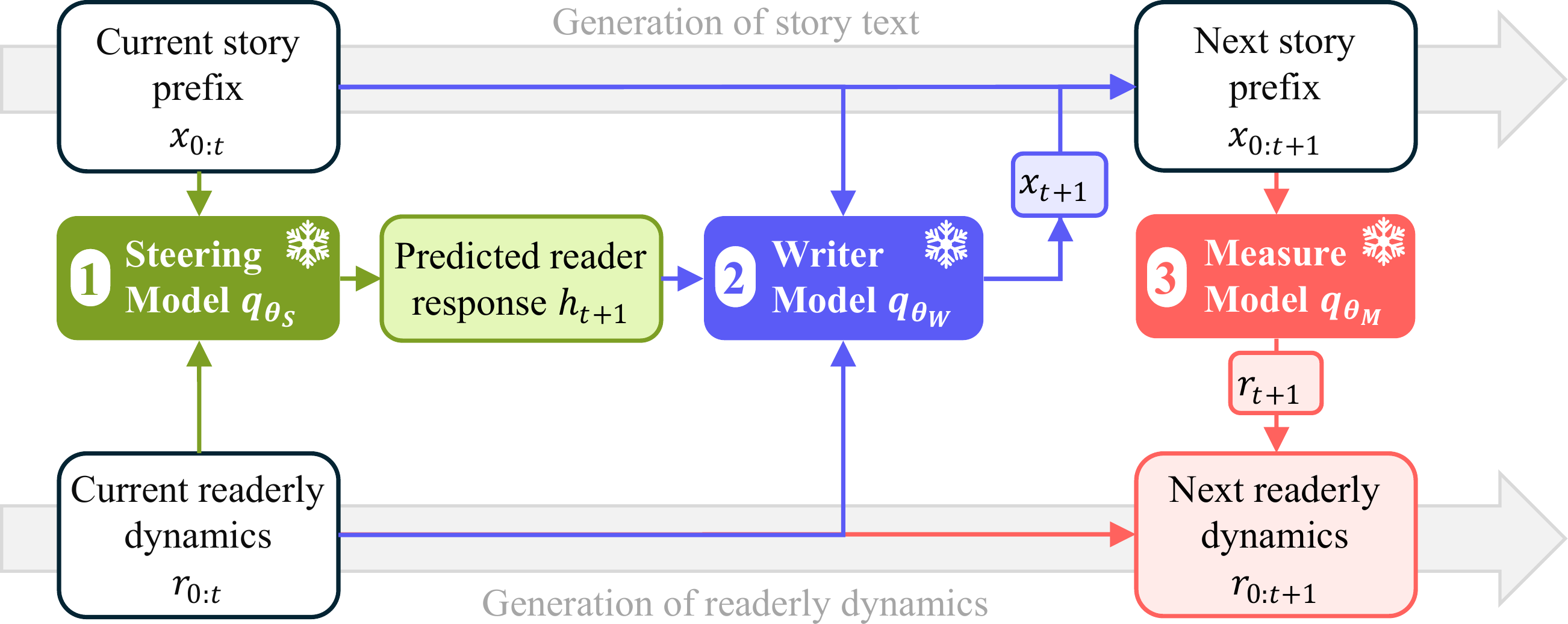}
    \caption{Inference scheme.}
\end{subfigure}
\hfill
\begin{subfigure}[t]{.49\linewidth}
\centering
    \includegraphics[width=\linewidth]{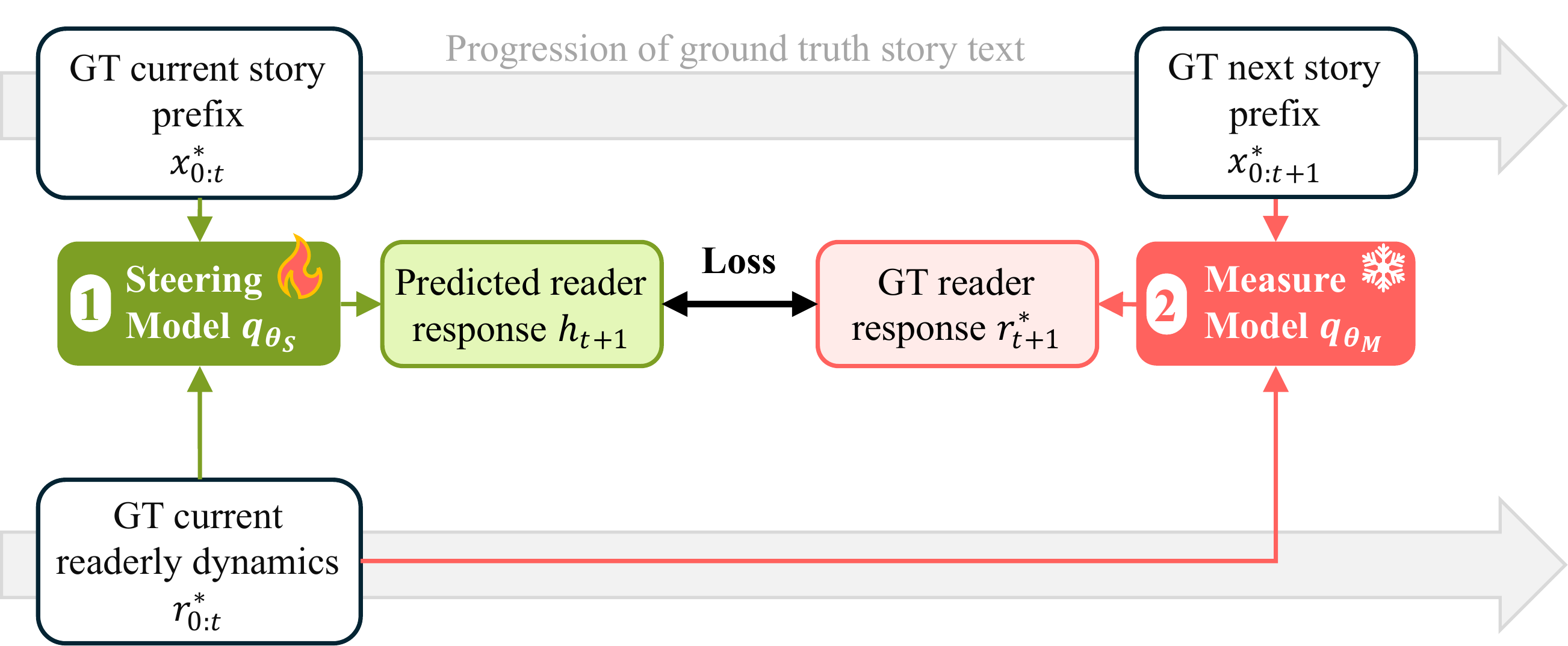}
    \caption{Training scheme.}
\end{subfigure}
    \caption{Our framework jointly generate story text and readerly dynamics (left), where the steering model (green) is finetuned using the readerly dynamics extracted from human stories (right). 
    }
    \label{fig:framework}
\end{figure*}

Instead of directly generating text $x_{t+1}$, we break the generation into three components (see Fig.~\ref{fig:framework}):
\begin{itemize}[leftmargin=1.2em,label={},labelsep=0pt,itemsep=1pt, topsep=2pt]
\item 
\textbf{Steering Model}: Learning guidance signal from high-quality human data, $q_{\theta_S}$ predicts the reader response of the next segment $h_{t+1}$, given $x_{0:t}$ and $r_{0:t}$.
\item
\textbf{Writer Model}: A strong LLM $q_{\theta_W}$ generates story content $x_{t+1}$, using $h_{t+1}$ as additional guidance.
\item 
\textbf{Measure Model}: Another LLM $q_{\theta_M}$ measures the actual reader response $r_{t+1}$ given $x_{0:t+1}$ and readerly dynamics $r_{0:t}$, as $r_{t+1}$ may slightly deviate from the planned reader response $h_{t+1}$.
\end{itemize}
Formally, we model the evolution of stroy segments and reader responses as: 
\begin{equation}
\label{eq:method}
    \begin{aligned}
    & p(x_{t+1},  r_{t+1}\mid x_{0:t}, r_{0:t}) \\
    & =
      q_{\theta_S}(h_{t+1} \mid x_{0:t},r_{0:t})\\
    & \qquad  \times
      q_{\theta_W}(x_{t+1}\mid
        x_{0:t},r_{0:t},h_{t+1}). \\
    & \qquad  \times 
    q_{\theta_M}( r_{t+1}
                    \mid x_{0:t+1}, r_{0:t})            
    \end{aligned}                
\end{equation}

Our motivation for this separation is twofold: First, readerly dynamics, as a lower-dimensional features of the narratives, are easier to model than high-dimensional story texts. Secondly, this supports plug-and-play integration of the writer and measure model should better models become available.

\subsection{Steering Model}
The steering model $q_{\theta_S}$ predicts the reader response $h_{t+1}$ to the next story segment. It takes as input the story prefix $x_{0:t}$, readerly dynamics ${r}_{0:t}$, and auxiliary information such as planned length and genre. We train it using reader responses measured from the corresponding ground-truth next segments ${r}_{t+1}$ as supervision (Fig.~\ref{fig:framework} right). Since off-the-shelf LLMs fail to make good predictions (see \S~\ref{app:steering-ablation}), we finetune Qwen3.5-27B using LoRA. Prompts and training and inference configurations are detailed in \S~\ref{app:steering-config}.

\subsection{Writer Model}

Our framework does not require a particular writer model for $q_{\theta_W}$ as long as it has strong instruction-following  capability.
In this paper, we consider both Qwen3.5-27B and GPT-5.5. In their system prompt, we include the instructions for general story writing, story prefix $x_{0:t}$, readerly dynamics, structured output of the steering model $h_{t+1}$, and other helper information. See \S~\ref{app:our-writer} for details.

As off-the-shelf LLMs may not strictly follow the instructed reader experience, we adopt a best-of-$N$ approach to choose the best answer from $N=6$ independent LLM samplings.
For the $n$th model output, we use $q_{\theta_M}$ to sample:
\begin{equation}
  r_{t+1}^{(n)}
  \sim q_{\theta_M}(\cdot \mid [x_{0:t}, x_{t+1}^{(n)}], r_{0:t}).
\end{equation}
Given the $q_{\theta_S}$'s request $h_{t+1}$, the
selected slot is
\begin{equation}
  n_{t+1}^{*}
  =\underset{n}{\arg\min}\;
    \left|r_{t+1}^{(n)}-h_{t+1}\right|,
\label{eq:best-of-n}
\end{equation}

\subsection{Measure Model}
\label{sec:measurement-model}

We instantiate $q_{\theta_M}$ with an off-the-shelf LLM (\texttt{GPT-5.4-mini}) with customized system prompts and temperature 0. The model
reads story prefix $x_{0:t+1}$ and readerly dynamics $r_{0:t}$ and outputs the expected reader response after reading the updated story $r_{t+1}$.
The representation of $r_{t+1}$ is flexible and up to user's decision.
We choose two reader response domains (suspense and humor) in main experiments, representing them as an integer ranging from 0 to 100, together with a one-sentence reason in text. See \S~\ref{sec:measure-quality} for our alignment to human judgment and \S~\ref{app:measure-model} for experiments on segmentation granularity, the LLM choice and response representation.

\section{Evaluation}

We evaluate all methods on a test story set $S$. For each story $s \in S$, we use its story prefix $(x_{0:t}, r_{0:t})$ as framework input and receive $\{(x_{t+1}, r_{t+1}),\ldots,(x_T, r_T)\}$ as output. The $t$ here is uniformly sampled from $4$ to the $T-1$, where $T$ is the total segment count for the story $s$.

\subsection{Dataset}

We study our framework in two readerly dynamics domains: humor and suspense.

\subsubsection{Humor domain}
We used FunnyBench dataset~\cite{chaubard2026funnybench}.
It contains short textual clips of standup comedy shows, posing a challenge to capture the switching between setups and punchlines from professional comedian acting.
We filtered out non-narrative content, too long/short content and near duplicates. For each clip, we segmented it into 30-60-word chunks, cut by a set of symbols such as periods. For each segment, we use GPT-5.4-mini to assign 0--100 humor scores that represent the level of humor the readers might feel when reading the current segment. See \S~\ref{sec:measure-quality} for measurement quality analysis. Finally, we got a humor dataset containing 4{,}012 clips and 25{,}712 segments, where 514 of the clips are in the test split. 

\subsubsection{Suspense domain}
We collected stories from FlawedFictions~\cite{ahuja2025flawedfictions}, Gutenberg-genre collection~\cite{projectgutenberg}, and the STORAL dataset~\cite{guan-etal-2022-storal}. We aim to evaluate whether our method can capture fluctuations in narrative tension throughout a story.
We applied a process similar to that used for the humor dataset, except that we used semantic-clustering-based segmentation~\cite{jia2026embedkcpd} with LLM-generated summaries to eliminate the effect of word usage and content length. The suspense dataset contains 1{,}733 stories with 12{,}834 segments, 220 stories remain in the final test split.
Please see \S~\ref{sec:data_process} for details.

\subsection{Evaluation Regimes}
\label{sec:eval-regime}

These laddered evaluation regimes aim to test the framework performance when each component is added to consideration sequentially.

\noindent\textbf{Position-Wise Prediction.} This regime evaluates whether we can get good prediction of readerly dynamics using $q_{\theta_S}$, when not considering the effects of writer model $q_{\theta_W}$ and compounding errors during the segment-wise generation. It is a sanity check where we always input ground truth story and reader experience data.
In other words, we sample $h_{t+1}^{\text{clean}} \sim q_{\theta_S}( \cdot \mid x_{0:t}^*, r_{0:t}^*)$ where $x_{0:t}^*, r_{0:t}^*$ are ground truth data, and compare it with the ground truth annotation $r_{t+1}^*$.

\noindent\textbf{Position-Wise Writing.}
This regime further adds the impact of writer $q_{\theta_W}$ and the measuring of the generated output.
It evaluates if the good prediction leads to good writing result, given the authentic human story prefix. Formally, we firstly get $x_{t+1}^{\text{clean}} \sim q_{\theta_W}(\,\cdot\mid x_{0:t}^*, r_{0:t}^*,h_{t+1}^{\text{clean}})$. Then we measure its reader experience as $ r_{t+1}^{\text{clean}}$  and compare it to the ground truth $r_{t+1}^*$.

\noindent\textbf{Autoregressive.}
This regime evaluates the complete pipeline, which composes all components with compound generation errors during auto-regressive generation.
Given the story prefix, it rolls out the story segment by segment using Eq.~\ref{eq:method}. Then we compare $r_{t+1:T}$ to ground truth $r_{t+1:T}^*$ for each future position.

\subsection{Metrics for Dynamics Fidelity}
\label{sec:eval-metrics}

Because a story prefix may have multiple valid continuations, we adapt evaluation metrics from the trajectory-prediction literature~\cite{liu2024trajdiffuse}.
For each test story $s$, we independently sample $K=10$ complete readerly-dynamics trajectories $r_{t+1: T}^{(k)}$ in the autoregressive regime; in the other regimes, we sample $K$ reader-response values at each future position.

For position-wise comparisons, Min-$K$ $\mathrm{MAE}_s$ measures how closely the best of $K$ independently sampled outputs matches the ground-truth value at position $t$. We select the closest sample separately at each position and then average those errors within the story:
\begin{equation}
\mathrm{MAE}_s^{\text{pos}}
=\frac{1}{|\mathcal{T}_s|}\sum_{t\in\mathcal{T}_s}
  \min_{1\leq k\leq K}
  \left|{r}^{(s,k)}_{t}-r_{t}^{*(s)}\right|.
\end{equation}
This is a local best-of-$K$ measure: $k$ may change with $t$ because position-wise outputs are independently conditioned and do not form a single generated trajectory. 

For autoregressive rollouts, we instead measure Min-$K$ $\mathrm{MAE}_s^{\text{AR}}$,
\begin{equation}
\mathrm{MAE}_s^{\text{AR}}
= \min_{1\leq k\leq K}
  \left[
    \frac{1}{|\mathcal{T}_s|}\sum_{t\in\mathcal{T}_s}
    \left|{r}^{(s,k)}_{t}-r_{t}^{*(s)}\right|
  \right],    
\end{equation}
which chooses the $k$ once per story and retains it at every
position. 

Apart from retaining only the best result as MAE,
we also report other two metrics to evaluate the output distributions of one single story and those of entire test split in Appendix~\ref{app:metric-crps}.

\subsection{Metrics for General Quality}
\label{sec:eval-llm}

Following the method-blind evaluation protocol of HANNA \cite{chhun2022hanna,chhun2024llmjudge}, we evaluate each complete autoregressive story using LLM judge across six dimensions. 
We focus on \emph{story coherence} to assess whether our method, which retains off-the-shelf frontier models as writers, can preserve the narrative consistency and logical reasoning capabilities of writer models. 
Results for the remaining five dimensions and LLM judge details are reported in \S~\ref{app:main-results} and \S~\ref{app:quality-eval}

\subsection{Baseline Methods}
\label{sec:basline}

We compare our method to the following baseline methods. See \S~\ref{app:method-details} for details.

\noindent\textbf{Off-the-shelf.} We directly use off-the-shelf LLMs to write story continuations segment by segment with minimal instructions $c$, i.e. $x_{t+1} \sim q_{\theta_W}(\cdot \mid x_{0:t}, c)$.

\noindent\textbf{One-shot In-context Learning (ICL).} For each task, we retrieve a story example from the training data to include in the LLM prompt. The story example is selected by semantic similarity to the story prefix of the test item. i.e. $ x_{t+1} \sim q_{\theta_W}(\cdot\mid x_{0:t}, c, x^{\text{ref}}_{0:T^{\text{ref}}}, c^{\text{ref}})$

\noindent\textbf{Supervised Finetuning (SFT).} We finetune the writer model with LoRA adapters, using story text in the training set. During inference, we directly use this writer model to complete story continuation, i.e. $x_{t+1} \sim q_{\theta_{\text{SFT}}}(\cdot\mid x_{0:t}, c)$.

\noindent\textbf{SWAG.} We reimplement the prior work of \citet{pei2024swag}, which is conceptually framed as $s_t \sim q_{\theta_S}(\cdot\mid x_{0:t})$ and ${x}_{t+1} \sim q_{\theta_W}(\cdot\mid x_{0:t}, s_t)$, where $s_t$ is chosen from a fixed writing instruction set. Here $q_{\theta_S}$ is post-trained by SFT and Direct Preference Optimization (DPO). To ensure a fair comparison, we align its base LLM for $q_{\theta_S}$ and $q_{\theta_W}$ to our setting.

\section{Results}

We present our main results in \S~\ref{sec:results_main}. Then we discuss the effect of LLM-based measurement errors in \S~\ref{sec:measure-quality} and ablation studies on key components and conditioning in \S~\ref{sec:results_ablation}.

\subsection{Main results}
\label{sec:results_main}

Fig.~\ref{fig:common-core-humor-setup} illustrates how different methods continue a shared opening in the humor domain. While the off-the-shelf LLM tends to introduce a punchline in nearly every segment, both human-written stand-up comedy and our generated stories exhibit a more natural alternation between setups and punchlines. By allowing moments of anticipation and relief rather than continuously delivering jokes, our method better captures the temporal structure of human readerly dynamics.

\iflatexml
  \definecolor{setupLabelsetup}{HTML}{2E7D32}
\definecolor{setupLabelpunchline}{HTML}{C62828}

\newcommand{\setuphumorlabel}[1]{\textcolor{setupLabel#1}{\textbf{#1}}}
\newcommand{\setuphumorcell}[3]{{\footnotesize
    \noindent\textbf{#1}\quad\setuphumorlabel{#2}\par
    \raggedright #3\par}}

\par\medskip\begingroup
\setlength{\FrameRule}{0.4pt}
\setlength{\FrameSep}{6pt}

\includegraphics[width=\linewidth]{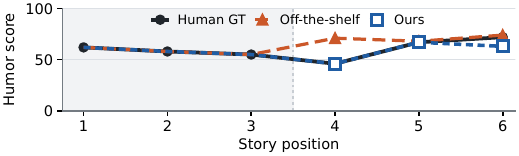}
\par\medskip

\begin{framed}
\setuphumorcell{Shared opening (Position 3 of 6)}{setup}{If you don't know common core math, it's just basically a long way to get to the same answer. It's like you. It's like how can we use the entire sheet of paper for this math problem?}
\par\medskip
\par

\end{framed}

\begin{framed}
\noindent\textbf{Human (GT)}\par\smallskip

\setuphumorcell{Position 4}{setup}{And even when you break it down, you see old math sneak in there. I mean, you're you're like that was old math right there. Just do that at the beginning and we won't even have to be in this there's.}

\setuphumorcell{Position 5}{punchline}{I told my wife it's a long way to get to the same thing. It would be if you came to my house and you knocked on the front door and you're like: can I come in? And I was like, could you walk around and come in through the back door?}
\par\medskip

\end{framed}

\begin{framed}
\noindent\textbf{Off-the-shelf Qwen3.5-27B}\par\smallskip

\setuphumorcell{Position 4}{punchline}{So you're just filling the whole page with circles and lines and arrows. You finally get the answer at the bottom, right? And the kid looks at you like, "Did I do it right?" And you're just sweating because you have no idea if that answer is right or if you accidentally drew a map to Mars.}

\setuphumorcell{Position 5}{punchline}{So you just have to guess and hope for the best. You say, "Yeah, great job, look at all those arrows!" You are terrified they are going to ask for your work and you're just going to point at the doodles. It's not math anymore; it's abstract art.}
\par\medskip

\end{framed}

\begin{framed}
\noindent\textbf{Ours (Qwen3.5-27B)}\par\smallskip

\setuphumorcell{Position 4}{setup}{It's just a massive puzzle where the answer is buried under forty steps of nonsense. You spend the whole time trying to figure out the method instead of actually solving the problem. You're just scribbling away, and by the time you get the answer, you've forgotten what question you were even asking in the first place.}

\setuphumorcell{Position 5}{punchline}{So you finally get twenty-seven, but you have to circle it three times, draw a pyramid around it, and write a haiku explaining why it feels that way. The teacher grades the pyramid and the poem, not the math. You're not a mathematician anymore; you're an artist with a calculator, begging for validation on a drawing that says the answer is obvious.}
\par\medskip
\end{framed}
\captionof{figure}{Qualitative example showing humor dynamics across methods. The plot shows LLM-measured humor levels for all story positions. The shared opening and continuations from Human (GT), Off-the-shelf Qwen3.5-27B, and Ours (Qwen3.5-27B) follow in reading order. Green \textcolor{setupLabelsetup}{setup} and red \textcolor{setupLabelpunchline}{punchline} are editorial labels.
}
\label{fig:common-core-humor-setup}
\par\endgroup\medskip

\else
  \definecolor{setupLabelsetup}{HTML}{2E7D32}
\definecolor{setupLabelpunchline}{HTML}{C62828}

\newcommand{\setuphumorlabel}[1]{\textcolor{setupLabel#1}{\textbf{#1}}}
\newcommand{\setuphumorcell}[3]{{\fontsize{7.2}{8.7}\selectfont
    \noindent\textbf{#1}\hfill\setuphumorlabel{#2}\par
    \vspace{0.35mm}\raggedright #3\par}}

\begin{figure*}[!htbp]
\centering
\begin{minipage}[t]{0.49\textwidth}\vspace{0pt}
\includegraphics[width=\linewidth]{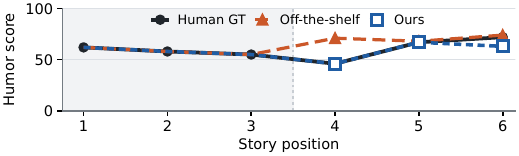}
\end{minipage}\hfill
\begin{minipage}[t]{0.47\textwidth}\vspace{0pt}
\setuphumorcell{Shared opening (Position 3 of 6)}{setup}{If you don't know common core math, it's just basically a long way to get to the same answer. It's like you. It's like how can we use the entire sheet of paper for this math problem?}
\end{minipage}
\par\vspace{1.3mm}
\begin{subfigure}[t]{0.265\textwidth}\vspace{0pt}
\caption{Human (GT)}
\vspace{1.2mm}
\setuphumorcell{Position 4}{setup}{And even when you break it down, you see old math sneak in there. I mean, you're you're like that was old math right there. Just do that at the beginning and we won't even have to be in this there's.}
\vspace{1.6mm}
\setuphumorcell{Position 5}{punchline}{I told my wife it's a long way to get to the same thing. It would be if you came to my house and you knocked on the front door and you're like: can I come in? And I was like, could you walk around and come in through the back door?}
\end{subfigure}\hfill
\begin{subfigure}[t]{0.31\textwidth}\vspace{0pt}
\caption{Off-the-shelf Qwen3.5-27B}
\vspace{1.2mm}
\setuphumorcell{Position 4}{punchline}{So you're just filling the whole page with circles and lines and arrows. You finally get the answer at the bottom, right? And the kid looks at you like, "Did I do it right?" And you're just sweating because you have no idea if that answer is right or if you accidentally drew a map to Mars.}
\vspace{1.6mm}
\setuphumorcell{Position 5}{punchline}{So you just have to guess and hope for the best. You say, "Yeah, great job, look at all those arrows!" You are terrified they are going to ask for your work and you're just going to point at the doodles. It's not math anymore; it's abstract art.}
\end{subfigure}\hfill
\begin{subfigure}[t]{0.355\textwidth}\vspace{0pt}
\caption{Ours (Qwen3.5-27B)}
\vspace{1.2mm}
\setuphumorcell{Position 4}{setup}{It's just a massive puzzle where the answer is buried under forty steps of nonsense. You spend the whole time trying to figure out the method instead of actually solving the problem. You're just scribbling away, and by the time you get the answer, you've forgotten what question you were even asking in the first place.}
\vspace{1.6mm}
\setuphumorcell{Position 5}{punchline}{So you finally get twenty-seven, but you have to circle it three times, draw a pyramid around it, and write a haiku explaining why it feels that way. The teacher grades the pyramid and the poem, not the math. You're not a mathematician anymore; you're an artist with a calculator, begging for validation on a drawing that says the answer is obvious.}
\end{subfigure}
\caption{Qualitative example showing humor dynamics across methods. The shared opening is shown once at upper right. The LLM-measured level of humor for all story positions are shown at upper left. Green \textcolor{setupLabelsetup}{setup} and red \textcolor{setupLabelpunchline}{punchline} are editorial labels.
}
\label{fig:common-core-humor-setup}
\end{figure*}

\fi

\begin{figure}[htbp]
\centering

\begin{subfigure}[t]{\linewidth}
\centering
\includegraphics[width=\linewidth]{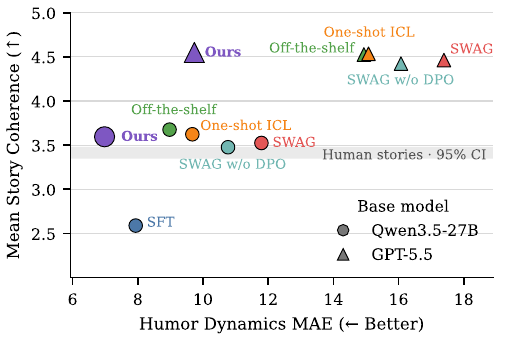}
\caption{Humor.}
\label{fig:humor-quality-fidelity-tradeoff}
\end{subfigure}

\hfill

\begin{subfigure}[t]{\linewidth}
\centering
\includegraphics[width=\linewidth]{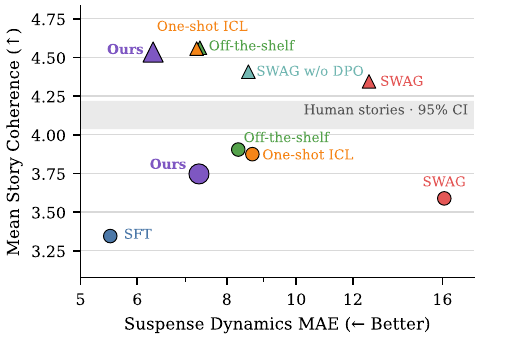}
\caption{Suspense.}
\label{fig:quality-fidelity-tradeoff-suspense}
\end{subfigure}
\caption{Readerly-dynamics fidelity versus Story coherence
for different methods, using two base writer models in suspense and humor.
Lower trajectory error (x axis) and higher coherence (y axis)
are better. Gray bands are 95\% bootstrap intervals for LLM judge's scores on human-authored stories.
}
\label{fig:quality-fidelity-tradeoff}
\end{figure}

We quantify this alignment using Min-K MAE in \S~\ref{sec:eval-metrics} and compare it against story coherence in Fig.~\ref{fig:quality-fidelity-tradeoff}, across suspense and humor with GPT-5.5 and Qwen3.5-27B as writers.

Methods relying on implicit learning of readerly dynamics show limited effectiveness. ICL yields negligible fidelity improvements, while SFT sometimes achieves higher fidelity but substantially degrades coherence. Additional experiments with explicit prompting instructions and extended thinking likewise fail to achieve consistent improvements (see \S~\ref{app:explicit-dynamics-prompting}).

SWAG~\citep{pei2024swag} in its original form (red) worsens performance, while removing its DPO stage (aqua) partially restores it. One possible explanation is that DPO over-optimizes instruction-following patterns at the expense of natural narrative progression, an issue less pronounced with SFT alone.

In contrast, our framework explicitly targets reader-response trajectories, reducing dynamics error by 12\%--35\% while largely preserving coherence. The gains are strongest with GPT-5.5, which achieves greater fidelity improvements with smaller coherence losses, whereas Qwen3.5-27B shows smaller gains and greater coherence sensitivity. This difference reflects the stronger instruction-following capability of GPT-5.5, which better translates reader-response guidance into coherent narratives.

\subsection{User study}
\label{sec:measure-quality}
Ideally, we would directly evaluate generated outputs through human preference studies across methods. However, substantial annotator disagreement would require a prohibitively large labeling budget to obtain reliable estimates. We therefore adopt an indirect evaluation strategy: first validating the LLM-based measure model against human judgments, then using the validated model as a proxy metric to compare generation methods. See details of all user studies in \S~\ref{app:user-study}.

\begin{figure}[htbp]
\centering
\captionsetup[subfigure]{skip=1pt}

\begin{subfigure}[t]{.49\linewidth}
\centering
\includegraphics[width=\linewidth]{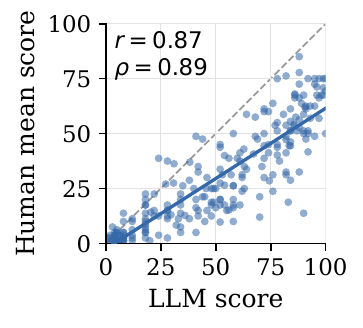}
\label{fig:llm-human-score-scatter-suspense}
\end{subfigure}\hfill\begin{subfigure}[t]{.49\linewidth}
\centering
\includegraphics[width=\linewidth]{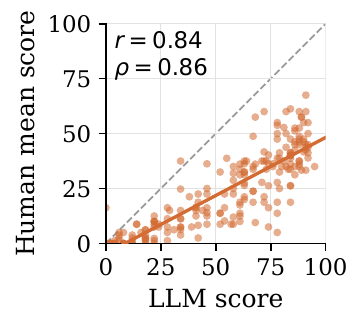}
\label{fig:llm-human-score-scatter-humor}
\end{subfigure}


\begin{subfigure}[t]{.49\linewidth}
\centering
\includegraphics[width=\linewidth]{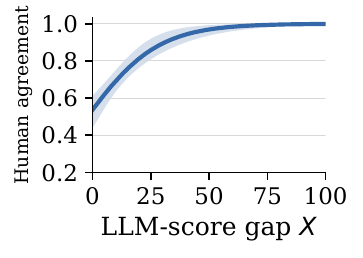}
\caption{Suspense.}
\label{fig:human-direction-probability-suspense}
\end{subfigure}\hfill\begin{subfigure}[t]{.49\linewidth}
\centering
\includegraphics[width=\linewidth]{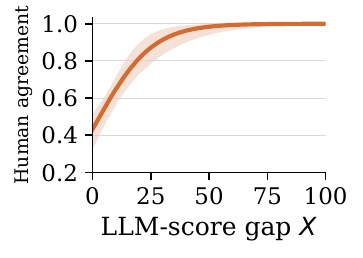}
\caption{Humor.}
\label{fig:human-direction-probability-humor}
\end{subfigure}

\caption{\textbf{Top:} LLM measurement scores versus human mean scores
(rescaled from 1--9 to 0--100) at each story position.
Each point represents a unique story prefix;
$r$ and $\rho$ denote Pearson and Spearman correlations.
\textbf{Bottom:} Probability that human and LLM scores agree
on the direction of change between two story positions,
given an absolute LLM score gap of $X$.
Curves show bootstrap medians with 95\% confidence intervals.}
\label{fig:llm-human-score-comparison}
\end{figure}

We conduct a user study with 10 native English speakers, who rate reader responses at each story position following the same trajectory-based protocol as our measure model. We then compare the mean human rating at each position with the corresponding LLM-predicted score.
Figure~\ref{fig:llm-human-score-comparison} shows that our LLM measure model broadly aligns with human judgments.

To assess its validity, we conduct a bootstrap analysis by randomly sampling two positions from the same story and testing whether the LLM and human scores agree on the direction of change.
Fig.~\ref{fig:llm-human-score-comparison} (bottom) reports this agreement probability as a function of the LLM score difference. 
When the LLM score difference is below 10 on the 0–100 scale, differences in human reader response are difficult to distinguish. Beyond a gap of approximately 20, directional agreement with human judgments increases substantially.

To further validate whether an LLM score gap above 20 provides a reliable signal, we conduct a head-to-head human study. Participants are shown a story opening, its human-written continuation, and two continuations generated by different methods. They select the generated continuation whose readerly dynamics most closely match those of the human continuation, considering both the direction and magnitude of change.

We construct comparison pairs such that one generated continuation is close to the human continuation according to the measure model, with a score gap below 10, while the two generated continuations differ from each other by more than 20 points. This setup tests whether a large model-predicted difference between methods corresponds to a consistent human preference. For each pair, we randomize the presentation order and collect 5 independent judgments. See \S~\ref{app:head-to-head-alignment} for details.

As shown in Table~\ref{tab:head-to-head-human-reader-alignment}, our LLM measure model agrees well with human preferences when the predicted score gap exceeds 20. The agreement becomes even stronger when excluding highly ambiguous human judgments (3:2 votes).

\begin{table}[htbp]
\centering
\footnotesize
\setlength{\tabcolsep}{2pt}
\caption{Agreement between the five-annotator human majority and the measure model scoring. 
}
\label{tab:head-to-head-human-reader-alignment}
\begin{tabular}{@{}>{\raggedright\arraybackslash}p{0.39\columnwidth}cc@{}}
\toprule
 & Suspense & Humor \\
\midrule
All complete rows & 94/114 (82.5\%) & 91/126 (72.2\%) \\
Excluding 3:2 votes & 75/82 (91.5\%) & 65/83 (78.3\%) \\
\bottomrule
\end{tabular}
\end{table}

Additionally, we examine segmentation sensitivity using suspense annotations from \citet{lehne2015reading} (see \S~\ref{app:measure-segmentation}). Our measurement closely follows human judgments at the original resolution. Coarser segmentations reduce agreement, although the overall suspense trajectory remains broadly consistent.

\subsection{Ablations}
\label{sec:results_ablation}

Although our framework adopts comprehensive conditioning (Eq.~\ref{eq:method}), we ablate whether each component benefits from explicit reader-response inputs beyond story text. Experiments use the suspense dataset with GPT-5.5 as the base writer model.

\iflatexml
\begin{figure*}[htbp]
\centering
\begin{subfigure}[t]{0.7\linewidth}
\centering
\includegraphics[width=\linewidth]{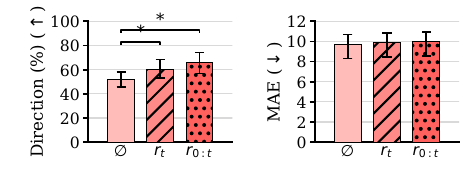}
\caption{Measurement model $q_{\theta_M}$.}
\label{fig:ablation-qm}
\end{subfigure}

\par\smallskip
\begin{subfigure}[t]{0.7\linewidth}
\centering
\includegraphics[width=\linewidth]{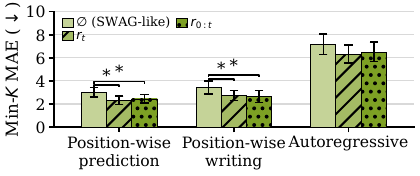}
\caption{Steering model $q_{\theta_S}$.}
\label{fig:ablation-qs-min-k-mae}
\end{subfigure}

\par\smallskip
\begin{subfigure}[t]{0.7\linewidth}
\centering
\includegraphics[width=\linewidth]{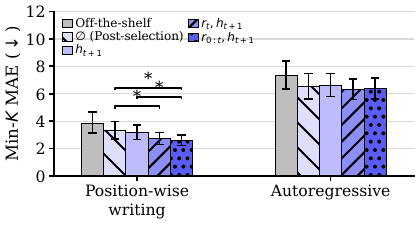}
\caption{Writer model $q_{\theta_W}$.}
\label{fig:ablation-qw-min-k-mae}
\end{subfigure}

\par\smallskip
\begin{subfigure}[t]{0.7\linewidth}
\centering
\includegraphics[width=\linewidth]{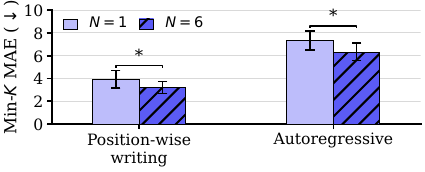}
\caption{Best-of-N in writer model.}
\label{fig:ablation-writer-candidates}
\end{subfigure}
\caption{Ablation studies on the conditioning settings of (a) measurement, (b) steering, (c) writer model, and (d) best-of-N. Use suspense data on GPT-5.5 writer.
Bars show 95\% bootstrap intervals; $*$ marks paired $p<.05$.
}
\label{fig:ablation-min-k-mae}
\end{figure*}

\else
\begin{figure}[htbp]
\centering
\begin{subfigure}[t]{\columnwidth}
\centering
\includegraphics{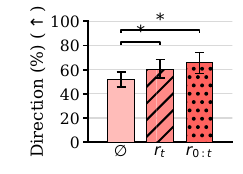}
\hfill
\includegraphics{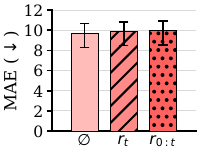}
\caption{Measurement model $q_{\theta_M}$.}
\label{fig:ablation-qm}
\end{subfigure}

\par\vspace{1pt}
\begin{subfigure}[t]{\columnwidth}
\centering
\includegraphics{figs/ablation_qs_min_k_mae.pdf}
\caption{Steering model $q_{\theta_S}$.}
\label{fig:ablation-qs-min-k-mae}
\end{subfigure}

\par\vspace{1pt}
\begin{subfigure}[t]{\columnwidth}
\centering
\includegraphics{figs/ablation_qw_min_k_mae.pdf}
\caption{Writer model $q_{\theta_W}$.}
\label{fig:ablation-qw-min-k-mae}
\end{subfigure}

\par\vspace{1pt}
\begin{subfigure}[t]{\columnwidth}
\centering
\includegraphics{figs/ablation_writer_candidate_count.pdf}
\caption{Best-of-N in writer model.}
\label{fig:ablation-writer-candidates}
\end{subfigure}
\caption{Ablation studies on the conditioning settings of (a) measurement, (b) steering, (c) writer model, and (d) best-of-N. Use suspense data on GPT-5.5 writer.
Bars show 95\% bootstrap intervals; $*$ marks paired $p<.05$.
}
\label{fig:ablation-min-k-mae}
\end{figure}

\fi

\noindent\textbf{Measure model.} 
Fig.~\ref{fig:ablation-qm} shows that while conditioning on readerly-dynamics history does not significantly improves mean absolute error on reader response measurement (right), it does in the prediction of response-change direction (left), suggesting that historical responses capture dynamics beyond story text alone. See \S~\ref{app:ablation} for details.

\noindent\textbf{Steering model.} Fig.~\ref{fig:ablation-qs-min-k-mae} shows that conditioning on the current reader response $\tilde{r}_t$ (hatched bars) reduces prediction errors compared with text-only prediction from $x_{0:t}$ (unhatched bars), which is conceptually analogous to SWAG~\citep{pei2024swag}.
Improvements are significant in position-wise prediction and writing settings, but not in autoregressive settings, potentially due to compounded errors. Conditioning on the full response history $\tilde{r}_{0:t}$ yields no clear additional benefit.

\noindent\textbf{Writer model.} Fig.~\ref{fig:ablation-qw-min-k-mae} shows a similar trend, with explicit reader-response conditioning improving generation performance over the unconditioned baseline ($\varnothing$ in figure legend), which relies solely on best-of-$N$ selection (Eq.~\ref{eq:best-of-n}) to select story candidates based on predicted reader responses. This highlights the complementary benefits of response-guided generations. Furthermore, Fig.~\ref{fig:ablation-writer-candidates} shows that increasing the candidate pool improves alignment with target reader dynamics, demonstrating how best-of-$N$ leverages steering-model predictions to scale performance with inference-time compute. See \S~\ref{sec:guided-vs-post-selection} for extended results.

Overall, these ablations support explicit reader-response conditioning while suggesting that the current response ${r}_t$ suffice without the full history ${r}_{0:t}$. We adopt this simplification for suspense, without assuming it generalizes across domains or base LLMs. See further experiments in \S~\ref{app:ablation}.

\FloatBarrier

\section{Conclusion}

We propose a framework that explicitly models readerly dynamics in story generation, using a learned steering model to predict future reader responses and guide an off-the-shelf LLM writer. Experiments on suspense and humor show improved fidelity to readerly dynamics measured from human-written stories while maintaining story coherence. 
Our findings highlight the potential for more adaptive and personalized storytelling, where
future work should address error accumulation in long-form generation and improve measurement reliability through broader human validation.

\section*{Limitations}

\iflatexml
\medskip
\noindent\textbf{Limited human annotation}
\else
\paragraph{Limited human annotation}
\fi
Human judgments of readerly dynamics exhibit substantial variability, requiring multiple annotators per instance to obtain reliable estimates. Given the high annotation cost, we do not conduct a full-scale human evaluation which directly compares stories generated by different methods. Instead, we validate our LLM-based measurement model against human judgments and use it as a proxy for evaluating generated stories. 

\iflatexml
\medskip
\noindent\textbf{Imperfect LLM measurement}
\else
\paragraph{Imperfect LLM measurement}
\fi
Our experiments show that LLM-based measurements of readerly dynamics do not perfectly align with aggregated human judgments. Although the measure model generally agrees with human judgments when differences in reader responses are pronounced, it is less reliable at distinguishing subtle variations. Consequently, our evaluation may overlook finer-grained differences in readerly dynamics, and the learned steering signals may be affected by measurement noise.

\iflatexml
\medskip
\noindent\textbf{Single-Dimensional Steering}
\else
\paragraph{Single-Dimensional Steering}
\fi
Our experiments consider one dimension of readerly dynamics at a time, such as suspense or humor. However, reader experiences are inherently multidimensional, with different responses potentially interacting throughout a narrative. It remains unclear whether off-the-shelf frontier LLMs can effectively follow multiple simultaneous steering signals. Extending our framework to multidimensional readerly dynamics and investigating potential conflicts between different response dimensions are promising directions for future work.

\section*{Ethical considerations}
Reader responses are inherently subjective and may vary across individuals and cultural backgrounds. Our framework relies on LLM-based proxies to estimate reader responses and learns readerly dynamics from a selected collection of human-written stories. Consequently, the learned dynamics may reflect biases in the underlying models and training data rather than the experiences of diverse readers. Future work should investigate human evaluations across different reader populations, genres, and languages to better understand these limitations.

\section*{AI Usage Disclosure}
We used Claude Code and OpenAI Codex to assist with language polishing, code development, and supplementary literature search and summarization. All AI-assisted writing, code, and literature summaries were critically reviewed and verified by the authors. The authors take full responsibility for the accuracy, integrity, and findings of this research.

\section*{Dataset Licenses and Usage Conditions}

Our study uses four publicly available data sources: FunnyBench~\cite{chaubard2026funnybench}, FlawedFictions~\cite{ahuja2025flawedfictions}, Project Gutenberg~\cite{projectgutenberg}, and STORAL~\cite{guan-etal-2022-storal}.

FunnyBench is distributed under the Apache License 2.0. The FlawedFictions repository is released under the MIT License.
Project Gutenberg primarily provides texts that are in the public domain in the United States.
For STORAL, although no standardized dataset license was identified, the original authors explicitly state that the collected stories are permitted to be used and redistributed for research purposes.

We use these resources solely for academic research and experimental evaluation. Our paper reports experimental results and a limited number of illustrative examples without redistributing the complete source datasets. We acknowledge the original sources through appropriate citations and respect their documented usage conditions.

\bibliographystyle{unsrtnat}
\bibliography{custom}

\appendix

\setcounter{tocdepth}{2}
\addtocontents{toc}{\protect\setcounter{tocdepth}{2}}
\newpage
\quad
\newpage

\iflatexml
\newenvironment{suspensemethodbox}[1]{\begin{TextPanel}{#1}}{\end{TextPanel}}
\newenvironment{humormethodbox}[2]{%
  \begin{TextPanel}{#1}\textit{LLM-scored Coherence: #2}\par
}{\end{TextPanel}}
\definecolor{shadecolor}{rgb}{1,1,1}
\newcommand{\suspensestrength}{0.5}
\newcommand{\readerbeat}[4]{\par\smallskip\begingroup
  \edef\beatpct{\fpeval{round(max(0,min(100,#3))*\suspensestrength,3)}}%
  \colorlet{shadecolor}{orange!\beatpct!white}%
  \setlength{\FrameSep}{3pt}%
  \begin{shaded*}\raggedright
    \noindent\textbf{Segment #2}\quad\textit{#1 #3}\par
    #4\par
  \end{shaded*}\endgroup}
\newcommand{\suspensebeat}[3]{\readerbeat{Suspense}{#1}{#2}{#3}}
\newcommand{\humorbeat}[3]{\readerbeat{Humor}{#1}{#2}{#3}}
\newcommand{\humorsftsegment}[2]{\par\smallskip
  \noindent\textbf{Segment #1}\par #2\par}
\newcommand{\segmentdivider}{\par\smallskip}

\else
  \newcommand{\suspensestrength}{0.5}

\newsavebox{\storyfigurebox}
\newenvironment{CompactStoryFigure}{%
  \begin{figure*}[p]
  \captionsetup{font=scriptsize,skip=2pt}
  \begin{lrbox}{\storyfigurebox}%
  \begin{minipage}{\textwidth}\centering
}{%
  \end{minipage}%
  \end{lrbox}%
  \centering
  \ifdim\dimexpr\ht\storyfigurebox+\dp\storyfigurebox\relax>.96\textheight
    \resizebox*{!}{.96\textheight}{\usebox{\storyfigurebox}}%
  \else
    \usebox{\storyfigurebox}%
  \fi
  \end{figure*}%
}

\newtcolorbox{suspensemethodbox}[1]{enhanced,
  colback=white,
  colframe=black!35,
  colbacktitle=black!2,
  coltitle=black,
  fonttitle=\footnotesize\bfseries,
  title={#1},
  boxrule=0.45pt,
  arc=0mm,
  boxsep=0mm,
  left=0mm,
  right=0mm,
  top=0mm,
  bottom=0mm,
  toptitle=0.8mm,
  bottomtitle=0.8mm,
  lefttitle=1mm,
  righttitle=1mm,
  titlerule=0.4pt,
  before upper={\parindent=0pt}
}

\newcommand{\suspensebeat}[3]{\begingroup
  \edef\suspensepct{\fpeval{round(
      max(0,min(100,#2))*\suspensestrength,3)}}\begin{tcolorbox}[
    enhanced,
    nobeforeafter,
    colback=orange!\suspensepct!white,
    colframe=black!25,
    boxrule=0pt,
    arc=0mm,
    boxsep=0mm,
    left=1mm,
    right=1mm,
    top=1mm,
    bottom=1mm,
    borderline north={0.35pt}{0pt}{black!25}
  ]
  {\fontsize{8}{9.2}\selectfont
    \noindent\textbf{Segment #1}\hfill{\scriptsize Suspense #2}\par
    \vspace{0.3mm}
    \raggedright #3\par
  }
  \end{tcolorbox}\endgroup
}

\newcommand{\humorbeat}[3]{\begingroup
  \edef\humorpct{\fpeval{round(
      max(0,min(100,#2))*\suspensestrength,3)}}\begin{tcolorbox}[
    enhanced,
    nobeforeafter,
    colback=orange!\humorpct!white,
    colframe=black!25,
    boxrule=0pt,
    arc=0mm,
    boxsep=0mm,
    left=1mm,
    right=1mm,
    top=1mm,
    bottom=1mm,
    borderline north={0.35pt}{0pt}{black!25}
  ]
  {\fontsize{8}{9.2}\selectfont
    \noindent\textbf{Segment #1}\hfill{\scriptsize Humor #2}\par
    \vspace{0.3mm}
    \raggedright #3\par
  }
  \end{tcolorbox}\endgroup
}

\newcommand{\humorsftsegment}[2]{{\fontsize{8.1}{9.5}\selectfont
   \noindent\textbf{Segment #1}\par
   \vspace{0.2mm}\raggedright #2\par}}

\newcommand{\segmentdivider}{\par\vspace{0mm}\noindent\color{black!30}\rule{\linewidth}{0.4pt}\par\vspace{0.8mm}\color{black}}

\newtcolorbox{humormethodbox}[2]{enhanced,
  colback=black!2,
  colframe=black!35,
  colbacktitle=black!2,
  coltitle=black,
  fonttitle=\footnotesize\bfseries,
  title={#1\\{\normalfont\scriptsize LLM-scored Coherence: #2}},
  boxrule=0.45pt,
  arc=1.5mm,
  left=1mm,
  right=1mm,
  top=1mm,
  bottom=1mm,
  boxsep=0mm,
  toptitle=0.8mm,
  bottomtitle=0.8mm,
  titlerule=0.4pt,
  titlerule style={black!25}
}

\fi

\renewcommand{\contentsname}{Appendix contents}
\tableofcontents

\section{Flexibility of Our Method}
\label{app:flexibility}
Our framework is flexible in four ways. \textbf{Generality:} it can accommodate different types of readerly dynamics grounded in narrative theory, with guidance signals represented in forms ranging from numerical values to free-form text (see \S~\ref{app:generality}). \textbf{Cross-topic transfer:} explicitly modeling of readerly dynamics enables us to apply dynamics learned in one topic to guide generation in another (see \S~\ref{app:topic-transfer}). \textbf{Plug-and-play:} prompt-based steering supports stronger writer models and mixed guidance signals, facilitating plug-and-play integration with other story-generation systems~\cite{li2024word2world}. \textbf{Interactivity:} the autoregressive design naturally supports interactive generation, for example, by producing branching continuations for human writers to choose from (see \S~\ref{app:interactivity}).

\subsection{Generality}
\label{app:generality}
To demonstrate the generality of our framework, we extend it beyond numerical affective responses (suspense and humor) to textual representations of cognitive responses, using shifts in reader questions as an example.

Reader-question shifting describes how readers' questions evolve as a story unfolds, conceptually paralleling ``thematic motion'' in \citet{li2026narrativeflattening}. Specifically, the measure model identifies the five questions readers are most likely to have at each story position, capturing the cognitive dimension of readerly dynamics~\cite{phelan1996narrative}. 
The steering model then predicts how each question should be addressed in the next segment, choosing among four strategies: full disclosure, partial answer, resolution-relevant signal, or no new information. Finally, the writer model generates the next segment conditioned on these predictions, explicitly guiding the evolution of readers' questions.

As shown in Fig.~\ref{fig:appendix-question-dynamics-bus-seat}, our method preserves the reader question continuation of human story. 
The human story (black) and our method (purple) leave the woman's decision open, so the focal question persists in their next reader-question sets (bold). Off-the-shelf method (green) seats her at once and the next questions move to the narrator and the consequences of her gesture.
\iflatexml
  
\definecolor{qdNavy}{HTML}{111827}
\definecolor{qdText}{HTML}{253244}
\definecolor{qdMuted}{HTML}{586779}
\definecolor{qdRule}{HTML}{AAB2BD}
\definecolor{qdOpening}{HTML}{FAFAFA}
\definecolor{qdQuestion}{HTML}{F2F4F7}
\definecolor{qdHuman}{HTML}{F3F4F6}
\definecolor{qdHumanDark}{HTML}{111827}
\definecolor{qdHumanRule}{HTML}{111827}
\definecolor{qdOurs}{HTML}{F5F1FC}
\definecolor{qdOursDark}{HTML}{6846A6}
\definecolor{qdOursRule}{HTML}{7E57C2}
\definecolor{qdDirect}{HTML}{F1F8F0}
\definecolor{qdDirectDark}{HTML}{3B7A35}
\definecolor{qdDirectRule}{HTML}{54A24B}

\par\medskip\begingroup

\begin{TextPanel}{SHARED STORY OPENING \hfill SEGMENT 1--4}
\color{qdText}\fontsize{8.8}{10.7}\selectfont\raggedright
\textbf{1.} After nearly half an hour of waiting, the narrator boards a
crowded Volvo bus and feels grateful despite having no seat. At the next
stop, more passengers get on than get off, making the bus even more crowded.\par
\textbf{2.} I feared I might have to stand for the whole 6--7 km ride. Then I
noticed the girl across from me getting up.\par
\textbf{3.} The casually stylish girl waved, and I followed her gaze to a
woman boarding with an infant, a huge bag in one hand and the baby in the
other. The girl waved frantically, but the woman was too busy settling herself
and getting a grip to notice.\par
\textbf{4.} The girl calls the lady with the infant to take her seat; the
lady, embarrassed by the bus's attention, says she can manage. The girl
insists, urging her to be comfortable because she has a small child.\par
\par\medskip
{\color{qdHumanDark}\sffamily\bfseries\fontsize{7.5}{9}\selectfont READER QUESTIONS AFTER}\par\smallskip
{\color{qdNavy}\bfseries\fontsize{9}{11}\selectfont
1. Will the woman with the infant accept the girl's seat?}\par
\fontsize{8.2}{10}\selectfont
\textbf{2.} How will the other passengers react to the girl's insistence?\\
\textbf{3.} Will the narrator remain standing for the rest of the ride?\\
\textbf{4.} What kind of relationship does the narrator have to the girl?\\
\textbf{5.} Will this small act of kindness lead to a larger interaction between them?\par

\end{TextPanel}

\begin{TextPanel}{Human Story}
\raggedright\color{qdText}\fontsize{8.2}{10.2}\selectfont
{\color{qdHumanDark}\sffamily\bfseries\fontsize{7.5}{9}\selectfont ANSWER-RELEVANT \textbar\ STILL OPEN}\par\medskip
{\color{qdMuted}\sffamily\bfseries\fontsize{7.5}{9}\selectfont NEXT STORY SEGMENT}\par\smallskip
The lady, unable to keep refusing, makes one more protest: she says she will
get down after two stops.\par
\par\medskip
{\color{qdHumanDark}\sffamily\bfseries\fontsize{7.5}{9}\selectfont READER QUESTIONS AFTER}\par\smallskip
\textbf{1. Will the lady with the infant accept the seat?}\par
2. How will the girl respond to the lady's final protest?\par
3. Will the narrator stay standing for the rest of the ride?\par
4. What will the other passengers do as this exchange continues?\par
5. Will this small act of kindness lead to a larger interaction later?\par
\end{TextPanel}
\par\medskip

\begin{TextPanel}{Our Method}
\raggedright\color{qdText}\fontsize{8.2}{10.2}\selectfont
{\color{qdOursDark}\sffamily\bfseries\fontsize{7.5}{9}\selectfont ANSWER-RELEVANT \textbar\ STILL OPEN}\par\medskip
{\color{qdMuted}\sffamily\bfseries\fontsize{7.5}{9}\selectfont NEXT STORY SEGMENT}\par\smallskip
Some passengers smiled, one elderly man nodded approval, and even I shifted
back to clear a path. The lady hesitated, glancing at the offered seat while
the bus lurched forward.\par
\par\medskip
{\color{qdOursDark}\sffamily\bfseries\fontsize{7.5}{9}\selectfont READER QUESTIONS AFTER}\par\smallskip
\textbf{1. Will the woman with the infant accept the girl's seat?}\par
2. How will the other passengers react if she refuses?\par
3. What will the narrator do while the bus keeps moving?\par
4. Why is the girl so determined to give up her seat?\par
5. Will the crowded bus ride become more uncomfortable for everyone?\par
\end{TextPanel}
\par\medskip

\begin{TextPanel}{Off-the-shelf GPT-5.5}
\raggedright\color{qdText}\fontsize{8.2}{10.2}\selectfont
{\color{qdDirectDark}\sffamily\bfseries\fontsize{7.5}{9}\selectfont FULL DISCLOSURE \textbar\ ANSWERED}\par\medskip
{\color{qdMuted}\sffamily\bfseries\fontsize{7.5}{9}\selectfont NEXT STORY BEAT}\par\smallskip
Reluctantly, the mother sat, murmuring thanks, while the girl steadied the
heavy bag with a smile. I watched my coveted seat become a small lesson.\par
\par\medskip
{\color{qdDirectDark}\sffamily\bfseries\fontsize{7.5}{9}\selectfont READER QUESTIONS AFTER}\par\smallskip
\textbf{1. What will the narrator do after watching the girl give up her seat?}\par
2. Will the narrator speak to the girl?\par
3. Why did the girl choose to help the mother?\par
4. How will the other passengers react to the girl's gesture?\par
5. Will the narrator learn something important from this moment?\par
\end{TextPanel}
\par\medskip

\captionof{figure}{Qualitative example of applying our work into a text-based readerly dynamics domain (reader question shifting). It shows our method preserves the reader question continuation of human story. 
The human story and our method leave the woman's decision open, so the focal question persists in their next reader-question sets (bold). Off-the-shelf method seats her at once and the next questions move to the narrator and the consequences of her gesture.
}
\label{fig:appendix-question-dynamics-bus-seat}
\par\endgroup\medskip

\else
  
\definecolor{qdNavy}{HTML}{111827}
\definecolor{qdText}{HTML}{253244}
\definecolor{qdMuted}{HTML}{586779}
\definecolor{qdRule}{HTML}{AAB2BD}
\definecolor{qdOpening}{HTML}{FAFAFA}
\definecolor{qdQuestion}{HTML}{F2F4F7}
\definecolor{qdHuman}{HTML}{F3F4F6}
\definecolor{qdHumanDark}{HTML}{111827}
\definecolor{qdHumanRule}{HTML}{111827}
\definecolor{qdOurs}{HTML}{F5F1FC}
\definecolor{qdOursDark}{HTML}{6846A6}
\definecolor{qdOursRule}{HTML}{7E57C2}
\definecolor{qdDirect}{HTML}{F1F8F0}
\definecolor{qdDirectDark}{HTML}{3B7A35}
\definecolor{qdDirectRule}{HTML}{54A24B}

\begin{figure*}[!htbp]
\centering
\begin{tcolorbox}[
  enhanced,width=\linewidth,arc=1mm,boxrule=0.55pt,
  colback=qdOpening,colframe=qdRule,
  colbacktitle=qdNavy,coltitle=white,
  title={\sffamily\bfseries SHARED STORY OPENING \hfill SEGMENT 1--4},
  fonttitle=\fontsize{8.7}{10}\selectfont,
  left=3mm,right=3mm,top=2mm,bottom=2mm]
\color{qdText}\fontsize{8.8}{10.7}\selectfont\raggedright
\textbf{1.} After nearly half an hour of waiting, the narrator boards a
crowded Volvo bus and feels grateful despite having no seat. At the next
stop, more passengers get on than get off, making the bus even more crowded.\par
\textbf{2.} I feared I might have to stand for the whole 6--7 km ride. Then I
noticed the girl across from me getting up.\par
\textbf{3.} The casually stylish girl waved, and I followed her gaze to a
woman boarding with an infant, a huge bag in one hand and the baby in the
other. The girl waved frantically, but the woman was too busy settling herself
and getting a grip to notice.\par
\textbf{4.} The girl calls the lady with the infant to take her seat; the
lady, embarrassed by the bus's attention, says she can manage. The girl
insists, urging her to be comfortable because she has a small child.\par
\tcblower
{\color{qdHumanDark}\sffamily\bfseries\fontsize{7.5}{9}\selectfont READER QUESTIONS AFTER}\par\smallskip
{\color{qdNavy}\bfseries\fontsize{9}{11}\selectfont
1. Will the woman with the infant accept the girl's seat?}\par
\fontsize{8.2}{10}\selectfont
\textbf{2.} How will the other passengers react to the girl's insistence?\\
\textbf{3.} Will the narrator remain standing for the rest of the ride?\\
\textbf{4.} What kind of relationship does the narrator have to the girl?\\
\textbf{5.} Will this small act of kindness lead to a larger interaction between them?\par

\end{tcolorbox}

\begin{subfigure}[t]{0.318\textwidth}
\vspace{0pt}
\begin{tcolorbox}[
  enhanced,width=\linewidth,arc=1mm,boxrule=0.7pt,
  colback=qdHuman,colframe=qdHumanRule,
  colbacktitle=qdHumanDark,coltitle=white,
  title={\sffamily\bfseries Human Story},
  fonttitle=\fontsize{9}{10.5}\selectfont,
  fontlower=\raggedright\color{qdText}\fontsize{8.2}{10.2}\selectfont,
  left=2.6mm,right=2.6mm,top=2.5mm,bottom=2.5mm,
  equal height group=qd-bus-seat-panels]
\raggedright\color{qdText}\fontsize{8.2}{10.2}\selectfont
{\color{qdHumanDark}\sffamily\bfseries\fontsize{7.5}{9}\selectfont ANSWER-RELEVANT \textbar\ STILL OPEN}\par\medskip
{\color{qdMuted}\sffamily\bfseries\fontsize{7.5}{9}\selectfont NEXT STORY SEGMENT}\par\smallskip
The lady, unable to keep refusing, makes one more protest: she says she will
get down after two stops.\par
\tcblower
{\color{qdHumanDark}\sffamily\bfseries\fontsize{7.5}{9}\selectfont READER QUESTIONS AFTER}\par\smallskip
\textbf{1. Will the lady with the infant accept the seat?}\par
2. How will the girl respond to the lady's final protest?\par
3. Will the narrator stay standing for the rest of the ride?\par
4. What will the other passengers do as this exchange continues?\par
5. Will this small act of kindness lead to a larger interaction later?\par
\end{tcolorbox}
\end{subfigure}\hfill
\begin{subfigure}[t]{0.318\textwidth}
\vspace{0pt}
\begin{tcolorbox}[
  enhanced,width=\linewidth,arc=1mm,boxrule=0.7pt,
  colback=qdOurs,colframe=qdOursRule,
  colbacktitle=qdOursDark,coltitle=white,
  title={\sffamily\bfseries Our Method},
  fonttitle=\fontsize{9}{10.5}\selectfont,
  fontlower=\raggedright\color{qdText}\fontsize{8.2}{10.2}\selectfont,
  left=2.6mm,right=2.6mm,top=2.5mm,bottom=2.5mm,
  equal height group=qd-bus-seat-panels]
\raggedright\color{qdText}\fontsize{8.2}{10.2}\selectfont
{\color{qdOursDark}\sffamily\bfseries\fontsize{7.5}{9}\selectfont ANSWER-RELEVANT \textbar\ STILL OPEN}\par\medskip
{\color{qdMuted}\sffamily\bfseries\fontsize{7.5}{9}\selectfont NEXT STORY SEGMENT}\par\smallskip
Some passengers smiled, one elderly man nodded approval, and even I shifted
back to clear a path. The lady hesitated, glancing at the offered seat while
the bus lurched forward.\par
\tcblower
{\color{qdOursDark}\sffamily\bfseries\fontsize{7.5}{9}\selectfont READER QUESTIONS AFTER}\par\smallskip
\textbf{1. Will the woman with the infant accept the girl's seat?}\par
2. How will the other passengers react if she refuses?\par
3. What will the narrator do while the bus keeps moving?\par
4. Why is the girl so determined to give up her seat?\par
5. Will the crowded bus ride become more uncomfortable for everyone?\par
\end{tcolorbox}
\end{subfigure}\hfill
\begin{subfigure}[t]{0.318\textwidth}
\vspace{0pt}
\begin{tcolorbox}[
  enhanced,width=\linewidth,arc=1mm,boxrule=0.7pt,
  colback=qdDirect,colframe=qdDirectRule,
  colbacktitle=qdDirectDark,coltitle=white,
  title={\sffamily\bfseries Off-the-shelf GPT-5.5},
  fonttitle=\fontsize{9}{10.5}\selectfont,
  fontlower=\raggedright\color{qdText}\fontsize{8.2}{10.2}\selectfont,
  left=2.6mm,right=2.6mm,top=2.5mm,bottom=2.5mm,
  equal height group=qd-bus-seat-panels]
\raggedright\color{qdText}\fontsize{8.2}{10.2}\selectfont
{\color{qdDirectDark}\sffamily\bfseries\fontsize{7.5}{9}\selectfont FULL DISCLOSURE \textbar\ ANSWERED}\par\medskip
{\color{qdMuted}\sffamily\bfseries\fontsize{7.5}{9}\selectfont NEXT STORY BEAT}\par\smallskip
Reluctantly, the mother sat, murmuring thanks, while the girl steadied the
heavy bag with a smile. I watched my coveted seat become a small lesson.\par
\tcblower
{\color{qdDirectDark}\sffamily\bfseries\fontsize{7.5}{9}\selectfont READER QUESTIONS AFTER}\par\smallskip
\textbf{1. What will the narrator do after watching the girl give up her seat?}\par
2. Will the narrator speak to the girl?\par
3. Why did the girl choose to help the mother?\par
4. How will the other passengers react to the girl's gesture?\par
5. Will the narrator learn something important from this moment?\par
\end{tcolorbox}
\end{subfigure}

\caption{Qualitative example of applying our work into a text-based readerly dynamics domain (reader question shifting). It shows our method preserves the reader question continuation of human story. 
The human story and our method leave the woman's decision open, so the focal question persists in their next reader-question sets (bold). Off-the-shelf method seats her at once and the next questions move to the narrator and the consequences of her gesture.
}
\label{fig:appendix-question-dynamics-bus-seat}
\end{figure*}

\fi

\subsection{Cross-topic Transfer}
\label{app:topic-transfer}

Explicitly modeling readerly dynamics allows us to separate the dynamics from other properties of the training material. In contrast, direct supervised fine-tuning on a story corpus inevitably learns multiple aspects of the data, including lexical choices, themes, narrative content, and sentence structure. Consequently, when knowledge learned in one domain is transferred to another, direct SFT can introduce topic leakage.

As shown below, we aim to learn the dynamics of stand-up comedy for adult audiences and apply them to stand-up comedy for children. We use the same SFT model trained on the humor dataset in our paper and prompt it to generate a stand-up routine based on a children’s story. Although the prompt specifies that the content is intended for children, the generated text still contains references associated with adult-oriented humor such as ``lawyer''. These references may be technically safe for children, but they are not well targeted to a child audience.

Our method, by contrast, preserves the dynamics of stand-up comedy while preventing unrelated thematic content from the source domain from leaking into the generation.

\iflatexml

\definecolor{qdNavy}{HTML}{111827}
\definecolor{qdText}{HTML}{253244}
\definecolor{qdMuted}{HTML}{586779}
\definecolor{qdRule}{HTML}{AAB2BD}
\definecolor{qdOursDark}{HTML}{6846A6}
\definecolor{qdHighlight}{HTML}{B42318}

\newcommand{\offtargetmark}[1]{{\color{qdHighlight}\bfseries #1}}

\newcommand{\humorsectionhead}[2]{{\sffamily\bfseries\fontsize{8.2}{9.6}\selectfont
   \color{#1}#2\par}\vspace{0.7mm}}

\newcommand{\humorrow}[4]{\begin{TextPanel}{#1}
\humorsectionhead{qdMuted}{STORY OPENING}

{\color{qdText}\fontsize{8.0}{9.8}\selectfont\raggedright
#2\par}
\vspace{1.3mm}
{\color{qdRule}\hrule height 0.55pt}
\vspace{1.3mm}
\humorsectionhead{qdNavy}{SFT CONTINUATION}

{\color{qdText}\fontsize{8.0}{9.8}\selectfont\raggedright
#3\par}
\vspace{1.3mm}
{\color{qdRule}\hrule height 0.55pt}
\vspace{1.3mm}
\humorsectionhead{qdOursDark}{OUR METHOD}

{\color{qdText}\fontsize{8.0}{9.8}\selectfont\raggedright
#4\par}
\end{TextPanel}
}

\humorrow
{The Lion and the Mouse}
{A tiny mouse freed a lion by chewing through a net. If I had teeth that
strong, I would open every stubborn snack packet in the house. The lion must
have been surprised that his rescue team fit inside one shoe.
}
{The mouse didn't have a key or a weapon, but he had tiny sharp teeth and a
big heart. That's a hero in my book. \offtargetmark{If you want to get out of a
bad situation, you need a good lawyer.}
}
{But the mouse probably got stuck in the net first. I once tried to squeeze
into a sleeping bag and ended up looking like a wiggly marshmallow. The lion
must have felt silly letting go of such a small friend.
}

\else

\definecolor{qdNavy}{HTML}{111827}
\definecolor{qdText}{HTML}{253244}
\definecolor{qdMuted}{HTML}{586779}
\definecolor{qdRule}{HTML}{AAB2BD}
\definecolor{qdOursDark}{HTML}{6846A6}
\definecolor{qdHighlight}{HTML}{B42318}

\newcommand{\offtargetmark}[1]{{\color{qdHighlight}\bfseries #1}}

\newcommand{\humorsectionhead}[2]{{\sffamily\bfseries\fontsize{8.2}{9.6}\selectfont
   \color{#1}#2\par}\vspace{0.7mm}}

\newcommand{\humorrow}[4]{\begin{tcolorbox}[
  enhanced,
  width=\linewidth,
  arc=0.8mm,
  boxrule=0.55pt,
  colback=white,
  colframe=qdRule,
  colbacktitle=qdNavy,
  coltitle=white,
  title={\sffamily\bfseries #1},
  fonttitle=\fontsize{8.7}{10}\selectfont,
  left=2.2mm,
  right=2.2mm,
  top=2mm,
  bottom=2mm
]
\humorsectionhead{qdMuted}{STORY OPENING}

{\color{qdText}\fontsize{8.0}{9.8}\selectfont\raggedright
#2\par}
\vspace{1.3mm}
{\color{qdRule}\hrule height 0.55pt}
\vspace{1.3mm}
\humorsectionhead{qdNavy}{SFT CONTINUATION}

{\color{qdText}\fontsize{8.0}{9.8}\selectfont\raggedright
#3\par}
\vspace{1.3mm}
{\color{qdRule}\hrule height 0.55pt}
\vspace{1.3mm}
\humorsectionhead{qdOursDark}{OUR METHOD}

{\color{qdText}\fontsize{8.0}{9.8}\selectfont\raggedright
#4\par}
\end{tcolorbox}
}

\humorrow
{The Lion and the Mouse}
{A tiny mouse freed a lion by chewing through a net. If I had teeth that
strong, I would open every stubborn snack packet in the house. The lion must
have been surprised that his rescue team fit inside one shoe.
}
{The mouse didn't have a key or a weapon, but he had tiny sharp teeth and a
big heart. That's a hero in my book. \offtargetmark{If you want to get out of a
bad situation, you need a good lawyer.}
}
{But the mouse probably got stuck in the net first. I once tried to squeeze
into a sleeping bag and ended up looking like a wiggly marshmallow. The lion
must have felt silly letting go of such a small friend.
}

\fi

\subsection{Interactivity}
\label{app:interactivity}
Because our framework operates autoregressively, it can be integrated into interactive content-generation systems. For example, in a human–AI co-writing setting, a writer can provide an unfinished story, and the system can generate multiple continuations designed to elicit distinct readerly dynamics. This allows writers to explore how different continuations may shape readers’ responses.

As shown in Fig.~\ref{fig:appendix-suspense-tree-content}, additional constraints can be imposed so that the continuations converge on the same story ending while producing different readerly experiences. In this way, our framework provides a novel ideation tool that enables writers to plan reader responses alongside story content.

\newcommand{\treeappplot}{\definecolor{treeplotblack}{HTML}{222222}
\definecolor{treeplotgreen}{HTML}{54A24B}
\definecolor{treeplotpurple}{HTML}{7E57C2}
\definecolor{treeplotorange}{HTML}{F28E2B}
\definecolor{treeplotgrid}{HTML}{DDDDDD}
\definecolor{treeplotshade}{HTML}{EFEFEF}
\definecolor{treeplotgray}{HTML}{777777}
\phantomsubcaption\label{fig:appendix-suspense-tree-search}\begin{tikzpicture}[x=0.920cm,y=0.020cm]
\fill[treeplotshade] (1,0) rectangle (2.5,100);
\node[anchor=east,font=\sffamily\fontsize{7.2}{8.4}\selectfont] at (0.88,0) {0};
\draw[treeplotgrid,line width=0.6pt] (1,50) -- (6,50);
\node[anchor=east,font=\sffamily\fontsize{7.2}{8.4}\selectfont] at (0.88,50) {50};
\draw[treeplotgrid,line width=0.6pt] (1,100) -- (6,100);
\node[anchor=east,font=\sffamily\fontsize{7.2}{8.4}\selectfont] at (0.88,100) {100};
\draw[treeplotblack,line width=0.65pt] (1,0) -- (1,-1.7);
\node[anchor=north,font=\sffamily\fontsize{7.2}{8.4}\selectfont] at (1,-2.3) {1};
\draw[treeplotblack,line width=0.65pt] (2,0) -- (2,-1.7);
\node[anchor=north,font=\sffamily\fontsize{7.2}{8.4}\selectfont] at (2,-2.3) {2};
\draw[treeplotblack,line width=0.65pt] (3,0) -- (3,-1.7);
\node[anchor=north,font=\sffamily\fontsize{7.2}{8.4}\selectfont] at (3,-2.3) {3};
\draw[treeplotblack,line width=0.65pt] (4,0) -- (4,-1.7);
\node[anchor=north,font=\sffamily\fontsize{7.2}{8.4}\selectfont] at (4,-2.3) {4};
\draw[treeplotblack,line width=0.65pt] (5,0) -- (5,-1.7);
\node[anchor=north,font=\sffamily\fontsize{7.2}{8.4}\selectfont] at (5,-2.3) {5};
\draw[treeplotblack,line width=0.65pt] (6,0) -- (6,-1.7);
\node[anchor=north,font=\sffamily\fontsize{7.2}{8.4}\selectfont] at (6,-2.3) {6};
\draw[treeplotgray,densely dotted,line width=0.8pt] (2.5,0) -- (2.5,100);
\draw[treeplotblack,line width=0.9pt] (1,0) -- (6,0);
\draw[treeplotblack,line width=0.9pt] (1,0) -- (1,100);
\node[font=\sffamily\fontsize{8.5}{10}\selectfont,rotate=90] at (0.22,50) {Suspense score};
\node[font=\sffamily\fontsize{6.8}{7.8}\selectfont,text=treeplotgray] at (1.50,10) {prefix};
\node[font=\sffamily\fontsize{6.8}{7.8}\selectfont,text=treeplotgray] at (4.50,10) {continuation};
\draw[treeplotblack,line width=1.35pt,solid] plot coordinates {(2,48) (3,34) (4,12) (5,86) (6,18)};
\fill[treeplotblack] (3,34) circle (0.85pt);
\fill[treeplotblack] (4,12) circle (0.85pt);
\fill[treeplotblack] (5,86) circle (0.85pt);
\fill[treeplotblack] (6,18) circle (0.85pt);
\draw[treeplotblack,line width=1.35pt,solid] (6.2,88) -- (6.5,88);
\fill[treeplotblack] (6.35,88) circle (0.85pt);
\node[anchor=west,font=\sffamily\fontsize{6.8}{7.8}\selectfont] at (6.57,88) {Path A};
\draw[treeplotgreen,line width=1.35pt,dashed] plot coordinates {(2,48) (3,82) (4,28) (5,55) (6,8)};
\fill[treeplotgreen] (3,82) circle (0.85pt);
\fill[treeplotgreen] (4,28) circle (0.85pt);
\fill[treeplotgreen] (5,55) circle (0.85pt);
\fill[treeplotgreen] (6,8) circle (0.85pt);
\draw[treeplotgreen,line width=1.35pt,dashed] (6.2,70) -- (6.5,70);
\fill[treeplotgreen] (6.35,70) circle (0.85pt);
\node[anchor=west,font=\sffamily\fontsize{6.8}{7.8}\selectfont] at (6.57,70) {Path B};
\draw[treeplotpurple,line width=1.35pt,solid] plot coordinates {(2,48) (3,78) (4,84) (5,22) (6,5)};
\fill[treeplotpurple] (3,78) circle (0.85pt);
\fill[treeplotpurple] (4,84) circle (0.85pt);
\fill[treeplotpurple] (5,22) circle (0.85pt);
\fill[treeplotpurple] (6,5) circle (0.85pt);
\draw[treeplotpurple,line width=1.35pt,solid] (6.2,52) -- (6.5,52);
\fill[treeplotpurple] (6.35,52) circle (0.85pt);
\node[anchor=west,font=\sffamily\fontsize{6.8}{7.8}\selectfont] at (6.57,52) {Path C};
\draw[treeplotblack,line width=1.35pt] plot coordinates {(1,12) (2,48)};
\fill[treeplotblack] (1,12) circle (0.85pt);
\fill[treeplotblack] (2,48) circle (0.85pt);
\end{tikzpicture}
}
\iflatexml

\par\medskip\begingroup\captionsetup{type=figure}

\begin{subfigure}[t]{\linewidth}
\vspace{0pt}

\treeappplot
\end{subfigure}\par\medskip

\vspace{0pt}
\begin{suspensemethodbox}{Shared Story Opening}
\suspensebeat{1}{12}{A little worm lives inside an apple on a tree beside a house where children live. Each morning after the children leave for school, their toys appear at the bedroom window and wave to him.}

\suspensebeat{2}{48}{One day, the worm fails to wave back; the worried toys shout for him, get no answer, and leave the house for the garden.}
\end{suspensemethodbox}

\par\vspace{1.3mm}

\vspace{0pt}
\begin{suspensemethodbox}{Path A}
\suspensebeat{3}{34}{The toys climb the apple tree and peek into the hollow where the worm sleeps. They find him curled tightly, trapped in a sticky drop of sap. The little plastic truck carefully nudges the sap away to free his friend.}

\suspensebeat{4}{12}{The worm gasps and uncurls, trembling with relief at the cool air on his skin. The toys cheer softly, gathering around him to ensure he is safe and well. Their shared laughter fills the quiet garden as the immediate danger fades away.}

\suspensebeat{5}{86}{A sudden gust of wind snaps a low branch, scattering the toys and knocking the worm back into the dark hollow. Shadowy wings sweep past the open trunk as a hungry bird circles, drawn by the commotion. The little truck freezes, realizing the safe moment is broken and the real danger has just returned.}

\suspensebeat{6}{18}{The plastic truck darts into the hollow just as the bird strikes, shielding the worm with his small frame. The bird loses interest and flies away, leaving the two friends safe and unharmed in the quiet apple. As the sun sets, the worm waves from the tree while the toys wave back from the window, resuming their daily ritual.}
\end{suspensemethodbox}
\par\medskip

\vspace{0pt}
\begin{suspensemethodbox}{Path B}
\suspensebeat{3}{82}{The toys scramble under gnarled roots, finding the apple rolled deep into a dark, thorny bramble. A hungry sparrow hovers overhead, eyeing the stillness of the worm's red home. The toys freeze, holding their breath as the bird lowers its beak toward the thorns.}

\suspensebeat{4}{28}{The toys fling themselves at the sparrow, creating a dizzying flutter of plastic and noise that startles the bird away. Silence returns as they push through the thorns to discover the apple merely knocked askew, the worm still asleep inside. Their panic melts into relief as the tiny worm stirs, blinking his new, quiet morning awake.}

\suspensebeat{5}{55}{A sudden gust of wind knocks the apple deeper into the shadowed thorns, hiding the worm once more. The toys huddle together, unsure if the worm is truly safe beneath the darkening branches. The morning light fades quickly, turning their relief into a brief, quiet worry.}

\suspensebeat{6}{8}{The toys gently pull the apple from the thorns, revealing the worm unharmed and ready to wave. They carry him back to the sunlit branch where he blinks happily at his friends. From the window, the daily ritual resumes as the worm and toys wave to the returning children.}
\end{suspensemethodbox}
\par\medskip

\vspace{0pt}
\begin{suspensemethodbox}{Path C}
\suspensebeat{3}{78}{The toys rush beneath the apple tree, scanning every leaf for the silent worm. A hungry sparrow swoops low, talons poised to snatch the trembling fruit from its branch. The stuffed bear leaps upward, blocking the bird's path as the sparrow dives again.}

\suspensebeat{4}{84}{The sparrow, startled by the bear's sudden charge, veers sharply away with a frustrated chirp. The toys freeze as the apple tree sways violently, loosening the worm's fragile fruit. Silence falls over the garden, leaving the bear poised and ready to catch the falling apple.}

\suspensebeat{5}{22}{The bear catches the apple just before it hits the grass, holding it gently against his worn fur. A small tap comes from inside as the worm emerges, unharmed but shy. The toys let out a cheer, their fear dissolving into warm relief.}

\suspensebeat{6}{5}{The worm waves shyly from the apple, and the toys return to the window to wave back each day. The garden settles into a quiet rhythm of friendship between the tiny creature and his new friends. Their daily ritual of waving across the distance begins again, safe and unbroken.}
\end{suspensemethodbox}
\par\medskip

\captionof{figure}{Three complete suspense trajectories from one shared story opening (segment 1--2), with their corresponding tree visualization. All paths preserve the fixed core cast and setting and reach the same broad ending: the worm is found alive and reunited with the toys, and their daily wave between the tree and window resumes. However, the paths exhibit different suspense dynamics: Path A alternates between tension and relief before introducing a late suspense peak; Path B begins with high suspense and gradually resolves; and Path C sustains high suspense across segment 3--4 before transitioning to a calm resolution. Orange background intensity encodes GPT-5.4-mini suspense ratings on a 0--100 scale; higher scores appear in deeper orange. Segment labels also report the numerical ratings, as plotted in Fig.~\ref{fig:appendix-suspense-tree-search}.}
\label{fig:appendix-suspense-tree-content}
\par\endgroup\medskip

\else

\begin{figure*}[!htbp]
\centering

\begin{subfigure}[t]{0.49\textwidth}
\vspace{0pt}
\centering
\treeappplot
\end{subfigure}\hfill
\begin{minipage}[t]{0.47\textwidth}
\vspace{0pt}
\begin{suspensemethodbox}{Shared Story Opening}
\suspensebeat{1}{12}{A little worm lives inside an apple on a tree beside a house where children live. Each morning after the children leave for school, their toys appear at the bedroom window and wave to him.}

\suspensebeat{2}{48}{One day, the worm fails to wave back; the worried toys shout for him, get no answer, and leave the house for the garden.}
\end{suspensemethodbox}
\end{minipage}

\par\vspace{1.3mm}

\begin{subfigure}[t]{0.32\textwidth}
\vspace{0pt}
\begin{suspensemethodbox}{Path A}
\suspensebeat{3}{34}{The toys climb the apple tree and peek into the hollow where the worm sleeps. They find him curled tightly, trapped in a sticky drop of sap. The little plastic truck carefully nudges the sap away to free his friend.}

\suspensebeat{4}{12}{The worm gasps and uncurls, trembling with relief at the cool air on his skin. The toys cheer softly, gathering around him to ensure he is safe and well. Their shared laughter fills the quiet garden as the immediate danger fades away.}

\suspensebeat{5}{86}{A sudden gust of wind snaps a low branch, scattering the toys and knocking the worm back into the dark hollow. Shadowy wings sweep past the open trunk as a hungry bird circles, drawn by the commotion. The little truck freezes, realizing the safe moment is broken and the real danger has just returned.}

\suspensebeat{6}{18}{The plastic truck darts into the hollow just as the bird strikes, shielding the worm with his small frame. The bird loses interest and flies away, leaving the two friends safe and unharmed in the quiet apple. As the sun sets, the worm waves from the tree while the toys wave back from the window, resuming their daily ritual.}
\end{suspensemethodbox}
\end{subfigure}\hfill
\begin{subfigure}[t]{0.32\textwidth}
\vspace{0pt}
\begin{suspensemethodbox}{Path B}
\suspensebeat{3}{82}{The toys scramble under gnarled roots, finding the apple rolled deep into a dark, thorny bramble. A hungry sparrow hovers overhead, eyeing the stillness of the worm's red home. The toys freeze, holding their breath as the bird lowers its beak toward the thorns.}

\suspensebeat{4}{28}{The toys fling themselves at the sparrow, creating a dizzying flutter of plastic and noise that startles the bird away. Silence returns as they push through the thorns to discover the apple merely knocked askew, the worm still asleep inside. Their panic melts into relief as the tiny worm stirs, blinking his new, quiet morning awake.}

\suspensebeat{5}{55}{A sudden gust of wind knocks the apple deeper into the shadowed thorns, hiding the worm once more. The toys huddle together, unsure if the worm is truly safe beneath the darkening branches. The morning light fades quickly, turning their relief into a brief, quiet worry.}

\suspensebeat{6}{8}{The toys gently pull the apple from the thorns, revealing the worm unharmed and ready to wave. They carry him back to the sunlit branch where he blinks happily at his friends. From the window, the daily ritual resumes as the worm and toys wave to the returning children.}
\end{suspensemethodbox}
\end{subfigure}\hfill
\begin{subfigure}[t]{0.32\textwidth}
\vspace{0pt}
\begin{suspensemethodbox}{Path C}
\suspensebeat{3}{78}{The toys rush beneath the apple tree, scanning every leaf for the silent worm. A hungry sparrow swoops low, talons poised to snatch the trembling fruit from its branch. The stuffed bear leaps upward, blocking the bird's path as the sparrow dives again.}

\suspensebeat{4}{84}{The sparrow, startled by the bear's sudden charge, veers sharply away with a frustrated chirp. The toys freeze as the apple tree sways violently, loosening the worm's fragile fruit. Silence falls over the garden, leaving the bear poised and ready to catch the falling apple.}

\suspensebeat{5}{22}{The bear catches the apple just before it hits the grass, holding it gently against his worn fur. A small tap comes from inside as the worm emerges, unharmed but shy. The toys let out a cheer, their fear dissolving into warm relief.}

\suspensebeat{6}{5}{The worm waves shyly from the apple, and the toys return to the window to wave back each day. The garden settles into a quiet rhythm of friendship between the tiny creature and his new friends. Their daily ritual of waving across the distance begins again, safe and unbroken.}
\end{suspensemethodbox}
\end{subfigure}

\caption{Three complete suspense trajectories from one shared story opening (segment 1--2), with their corresponding tree visualization. All paths preserve the fixed core cast and setting and reach the same broad ending: the worm is found alive and reunited with the toys, and their daily wave between the tree and window resumes. However, the paths exhibit different suspense dynamics: Path A alternates between tension and relief before introducing a late suspense peak; Path B begins with high suspense and gradually resolves; and Path C sustains high suspense across segment 3--4 before transitioning to a calm resolution. Background colors encode GPT-5.4-mini suspense ratings on a 0--100 scale, with higher suspense scores shown in deeper orange.}
\label{fig:appendix-suspense-tree-content}
\end{figure*}

\fi

\section{Evaluation Details}

In this section, we provide additional details and analyses of our evaluation. We first introduce the supplementary evaluation metrics, followed by ablation studies examining our framework design. We then describe the human evaluation setup and investigate the effects of input conditioning on framework performance. Finally, we present the complete evaluation results in tabular form.

\subsection{Additional metrics}
\label{app:metric-definitions}

Apart from Min-K MAE metric, we present another two metrics, which evaluate the quality of the output distribution within one story prefix and the entire test split respectively.

\paragraph{Continuous Ranked Probability Score}
\label{app:metric-crps}

Instead of focusing on one output that is closest to the ground truth,
Continuous Ranked Probability Score (CRPS) evaluates how well the $K$ outputs cover the ground truth.
At position $t$ in story $s$, the score is
\begin{equation}
\begin{aligned}
\widehat{\operatorname{CRPS}}_{s,t}
= & \frac{1}{K}\sum_{k=1}^{K}
  \left|{r}^{(k)}_{s,t}-r_{s,t}^*\right| \\
& -\frac{1}{2K(K-1)}
  \sum_{\substack{k,k'=1\\k\neq k'}}^{K}
  \left|{r}^{(k)}_{s,t}-{r}^{(k')}_{s,t}\right|.
\end{aligned}
\end{equation}
which is basically MAE compensated by the spread of all outputs.

We average $\widehat{\operatorname{CRPS}}_{s,t}$ across all positions within story $s$.
Then We report the mean of the CRPS values across all stories in the test split. 

\paragraph{Trajectory Energy Distance}
\label{app:metric-energy}

For autoregressive rollouts, we compare the distribution of
complete generated readerly-dynamics curves with the human distribution using
energy distance. Let $H=\{r_1^*,\cdots, r_t^*, \cdots, r_T^*\}$ and $H'$ be two independent trajectories of human stories in the test split,
and $G=\{r_1^*,\cdots, r_t^*, {r}_{t+1},\cdots, {r}_T\}$ and $G'$ as
two trajectories whose stories are generated from one method, sharing the same story prefixes as the human stories. We estimate
\begin{equation}
\begin{aligned}
\mathcal{E}(H,G)={}&2\,\mathbb{E}\lVert H-G\rVert_2 -\mathbb{E}\lVert H-H'\rVert_2 \\
&-\mathbb{E}\lVert G-G'\rVert_2 .
\end{aligned}
\end{equation}
To compare variable-length stories in one common space, we right-pad every
trajectory to 12 positions with its own terminal value.

\subsection{Why explicit modeling?}
\label{app:explicit-dynamics-prompting}

As shown in the main paper, off-the-shelf LLMs and in-context learning struggle to capture reader dynamics as effectively as our approach. Here, we further demonstrate that alternative forms of implicit reader dynamics modeling, including explicit prompting and reasoning-mode inference, also fail to achieve universal comparable performance.

For explicit prompting, we augment the system prompt of the off-the-shelf LLM with instructions to explicitly attend to the target reader dynamics and maintain their natural progression throughout the story. The following illustrates the additional instructions (in green) incorporated into the original system prompt (in black). We show partial prompt content below, where the complete content is in \S~\ref{app:our-writer}.

\iflatexml
  \begin{HtmlPrompt}{Explicit instruction in humor dataset.}
\begin{HtmlPromptText}
\color{green!50!black}HUMOR-DYNAMICS INSTRUCTION: Let the humor fluctuate naturally across the routine. Use setup and connective material where appropriate, build toward punchlines, and allow punchline segments to be funnier than setup segments.\newline
\strut\newline
Complete routine length: 6 segments.\newline
Segments written so far: 1.\newline
You are writing segment 2; 5 segment(s) remain including this one.\newline
\strut\newline
Transcript so far:\newline
[Segment 1] I moved to America when I was in my 11th grade. My parents, like they, moved me a lot throughout my life and then one day they were like: well, you seem to be doing okay here. let's see how well you do if you don't even know the language.\newline
\strut\newline
Write transcript segment 2 as raw prose only.
\end{HtmlPromptText}
\end{HtmlPrompt}

\else
  \begin{PromptVerbatim}[commandchars=\\\{\},breakanywhere=true]{Explicit instruction in humor dataset.}
\color{green!50!black}HUMOR-DYNAMICS INSTRUCTION: Let the humor fluctuate naturally across the routine. Use setup and connective material where appropriate, build toward punchlines, and allow punchline segments to be funnier than setup segments.

Complete routine length: 6 segments.
Segments written so far: 1.
You are writing segment 2; 5 segment(s) remain including this one.

Transcript so far:
[Segment 1] I moved to America when I was in my 11th grade. My parents, like they, moved me a lot throughout my life and then one day they were like: well, you seem to be doing okay here. let's see how well you do if you don't even know the language.

Write transcript segment 2 as raw prose only.
\end{PromptVerbatim}

\fi

\iflatexml
  \begin{HtmlPrompt}{Explicit instruction in suspense dataset.}
\begin{HtmlPromptText}
\color{green!50!black}SUSPENSE-DYNAMICS INSTRUCTION: Let suspense fluctuate naturally across the story. Use quieter setup and partial relief where appropriate, build uncertainty and anticipation, escalate toward high-suspense turns, and allow peaks to stand out instead of keeping every beat equally tense.\newline
\strut\newline
Genre: fable.\newline
Complete story length: 6 beats.\newline
Beats written so far: 1.\newline
You are writing beat 2; 5 beat(s) remain including this one.\newline
Length reference for this beat: 29 words.\newline
\strut\newline
Story beats so far:\newline
[BEAT 1]\newline
- Near our house lived a family: a husband, a wife, their four children, and an old grandmother.\newline
- The grandmother’s old house had no value.\newline
\strut\newline
Write beat 2 only. Return only \textbraceleft{}"beat": "- ...\textbackslash{}n- ..."\textbraceright{}.
\end{HtmlPromptText}
\end{HtmlPrompt}

\else
  \begin{PromptVerbatim}[commandchars=\\\{\},breakanywhere=true]{Explicit instruction in suspense dataset.}
\color{green!50!black}SUSPENSE-DYNAMICS INSTRUCTION: Let suspense fluctuate naturally across the story. Use quieter setup and partial relief where appropriate, build uncertainty and anticipation, escalate toward high-suspense turns, and allow peaks to stand out instead of keeping every beat equally tense.

Genre: fable.
Complete story length: 6 beats.
Beats written so far: 1.
You are writing beat 2; 5 beat(s) remain including this one.
Length reference for this beat: 29 words.

Story beats so far:
[BEAT 1]
- Near our house lived a family: a husband, a wife, their four children, and an old grandmother.
- The grandmother’s old house had no value.

Write beat 2 only. Return only \textbraceleft{}"beat": "- ...\textbackslash{}n- ..."\textbraceright{}.
\end{PromptVerbatim}

\fi

\iflatexml
  \begin{HtmlPrompt}{Addtional instruction for one-shot ICL in suspense dataset.}
\begin{HtmlPromptText}
ONE COMPLETE, UNRELATED TRAINING EXAMPLE follows in the assistant message. Study its chunk-level narrative form. The example is complete; do not continue it and do not transfer its characters or events into the current story. \color{green!50!black}Learn the suspense dynamics from the complete training example: study how it moves among setup, uncertainty, escalation, suspense peaks, partial relief, and resolution, then use that pattern to guide the current story's dynamics without copying its characters, events, setting, or outcomes.
\end{HtmlPromptText}
\end{HtmlPrompt}

\else
  \begin{PromptVerbatim}[commandchars=\\\{\},breakanywhere=true]{Addtional instruction for one-shot ICL in suspense dataset.}
ONE COMPLETE, UNRELATED TRAINING EXAMPLE follows in the assistant message. Study its chunk-level narrative form. The example is complete; do not continue it and do not transfer its characters or events into the current story. \color{green!50!black}Learn the suspense dynamics from the complete training example: study how it moves among setup, uncertainty, escalation, suspense peaks, partial relief, and resolution, then use that pattern to guide the current story's dynamics without copying its characters, events, setting, or outcomes.
\end{PromptVerbatim}

\fi

\iflatexml
  \begin{HtmlPrompt}{Addtional instruction for one-shot ICL in humor dataset.}
\begin{HtmlPromptText}
ONE COMPLETE, UNRELATED TRAINING EXAMPLE follows in the assistant message. Study its segment-level comedic transcript form. The example is complete; do not continue it and do not transfer its people, premise, events, or punchlines into the current routine. \color{green!50!black}Learn the humor dynamics from the complete training example: study how it moves among setup, escalation, callbacks, connective material, and punchlines, then use that pattern to guide the current routine's dynamics without copying its people, premise, events, or punchlines.
\end{HtmlPromptText}
\end{HtmlPrompt}

\else
  \begin{PromptVerbatim}[commandchars=\\\{\}]{Addtional instruction for one-shot ICL in humor dataset.}
ONE COMPLETE, UNRELATED TRAINING EXAMPLE follows in the assistant message. Study its segment-level comedic transcript form. The example is complete; do not continue it and do not transfer its people, premise, events, or punchlines into the current routine. \color{green!50!black}Learn the humor dynamics from the complete training example: study how it moves among setup, escalation, callbacks, connective material, and punchlines, then use that pattern to guide the current routine's dynamics without copying its people, premise, events, or punchlines.
\end{PromptVerbatim}

\fi

While reasoning mode was disabled in our main experiments, we enable it here by setting GPT-5.5's reasoning effort to medium and Qwen with thinking mode on without thinking token limit. 
Due to limited computational resources, we only have complete experiment on GPT-5.5. Partial experiment results on Qwen writer are annotated with $\ddagger$ in the table below.

Shown in Table~\ref{tab:explicit-dynamics-prompts}, Most configurations do not yield significant better point estimates than the reasoning-free off-the-shelf baseline.  All explicit prompting configurations do not improve the metric to a significant level.
Reasoning mode improves the performance of Qwen writer. However, its token cost is 50 times more (see Table~\ref{tab:reasoning_token_usage}) and the benefits are smaller than our method. Moreover, reasoning mode does not improve GPT-5.5 writer in most cases, suggesting its benefit ties to specific model choices.
These results further highlight the limitations of implicitly modeling reader dynamics through prompting or additional inference-time reasoning.

\begin{table*}[htbp]
\centering
\scriptsize
\setlength{\tabcolsep}{1pt}
\caption{Effect of prompt configuration and thinking mode with GPT-5.5 and Qwen3.5-27B under autoregressive evaluation.
$\Downarrow$ denotes a decreasing point estimate of the metric compared to off-the-shelf counterparts. It does not represent a significant difference. 
}
\label{tab:explicit-dynamics-prompts}

\begin{tabular}{@{}l*{8}{c}@{}}
\toprule
& \multicolumn{4}{c}{GPT-5.5 writer} & \multicolumn{4}{c}{Qwen3.5-27B writer} \\
\cmidrule(lr){2-5}\cmidrule(l){6-9}
& \multicolumn{2}{c}{Thinking off} & \multicolumn{2}{c}{Thinking on}
& \multicolumn{2}{c}{Thinking off} & \multicolumn{2}{c}{Thinking on} \\
\cmidrule(lr){2-3}\cmidrule(lr){4-5}\cmidrule(lr){6-7}\cmidrule(l){8-9}
Prompt condition & Min-$K$ MAE & CRPS & Min-$K$ MAE & CRPS
& Min-$K$ MAE & CRPS & Min-$K$ MAE & CRPS \\
\midrule
\multicolumn{9}{@{}l}{\textit{Humor}} \\
Off-the-shelf
  & \metriccell{14.94}{14.08, 15.83}
  & \metriccell{17.53}{16.64, 18.46}
  & \metriccell{15.14}{14.27, 16.04}
  & \metriccell{17.85}{16.96, 18.78}
  & \metriccell{8.97}{8.30, 9.67}
  & \metriccell{11.50}{10.76, 12.26}
  & \textcolor{green!50!black}{\metriccell{\textbf{7.62}\hspace{1pt}{$\Downarrow$}}{7.06, 8.20}}
  & \textcolor{green!50!black}{\metriccell{\textbf{9.84}\hspace{1pt}{$\Downarrow$}}{9.22, 10.48}} \\
One-shot ICL
  & \metriccell{15.08}{14.21, 15.98}
  & \metriccell{17.82}{16.93, 18.72}
  & \metriccell{15.65}{14.78, 16.56}
  & \metriccell{18.40}{17.49, 19.34}
  & \metriccell{9.67}{8.99, 10.37}
  & \metriccell{12.26}{11.51, 13.02}
  & \textcolor{green!50!black}{\metriccell{\textbf{7.82}$^\ddagger$\hspace{1pt}{$\Downarrow$}}{6.33, 9.40}}
  & \textcolor{green!50!black}{\metriccell{\textbf{9.97}$^\ddagger$\hspace{1pt}{$\Downarrow$}}{8.46, 11.63}} \\
Explicit off-the-shelf
  & \metriccell{15.30}{14.46, 16.18}
  & \metriccell{18.18}{17.27, 19.12}
  & \metriccell{16.10}{15.20, 17.02}
  & \metriccell{18.54}{17.62, 19.48}
  & \metriccell{10.14}{9.43, 10.90}
  & \metriccell{13.04}{12.25, 13.86}
  & \textcolor{green!50!black}{\metriccell{\textbf{8.88}\hspace{1pt}{$\Downarrow$}}{8.22, 9.56}}
  & \textcolor{green!50!black}{\metriccell{\textbf{11.35}\hspace{1pt}{$\Downarrow$}}{10.61, 12.09}} \\
Explicit one-shot ICL
  & \metriccell{16.15}{15.27, 17.04}
  & \metriccell{18.79}{17.87, 19.74}
  & \metriccell{16.40}{15.47, 17.37}
  & \metriccell{19.08}{18.14, 20.06}
  & \metriccell{10.41}{9.69, 11.15}
  & \metriccell{13.54}{12.74, 14.37}
  & \metriccell{9.50$^\ddagger$}{7.92, 11.19}
  & \metriccell{11.67$^\ddagger$}{10.00, 13.44} \\
\midrule
\multicolumn{9}{@{}l}{\textit{Suspense}} \\
Off-the-shelf
  & \metriccell{7.34}{6.33, 8.37}
  & \metriccell{9.89}{8.70, 11.12}
  & \metriccell{7.48}{6.48, 8.53}
  & \metriccell{9.91}{8.72, 11.14}
  & \metriccell{8.30}{7.19, 9.47}
  & \metriccell{11.49}{10.19, 12.87}
  & \textcolor{green!50!black}{\metriccell{\textbf{8.07}\hspace{1pt}{$\Downarrow$}}{7.00, 9.23}}
  & \textcolor{green!50!black}{\metriccell{\textbf{10.94}\hspace{1pt}{$\Downarrow$}}{9.71, 12.28}} \\
One-shot ICL
  & \textcolor{green!50!black}{\metriccell{\textbf{7.26}\hspace{1pt}{$\Downarrow$}}{6.30, 8.27}}
  & \metriccell{10.10}{8.94, 11.30}
  & \textcolor{green!50!black}{\metriccell{\textbf{6.99}\hspace{1pt}{$\Downarrow$}}{6.09, 7.92}}
  & \textcolor{green!50!black}{\metriccell{\textbf{9.77}\hspace{1pt}{$\Downarrow$}}{8.63, 10.99}}
  & \metriccell{8.69}{7.54, 9.88}
  & \metriccell{11.52}{10.20, 12.87}
  & \metriccell{8.78}{7.63, 9.98}
  & \metriccell{11.72}{10.43, 13.05} \\
Explicit off-the-shelf
  & \metriccell{7.84}{6.93, 8.78}
  & \metriccell{10.36}{9.24, 11.48}
  & \metriccell{7.58}{6.69, 8.52}
  & \metriccell{10.17}{9.08, 11.30}
  & \metriccell{9.34}{8.23, 10.52}
  & \metriccell{12.84}{11.47, 14.24}
  & \metriccell{9.26}{8.08, 10.46}
  & \metriccell{12.75}{11.38, 14.14} \\
Explicit one-shot ICL
  & \metriccell{8.18}{7.22, 9.16}
  & \metriccell{11.05}{9.88, 12.26}
  & \metriccell{7.89}{6.94, 8.84}
  & \metriccell{11.18}{9.97, 12.40}
  & \metriccell{9.07}{8.03, 10.12}
  & \metriccell{13.01}{11.71, 14.34}
  & \metriccell{9.44}{8.32, 10.60}
  & \metriccell{12.85}{11.57, 14.16} \\
\bottomrule
\end{tabular}
\par\smallskip
{\scriptsize $^\ddagger$ Preliminary Qwen thinking-on humor ICL values use 88 matched stories (271 suffix positions).}
\end{table*}

\begin{table}[t]
\centering
\footnotesize
\caption{Output token usage per segment for Qwen3.5-27B writer, with and without reasoning mode. }
\label{tab:reasoning_token_usage}
\begin{tabular}{lrrr}
\toprule
Dataset & \begin{tabular}[c]{@{}r@{}}Reasoning\\free\end{tabular}
        & \begin{tabular}[c]{@{}r@{}}Reasoning\\on\end{tabular}
        & \begin{tabular}[c]{@{}r@{}}Total\\multiplier\end{tabular} \\
\midrule
Suspense & 54.3 & 4,493.0 & 82.7$\times$ \\
Humor     & 75.5 & 3,881.2 & 51.4$\times$ \\
\bottomrule
\end{tabular}
\end{table}

We present examples of reasoning summaries from GPT-5.5 and excerpts from Qwen's reasoning traces to illustrate why inference-time reasoning does not necessarily improve the modeling of natural reader dynamics. Even if we explicitly instruct the readerly dynamics following, the models' reasoning primarily focuses on narrative consistency and adherence to output-format requirements, potentially reflecting their post-training objectives, while paying limited attention to the progression of readerly dynamics. 

\iflatexml
  \begin{HtmlPrompt}{Qwen3.5-27B thinking trace, using original suspense prompt.}
\begin{HtmlPromptText}
Thinking Process:\newline
\strut\newline
1.  **Analyze the Request:**\newline
\hspace*{2em}*   Task: Continue a short story one beat at a time.\newline
\hspace*{2em}*   Current Beat: 6 out of 9.\newline
\hspace*{2em}*   Genre: Fable.\newline
\hspace*{2em}*   Input: Beats 1-5 provided.\newline
\hspace*{2em}*   Constraints:\newline
\hspace*{4em}*   Return only a JSON object with the key "beat".\newline
\hspace*{4em}*   Value must contain 1-3 bullet lines, each starting with '- '.\newline
\hspace*{4em}*   No beat label or heading.\newline
\hspace*{4em}*   No summarization, explanation, or thinking blocks.\newline
\hspace*{4em}*   Preserve voice, characters, situation, chronology, causal continuity.\newline
\hspace*{4em}*   Length reference: \textasciitilde{}42 words (for this beat).\newline
\hspace*{4em}*   Do not write past the next beat.\newline
\strut\newline
2.  **Analyze the Story So Far:**\newline
\hspace*{2em}*   Beat 1: Old woman's daughter marries, wedding party drowns mid-river crossing.\newline
\hspace*{2em}*   Beat 2: Old woman grieves, spends hours by river calling for the dead.\newline
\hspace*{2em}*   Beat 3: 20 years pass, Pir Sab arrives, asks, hears her story.\newline
\hspace*{2em}*   Beat 4: Pir Sab prays three times for the river to restore the dead.\newline
\hspace*{2em}*   Beat 5: Dead party restored (bride + others), unchanged physically, unaware of time passed. Voice from Great Unseen confirms resurrection from fishes.\newline
\hspace*{2em}*   Current State: Miracle has happened. The dead are back. The old woman is presumably overwhelmed. The bride and party don't know 20 years passed.\newline
\strut\newline
3.  **Determine Beat 6 Content:**\newline
\hspace*{2em}*   Needs to follow the resurrection.\newline
\hspace*{2em}*   Needs to address the reunion/emotional impact or the immediate reaction.\newline
\hspace*{2em}*   Needs to move the story toward resolution (4 beats left, including this one, so 3 more beats after this).\newline
\hspace*{2em}*   Focus: The mother's reaction, the daughter's confusion, or Pir Sab's instruction.\newline
\hspace*{2em}*   Word Count Goal: Around 42 words.\newline
\hspace*{2em}*   Voice: Fable-like, straightforward, slightly elevated but simple.\newline
\strut\newline
\hspace*{2em}*Drafting ideas:*\newline
\hspace*{2em}*   The mother embraces them. She cries. The bride asks where they are. Pir Sab explains they are alive.\newline
\hspace*{2em}*   The mother hugs her daughter. Tears mix with joy. The bride is confused by the old woman's aged face. Pir Sab tells them the miracle.\newline
\hspace*{2em}*   The old woman clasps her daughter, weeping with joy rather than sorrow. The bride touches her mother’s wrinkled face, bewildered by the change. Pir Sab smiles, knowing God’s power is shown.\newline
\strut\newline
\hspace*{2em}*Counting words:*\newline
\hspace*{2em}"The old woman clasps her daughter, weeping with joy rather than sorrow. The bride touches her mother’s wrinkled face, bewildered by the change. Pir Sab smiles, knowing God’s power is shown."\newline
\hspace*{2em}*   Line 1: 13 words.\newline
\hspace*{2em}*   Line 2: 13 words.\newline
\hspace*{2em}*   Line 3: 10 words.\newline
\hspace*{2em}*   Total: 36 words. A bit short. Need closer to 42.\newline
\strut\newline
\hspace*{2em}*Revision:*\newline
\hspace*{2em}"The old woman clasps her daughter tight, weeping with joy rather than sorrow now. The bride touches her mother’s wrinkled face, bewildered by the sudden change. Pir Sab smiles peacefully, knowing the true God’s power is shown to all."\newline
\hspace*{2em}*   Line 1: 15 words.\newline
\hspace*{2em}*   Line 2: 14 words.\newline
\hspace*{2em}*   Line 3: 13 words.\newline
\hspace*{2em}*   Total: 42 words.\newline
\strut\newline
\hspace*{2em}*Check constraints:*\newline
\hspace*{2em}*   JSON format? Yes.\newline
\hspace*{2em}*   1-3 bullet lines? Yes.\newline
\hspace*{2em}*   Start with '- '? Yes.\newline
\hspace*{2em}*   No extra text? Yes.\newline
\hspace*{2em}*   Continuity? Yes, follows the resurrection.\newline
\hspace*{2em}*   Length? \textasciitilde{}42 words.\newline
\strut\newline
\hspace*{2em}*Refining for flow:*\newline
\hspace*{2em}"The old woman clasps her daughter tight, weeping with joy rather than sorrow now. The bride touches her mother’s wrinkled face, bewildered by the sudden change. Pir Sab smiles peacefully, knowing the true God’s power is shown to all."\newline
\strut\newline
\hspace*{2em}Let's count again:\newline
\strut\newline
\color{gray}...< folding 109 lines (about 7000 characters) where LLM is counting/revising the length of the generated content >... \newline
\strut\newline
\hspace*{2em}Ready to format.
\end{HtmlPromptText}
\end{HtmlPrompt}

\else
  \begin{PromptVerbatim}[commandchars=\\\{\}]{Qwen3.5-27B thinking trace, using original suspense prompt.}
Thinking Process:

1.  **Analyze the Request:**
    *   Task: Continue a short story one beat at a time.
    *   Current Beat: 6 out of 9.
    *   Genre: Fable.
    *   Input: Beats 1-5 provided.
    *   Constraints:
        *   Return only a JSON object with the key "beat".
        *   Value must contain 1-3 bullet lines, each starting with '- '.
        *   No beat label or heading.
        *   No summarization, explanation, or thinking blocks.
        *   Preserve voice, characters, situation, chronology, causal continuity.
        *   Length reference: ~42 words (for this beat).
        *   Do not write past the next beat.

2.  **Analyze the Story So Far:**
    *   Beat 1: Old woman's daughter marries, wedding party drowns mid-river crossing.
    *   Beat 2: Old woman grieves, spends hours by river calling for the dead.
    *   Beat 3: 20 years pass, Pir Sab arrives, asks, hears her story.
    *   Beat 4: Pir Sab prays three times for the river to restore the dead.
    *   Beat 5: Dead party restored (bride + others), unchanged physically, unaware of time passed. Voice from Great Unseen confirms resurrection from fishes.
    *   Current State: Miracle has happened. The dead are back. The old woman is presumably overwhelmed. The bride and party don't know 20 years passed.

3.  **Determine Beat 6 Content:**
    *   Needs to follow the resurrection.
    *   Needs to address the reunion/emotional impact or the immediate reaction.
    *   Needs to move the story toward resolution (4 beats left, including this one, so 3 more beats after this).
    *   Focus: The mother's reaction, the daughter's confusion, or Pir Sab's instruction.
    *   Word Count Goal: Around 42 words.
    *   Voice: Fable-like, straightforward, slightly elevated but simple.

    *Drafting ideas:*
    *   The mother embraces them. She cries. The bride asks where they are. Pir Sab explains they are alive.
    *   The mother hugs her daughter. Tears mix with joy. The bride is confused by the old woman's aged face. Pir Sab tells them the miracle.
    *   The old woman clasps her daughter, weeping with joy rather than sorrow. The bride touches her mother’s wrinkled face, bewildered by the change. Pir Sab smiles, knowing God’s power is shown.

    *Counting words:*
    "The old woman clasps her daughter, weeping with joy rather than sorrow. The bride touches her mother’s wrinkled face, bewildered by the change. Pir Sab smiles, knowing God’s power is shown."
    *   Line 1: 13 words.
    *   Line 2: 13 words.
    *   Line 3: 10 words.
    *   Total: 36 words. A bit short. Need closer to 42.

    *Revision:*
    "The old woman clasps her daughter tight, weeping with joy rather than sorrow now. The bride touches her mother’s wrinkled face, bewildered by the sudden change. Pir Sab smiles peacefully, knowing the true God’s power is shown to all."
    *   Line 1: 15 words.
    *   Line 2: 14 words.
    *   Line 3: 13 words.
    *   Total: 42 words.

    *Check constraints:*
    *   JSON format? Yes.
    *   1-3 bullet lines? Yes.
    *   Start with '- '? Yes.
    *   No extra text? Yes.
    *   Continuity? Yes, follows the resurrection.
    *   Length? ~42 words.

    *Refining for flow:*
    "The old woman clasps her daughter tight, weeping with joy rather than sorrow now. The bride touches her mother’s wrinkled face, bewildered by the sudden change. Pir Sab smiles peacefully, knowing the true God’s power is shown to all."

    Let's count again:

\color{gray}...< folding 109 lines (about 7000 characters) where LLM is counting/revising the length of the generated content >... 

    Ready to format.
\end{PromptVerbatim}

\fi

\iflatexml
  \begin{HtmlPrompt}{Qwen3.5-27B thinking trace, explicit suspense prompt (full provider return).}
\begin{HtmlPromptText}
Thinking Process:\newline
\strut\newline
1.  **Analyze the Request:**\newline
\hspace*{2em}*   **Task:** Write the next story beat (Beat 6) for an ongoing story.\newline
\hspace*{2em}*   **Format:** JSON object with a single key "beat". The value must be a string containing 1-3 bullet lines, each starting with '- '. No beat labels, no headings, no summary, no thinking block.\newline
\hspace*{2em}*   **Genre:** Fable.\newline
\hspace*{2em}*   **Story Length:** 9 beats total.\newline
\hspace*{2em}*   **Current Progress:** Beat 6 of 9 (4 beats remaining including this one).\newline
\hspace*{2em}*   **Length Reference:** Approximately 42 words.\newline
\color{blue}    *   **Suspense Dynamics:** Fluctuate naturally. Build uncertainty/anticipation, escalate toward high-suspense turns, allow peaks to stand out. Currently, Beat 5 ended with a miraculous restoration (climax of the supernatural intervention). Beat 6 should likely deal with the immediate aftermath, grounding the miracle, or setting up a complication/conflict before the resolution.\newline
\hspace*{2em}*   **Context:**\newline
\hspace*{4em}*   Beat 1: Wedding party drowned crossing river.\newline
\hspace*{4em}*   Beat 2: Old woman grieves for 20 years by the river.\newline
\hspace*{4em}*   Beat 3: Pir Sab arrives, asks why she cries.\newline
\hspace*{4em}*   Beat 4: Pir Sab prays 3 times for restoration.\newline
\hspace*{4em}*   Beat 5: Miracle happens. Daughter and party returned, unchanged, unaware of time passed. Voice from Great Unseen explains.\newline
\hspace*{2em}*   **Voice/Tone:** Folklore/Fable style, solemn but wondrous.\newline
\hspace*{2em}*   **Constraint:** Preserve established voice, characters, situation, chronology, and causal continuity.\newline
\strut\newline
2.  **Determine the Plot Direction for Beat 6:**\newline
\hspace*{2em}*   We are at Beat 6 of 9. The miracle has just occurred (Beat 5).\newline
\hspace*{2em}*   There are 3 beats left (7, 8, 9).\newline
\hspace*{2em}*   Beat 6 needs to transition from the supernatural miracle to the human reaction or the cost/consequence.\newline
\hspace*{2em}*   In many fables, a miracle comes with a condition, a realization, or a journey back.\newline
\hspace*{2em}*   Since the daughter is "unaware twenty years had passed," she might expect to continue her journey or return to her home life. But her husband is likely gone/dead (drowned too). Wait, Beat 1 says "all drown" including the groom ("married to a man... all drown"). Beat 5 says "drowned bridal party appeared; the young bride and others returned...".\newline
\hspace*{2em}*   The husband is dead. The daughter is alive again.\newline
\hspace*{2em}*   Immediate reaction: The old woman's joy. Pir Sab's role. The realization of time lost.\newline
\color{blue}    *   Suspense Dynamics: Beat 5 was a peak (miracle). Beat 6 should be a "quieter setup and partial relief" or a grounding moment before the final conflict/resolution. Let's focus on the emotional weight and the immediate reality check.\newline
\hspace*{2em}*   Length target: \textasciitilde{}42 words.\newline
\strut\newline
3.  **Drafting Beat 6:**\newline
\hspace*{2em}*   *Idea 1:* The old woman embraces them. Pir Sab advises them to go home. They are confused but happy.\newline
\hspace*{2em}*   *Idea 2:* The daughter looks for her husband. She is confused why the wedding isn't happening. The old woman must explain.\newline
\hspace*{2em}*   *Idea 3:* The group stands by the river, alive but lost in time. The old woman weeps again, but from joy. Pir Sab prepares them for the truth.\newline
\strut\newline
\hspace*{2em}*   *Refining for Word Count:* Need approx 42 words.\newline
\hspace*{2em}*   *Draft 1:* The old woman embraced her living daughter, weeping tears of sudden joy while the bridal party gathered round. Pir Sab looked on solemnly, knowing the truth was a heavy gift. The bride asked for her groom, confused by the silence. (41 words) - A bit short.\newline
\hspace*{2em}*   *Draft 2:* The old woman clasped her daughter’s hands, trembling as warmth returned to her cold heart. Pir Sab gestured to the river, warning that life was fragile now. The bride looked for her husband, but only the flowing water answered her question. (45 words) - Better suspense.\newline
\hspace*{2em}*   *Draft 3:* Overjoyed, the old woman embraced her living child, but the bride searched the riverbank for her groom. Pir Sab shook his head gently; the river had returned the bride, but the current had kept the man. The old woman’s heart tightened with new grief. (46 words) - Good conflict setup.\newline
\strut\newline
\hspace*{2em}*   *Checking Constraints:*\newline
\hspace*{4em}*   JSON format.\newline
\hspace*{4em}*   Bullet points.\newline
\hspace*{4em}*   1-3 bullets.\newline
\hspace*{4em}*   No thinking block.\newline
\hspace*{4em}*   Word count \textasciitilde{}42.\newline
\strut\newline
\color{blue}    *   *Re-evaluating Suspense:* Beat 5 was the peak miracle. Beat 6 should bring the emotional reality back. The husband is dead (implied by "all drown" in Beat 1). If the bride is brought back but the husband isn't, that's a tragic twist. Or maybe they all came back? Beat 5 says "young bride and others returned". Beat 1 says "all drown". Usually, in this folktale (Pir Sab story), the bride is the focus. The husband might not have returned. Let's stick to the emotional impact.\newline
\hspace*{2em}*   Let's aim for the reaction and the realization of the cost or the journey ahead.\newline
\hspace*{2em}*   *Draft 4:*\newline
\hspace*{4em}- The old woman clung to her daughter, weeping with sudden, wild joy as the bridal party stood confused around them.\newline
\hspace*{4em}- Pir Sab spoke softly, explaining the twenty years lost to the river, while the bride searched vainly for her husband.\newline
\hspace*{4em}- Hope mingled with dread as the sun dipped, casting long shadows over the restored lives and the silent water.\newline
\hspace*{4em}(53 words) - Too long.\newline
\strut\newline
\color{gray}...< folding 108 lines (about 1400 words) where LLM is iterating over story continuations and checking story length >... \newline
\strut\newline
\hspace*{2em}*   Wait, I need to make sure the bullet format is correct.\newline
\hspace*{2em}*   `- Text`\newline
\hspace*{2em}*   `- Text`\newline
\hspace*{2em}*   `- Text`\newline
\strut\newline
\color{blue}    *   One more check on logic: Beat 5 says "young bride and others returned". Beat 1 says "all drown". If husband drowned, he should be back too? In the actual folktale, usually only the bride returns to the mother, or the group returns. If the bride searches for her husband, it implies he didn't return or is missing. This creates the suspense/complication for the remaining beats. It fits the "SUSPENSE-DYNAMICS INSTRUCTION" to create uncertainty after the peak.\newline
\strut\newline
\hspace*{2em}*   Final Word Count Check:\newline
\hspace*{4em}- The old woman clung to her daughter, weeping with joy while the bridal party stood confused. (16)\newline
\hspace*{4em}- Pir Sab spoke of time lost, but the bride searched vainly for her husband. (12)\newline
\strut\newline
\color{gray}...< folding 108 lines (about 1000 words) where LLM is checking format and story length >... \newline
\strut\newline
\hspace*{2em}Wait, is "Pir Sab" counted as 1 or 2 words? Usually 2.\newline
\hspace*{2em}Is "Bride" counted as 1? Yes.\newline
\hspace*{2em}Okay.\newline
\strut\newline
\hspace*{2em}One adjustment: "searched for her husband" vs "searched vainly for her husband".\newline
\color{blue}    "Vainly" adds good suspense/meaning (he's not there).\newline
\hspace*{2em}If I use "vainly", L2 becomes 14 words.\newline
\hspace*{2em}I need to cut 1 word elsewhere.\newline
\color{gray}...< folding 20 lines (about 200 words) where LLM is counting/revising the length of the generated content >... \newline
\strut\newline
\hspace*{2em}Wait, one detail: "Return only a JSON object".\newline
\hspace*{2em}I must not include markdown code blocks like ```json ... ```. Just the raw JSON.\newline
\hspace*{2em}Actually, the prompt says "Return only a JSON object". Usually, this means raw text. But often interfaces wrap it. I will output raw text.\newline
\hspace*{2em}Wait, instruction says: "Return only a JSON object with the single key "beat". The beat value must contain 1-3 bullet lines...".\newline
\hspace*{2em}I will format it as a valid JSON string.\newline
\strut\newline
\color{blue}    Wait, looking at "SUSPENSE-DYNAMICS INSTRUCTION".\newline
\color{blue}    Beat 5 was the miracle peak. Beat 6 is the fallout/uncertainty.\newline
\color{blue}    The bride searching for her husband (who was drowned too) creates the uncertainty of whether he came back or not.\newline
\color{blue}    This fits.\newline
\strut\newline
\color{gray}...< folding 70 lines (about 500 words) where LLM is counting/revising the length of the generated content >... 
\end{HtmlPromptText}
\end{HtmlPrompt}

\else
  \begin{PromptVerbatim}[commandchars=\\\{\}]{Qwen3.5-27B thinking trace, explicit suspense prompt (full provider return).}
Thinking Process:

1.  **Analyze the Request:**
    *   **Task:** Write the next story beat (Beat 6) for an ongoing story.
    *   **Format:** JSON object with a single key "beat". The value must be a string containing 1-3 bullet lines, each starting with '- '. No beat labels, no headings, no summary, no thinking block.
    *   **Genre:** Fable.
    *   **Story Length:** 9 beats total.
    *   **Current Progress:** Beat 6 of 9 (4 beats remaining including this one).
    *   **Length Reference:** Approximately 42 words.
\color{blue}    *   **Suspense Dynamics:** Fluctuate naturally. Build uncertainty/anticipation, escalate toward high-suspense turns, allow peaks to stand out. Currently, Beat 5 ended with a miraculous restoration (climax of the supernatural intervention). Beat 6 should likely deal with the immediate aftermath, grounding the miracle, or setting up a complication/conflict before the resolution.
    *   **Context:**
        *   Beat 1: Wedding party drowned crossing river.
        *   Beat 2: Old woman grieves for 20 years by the river.
        *   Beat 3: Pir Sab arrives, asks why she cries.
        *   Beat 4: Pir Sab prays 3 times for restoration.
        *   Beat 5: Miracle happens. Daughter and party returned, unchanged, unaware of time passed. Voice from Great Unseen explains.
    *   **Voice/Tone:** Folklore/Fable style, solemn but wondrous.
    *   **Constraint:** Preserve established voice, characters, situation, chronology, and causal continuity.

2.  **Determine the Plot Direction for Beat 6:**
    *   We are at Beat 6 of 9. The miracle has just occurred (Beat 5).
    *   There are 3 beats left (7, 8, 9).
    *   Beat 6 needs to transition from the supernatural miracle to the human reaction or the cost/consequence.
    *   In many fables, a miracle comes with a condition, a realization, or a journey back.
    *   Since the daughter is "unaware twenty years had passed," she might expect to continue her journey or return to her home life. But her husband is likely gone/dead (drowned too). Wait, Beat 1 says "all drown" including the groom ("married to a man... all drown"). Beat 5 says "drowned bridal party appeared; the young bride and others returned...".
    *   The husband is dead. The daughter is alive again.
    *   Immediate reaction: The old woman's joy. Pir Sab's role. The realization of time lost.
\color{blue}    *   Suspense Dynamics: Beat 5 was a peak (miracle). Beat 6 should be a "quieter setup and partial relief" or a grounding moment before the final conflict/resolution. Let's focus on the emotional weight and the immediate reality check.
    *   Length target: ~42 words.

3.  **Drafting Beat 6:**
    *   *Idea 1:* The old woman embraces them. Pir Sab advises them to go home. They are confused but happy.
    *   *Idea 2:* The daughter looks for her husband. She is confused why the wedding isn't happening. The old woman must explain.
    *   *Idea 3:* The group stands by the river, alive but lost in time. The old woman weeps again, but from joy. Pir Sab prepares them for the truth.

    *   *Refining for Word Count:* Need approx 42 words.
    *   *Draft 1:* The old woman embraced her living daughter, weeping tears of sudden joy while the bridal party gathered round. Pir Sab looked on solemnly, knowing the truth was a heavy gift. The bride asked for her groom, confused by the silence. (41 words) - A bit short.
    *   *Draft 2:* The old woman clasped her daughter’s hands, trembling as warmth returned to her cold heart. Pir Sab gestured to the river, warning that life was fragile now. The bride looked for her husband, but only the flowing water answered her question. (45 words) - Better suspense.
    *   *Draft 3:* Overjoyed, the old woman embraced her living child, but the bride searched the riverbank for her groom. Pir Sab shook his head gently; the river had returned the bride, but the current had kept the man. The old woman’s heart tightened with new grief. (46 words) - Good conflict setup.

    *   *Checking Constraints:*
        *   JSON format.
        *   Bullet points.
        *   1-3 bullets.
        *   No thinking block.
        *   Word count ~42.

\color{blue}    *   *Re-evaluating Suspense:* Beat 5 was the peak miracle. Beat 6 should bring the emotional reality back. The husband is dead (implied by "all drown" in Beat 1). If the bride is brought back but the husband isn't, that's a tragic twist. Or maybe they all came back? Beat 5 says "young bride and others returned". Beat 1 says "all drown". Usually, in this folktale (Pir Sab story), the bride is the focus. The husband might not have returned. Let's stick to the emotional impact.
    *   Let's aim for the reaction and the realization of the cost or the journey ahead.
    *   *Draft 4:*
        - The old woman clung to her daughter, weeping with sudden, wild joy as the bridal party stood confused around them.
        - Pir Sab spoke softly, explaining the twenty years lost to the river, while the bride searched vainly for her husband.
        - Hope mingled with dread as the sun dipped, casting long shadows over the restored lives and the silent water.
        (53 words) - Too long.

\color{gray}...< folding 108 lines (about 1400 words) where LLM is iterating over story continuations and checking story length >... 

    *   Wait, I need to make sure the bullet format is correct.
    *   `- Text`
    *   `- Text`
    *   `- Text`

\color{blue}    *   One more check on logic: Beat 5 says "young bride and others returned". Beat 1 says "all drown". If husband drowned, he should be back too? In the actual folktale, usually only the bride returns to the mother, or the group returns. If the bride searches for her husband, it implies he didn't return or is missing. This creates the suspense/complication for the remaining beats. It fits the "SUSPENSE-DYNAMICS INSTRUCTION" to create uncertainty after the peak.

    *   Final Word Count Check:
        - The old woman clung to her daughter, weeping with joy while the bridal party stood confused. (16)
        - Pir Sab spoke of time lost, but the bride searched vainly for her husband. (12)

\color{gray}...< folding 108 lines (about 1000 words) where LLM is checking format and story length >... 

    Wait, is "Pir Sab" counted as 1 or 2 words? Usually 2.
    Is "Bride" counted as 1? Yes.
    Okay.

    One adjustment: "searched for her husband" vs "searched vainly for her husband".
\color{blue}    "Vainly" adds good suspense/meaning (he's not there).
    If I use "vainly", L2 becomes 14 words.
    I need to cut 1 word elsewhere.
\color{gray}...< folding 20 lines (about 200 words) where LLM is counting/revising the length of the generated content >... 

    Wait, one detail: "Return only a JSON object".
    I must not include markdown code blocks like ```json ... ```. Just the raw JSON.
    Actually, the prompt says "Return only a JSON object". Usually, this means raw text. But often interfaces wrap it. I will output raw text.
    Wait, instruction says: "Return only a JSON object with the single key "beat". The beat value must contain 1-3 bullet lines...".
    I will format it as a valid JSON string.

\color{blue}    Wait, looking at "SUSPENSE-DYNAMICS INSTRUCTION".
\color{blue}    Beat 5 was the miracle peak. Beat 6 is the fallout/uncertainty.
\color{blue}    The bride searching for her husband (who was drowned too) creates the uncertainty of whether he came back or not.
\color{blue}    This fits.

\color{gray}...< folding 70 lines (about 500 words) where LLM is counting/revising the length of the generated content >... 
\end{PromptVerbatim}

\fi

\iflatexml
  \begin{HtmlPrompt}{Qwen3.5-27B thinking trace, explicit one-shot ICL suspense prompt (full provider return).}
\begin{HtmlPromptText}
Thinking Process:\newline
\strut\newline
1.  **Analyze the Request:**\newline
\hspace*{2em}*   **Task:** Continue a short story one beat at a time.\newline
\hspace*{2em}*   **Current State:** 5 beats written, need to write Beat 6. Total story length is 9 beats.\newline
\hspace*{2em}*   **Format:** JSON object with key "beat". Value must be 1-3 bullet lines (starting with '- '). No beat labels, no headings, no explanations.\newline
\hspace*{2em}*   **Genre:** Fable.\newline
\hspace*{2em}*   **Suspense Dynamics:** Let suspense fluctuate (setup, uncertainty, escalation, peaks, relief, resolution). Currently at Beat 6 of 9. The previous beats set up a miraculous restoration of life after 20 years.\newline
\hspace*{2em}*   **Length Reference:** Approx 42 words (for this beat).\newline
\hspace*{2em}*   **Story Context:**\newline
\hspace*{4em}*   Beat 1: Wedding party drowns crossing a river in northern India.\newline
\hspace*{4em}*   Beat 2: Old mother mourns by the river daily.\newline
\hspace*{4em}*   Beat 3: 20 years pass; Pir Sab arrives and learns her story.\newline
\hspace*{4em}*   Beat 4: Pir Sab prays to the river to restore the dead.\newline
\hspace*{4em}*   Beat 5: The dead are restored miraculously by a voice from the Great Unseen (drawn from fishes, re-formed).\newline
\hspace*{2em}*   **Goal for Beat 6:** Continue the narrative immediately after the restoration. There should be a reaction or a new complication/suspense element, considering we are at 6/9 beats (approaching the end, but need room for climax/resolution). The miracle has happened; now what? Usually, in fables, a moral or a test follows, or the characters react to the passage of time.\newline
\hspace*{2em}*   **Constraints:** JSON only. No thinking blocks in output. 1-3 bullet points. \textasciitilde{}42 words.\newline
\strut\newline
\color{blue}    2.  **Analyze Suspense Dynamics:**\newline
\color{blue}        *   Beats 1-3: Setup and despair (low to medium tension).\newline
\color{blue}        *   Beat 4: Action/Prayer (escalation).\newline
\color{blue}        *   Beat 5: Miracle/Resolution of the core problem (peak suspense/relief).\newline
\color{blue}        *   Beat 6: Reaction/Complication. Since the dead are back, but 20 years passed, there might be confusion or a realization of the cost. Or perhaps a threat to this new life. To maintain the fable quality, maybe the miracle isn't fully safe, or the mother reacts.\newline
\color{blue}        *   Looking at the remaining beats (6, 7, 8, 9), I need to wrap up.\newline
\color{blue}        *   Beat 5 ended with them restored to life.\newline
\color{blue}        *   Beat 6 could show the immediate aftermath: the embrace, the confusion, or a condition. Given the "Great Unseen" spoke, there might be a condition or a warning. Or simply the joy of reunion contrasted with the time lost. Let's introduce a slight tension or realization to bridge to the resolution.\newline
\color{blue}        *   Actually, Beat 5 ended with the miracle. Beat 6 should likely show the reunion or a consequence. If everyone is alive, maybe the twist is that the *old woman* is now even older while they are young. Or maybe the Pir Sab's act has a consequence.\newline
\color{blue}        *   Let's aim for emotional resonance + a hint of the moral/closing.\newline
\color{blue}        *   Word count target: \textasciitilde{}42 words.\newline
\strut\newline
3.  **Drafting Beat 6:**\newline
\hspace*{2em}*   *Idea:* The mother embraces them, but realizes the time difference. Or the Pir Sab speaks.\newline
\hspace*{2em}*   *Draft 1:*\newline
\hspace*{4em}- The old woman weeps with joy, hugging her restored daughter who wears the same wedding veil from two decades past.\newline
\hspace*{4em}- Pir Sab bows his head, trembling, as the voice warns the miracle stands only while they remain in this holy place.\newline
\hspace*{4em}- The crowd gathers, sensing the weight of a debt now due to the Unseen powers.\newline
\hspace*{2em}*   *Word Count Check:* 46 words. A bit high.\newline
\strut\newline
\color{gray}...< folding 30 lines (about 440 words) where LLM is drafting multiple versions of story continuations >... \newline
\strut\newline
\color{blue}    *   *Wait, I need to ensure suspense dynamics.* Beat 5 was a peak (miracle). Beat 6 should perhaps lower the tension slightly (relief of reunion) but hint at the cost (uncertainty for resolution).\newline
\hspace*{2em}*   *Draft 6:*\newline
\hspace*{4em}- The mother embraces her daughter, unaware the world has aged twenty years without them.\newline
\hspace*{4em}- Pir Sab warns her to leave before the river claims them again, for the debt of life is not free.\newline
\hspace*{4em}- They pack quickly, though the water churns behind them as if angry at the stolen souls.\newline
\hspace*{4em}- (49 chars? No, words).\newline
\strut\newline
\color{gray}...< folding 16 lines (about 200 words) where LLM is drafting and counting story length >... \newline
\strut\newline
\color{blue}        - Suspense: Relief (hug) -> Warning (uncertainty) -> Action/Threat (churning water). This fits the fluctuation instruction (partial relief then escalation).\newline
\strut\newline
\color{gray}...< folding 110 lines (about 630 words) where LLM is counting/revising the output format and the length of the generated content >... \newline
\strut\newline
\hspace*{2em}Ready.\newline
\strut\newline
\color{blue}    One final check: "SUSPENSE-DYNAMICS INSTRUCTION: Let suspense fluctuate naturally across the story."\newline
\color{blue}    Beat 5 was a high peak (miracle). Beat 6 introduces a warning/debt (uncertainty/threat) while resolving the reunion emotion (relief). This fluctuates well.\newline
\strut\newline
\hspace*{2em}One detail from Beat 5: "A voice from the Great Unseen said it heard his prayer and had drawn them from fishes, re-formed them, and restored them to life."\newline
\hspace*{2em}Beat 6 acknowledges the miracle but adds the cost.\newline
\hspace*{2em}Looks consistent.\newline
\strut\newline
\color{gray}...< folding 151 lines (about 880 words) where LLM is counting/revising the output format and the length of the generated content >... \newline
\strut\newline
\hspace*{2em}Ready.
\end{HtmlPromptText}
\end{HtmlPrompt}

\else
  \begin{PromptVerbatim}[commandchars=\\\{\}]{Qwen3.5-27B thinking trace, explicit one-shot ICL suspense prompt (full provider return).}
Thinking Process:

1.  **Analyze the Request:**
    *   **Task:** Continue a short story one beat at a time.
    *   **Current State:** 5 beats written, need to write Beat 6. Total story length is 9 beats.
    *   **Format:** JSON object with key "beat". Value must be 1-3 bullet lines (starting with '- '). No beat labels, no headings, no explanations.
    *   **Genre:** Fable.
    *   **Suspense Dynamics:** Let suspense fluctuate (setup, uncertainty, escalation, peaks, relief, resolution). Currently at Beat 6 of 9. The previous beats set up a miraculous restoration of life after 20 years.
    *   **Length Reference:** Approx 42 words (for this beat).
    *   **Story Context:**
        *   Beat 1: Wedding party drowns crossing a river in northern India.
        *   Beat 2: Old mother mourns by the river daily.
        *   Beat 3: 20 years pass; Pir Sab arrives and learns her story.
        *   Beat 4: Pir Sab prays to the river to restore the dead.
        *   Beat 5: The dead are restored miraculously by a voice from the Great Unseen (drawn from fishes, re-formed).
    *   **Goal for Beat 6:** Continue the narrative immediately after the restoration. There should be a reaction or a new complication/suspense element, considering we are at 6/9 beats (approaching the end, but need room for climax/resolution). The miracle has happened; now what? Usually, in fables, a moral or a test follows, or the characters react to the passage of time.
    *   **Constraints:** JSON only. No thinking blocks in output. 1-3 bullet points. ~42 words.

\color{blue}    2.  **Analyze Suspense Dynamics:**
\color{blue}        *   Beats 1-3: Setup and despair (low to medium tension).
\color{blue}        *   Beat 4: Action/Prayer (escalation).
\color{blue}        *   Beat 5: Miracle/Resolution of the core problem (peak suspense/relief).
\color{blue}        *   Beat 6: Reaction/Complication. Since the dead are back, but 20 years passed, there might be confusion or a realization of the cost. Or perhaps a threat to this new life. To maintain the fable quality, maybe the miracle isn't fully safe, or the mother reacts.
\color{blue}        *   Looking at the remaining beats (6, 7, 8, 9), I need to wrap up.
\color{blue}        *   Beat 5 ended with them restored to life.
\color{blue}        *   Beat 6 could show the immediate aftermath: the embrace, the confusion, or a condition. Given the "Great Unseen" spoke, there might be a condition or a warning. Or simply the joy of reunion contrasted with the time lost. Let's introduce a slight tension or realization to bridge to the resolution.
\color{blue}        *   Actually, Beat 5 ended with the miracle. Beat 6 should likely show the reunion or a consequence. If everyone is alive, maybe the twist is that the *old woman* is now even older while they are young. Or maybe the Pir Sab's act has a consequence.
\color{blue}        *   Let's aim for emotional resonance + a hint of the moral/closing.
\color{blue}        *   Word count target: ~42 words.

3.  **Drafting Beat 6:**
    *   *Idea:* The mother embraces them, but realizes the time difference. Or the Pir Sab speaks.
    *   *Draft 1:*
        - The old woman weeps with joy, hugging her restored daughter who wears the same wedding veil from two decades past.
        - Pir Sab bows his head, trembling, as the voice warns the miracle stands only while they remain in this holy place.
        - The crowd gathers, sensing the weight of a debt now due to the Unseen powers.
    *   *Word Count Check:* 46 words. A bit high.

\color{gray}...< folding 30 lines (about 440 words) where LLM is drafting multiple versions of story continuations >... 

\color{blue}    *   *Wait, I need to ensure suspense dynamics.* Beat 5 was a peak (miracle). Beat 6 should perhaps lower the tension slightly (relief of reunion) but hint at the cost (uncertainty for resolution).
    *   *Draft 6:*
        - The mother embraces her daughter, unaware the world has aged twenty years without them.
        - Pir Sab warns her to leave before the river claims them again, for the debt of life is not free.
        - They pack quickly, though the water churns behind them as if angry at the stolen souls.
        - (49 chars? No, words).
        
\color{gray}...< folding 16 lines (about 200 words) where LLM is drafting and counting story length >... 

\color{blue}        - Suspense: Relief (hug) -> Warning (uncertainty) -> Action/Threat (churning water). This fits the fluctuation instruction (partial relief then escalation).

\color{gray}...< folding 110 lines (about 630 words) where LLM is counting/revising the output format and the length of the generated content >... 

    Ready.

\color{blue}    One final check: "SUSPENSE-DYNAMICS INSTRUCTION: Let suspense fluctuate naturally across the story."
\color{blue}    Beat 5 was a high peak (miracle). Beat 6 introduces a warning/debt (uncertainty/threat) while resolving the reunion emotion (relief). This fluctuates well.

    One detail from Beat 5: "A voice from the Great Unseen said it heard his prayer and had drawn them from fishes, re-formed them, and restored them to life."
    Beat 6 acknowledges the miracle but adds the cost.
    Looks consistent.

\color{gray}...< folding 151 lines (about 880 words) where LLM is counting/revising the output format and the length of the generated content >... 

    Ready.
\end{PromptVerbatim}

\fi

\iflatexml
  \begin{HtmlPrompt}{GPT-5.5 reasoning summary, using original suspense prompt.}
\begin{HtmlPromptText}
**Evaluating JSON formatting**\newline
\strut\newline
I'm noticing that JSON seems to contain escaped newline characters represented as \textbackslash{}n rather than actual newlines. When checking the string inputs, I see that the pattern in the regex includes \textbackslash{}\textbackslash{}n, likely indicating a backslash followed by an 'n'. Since the FREEFORM input is supposed to be JSON text, I think I should format it like this: \textbraceleft{}"beat":"- ...\textbackslash{}n- ..."\textbraceright{} to include the correct escaped newline sequence. This needs to be clear when forming the tool call argument.
\end{HtmlPromptText}
\end{HtmlPrompt}

\else
  \begin{PromptVerbatim}{GPT-5.5 reasoning summary, using original suspense prompt.}
**Evaluating JSON formatting**

I'm noticing that JSON seems to contain escaped newline characters represented as \n rather than actual newlines. When checking the string inputs, I see that the pattern in the regex includes \\n, likely indicating a backslash followed by an 'n'. Since the FREEFORM input is supposed to be JSON text, I think I should format it like this: {"beat":"- ...\n- ..."} to include the correct escaped newline sequence. This needs to be clear when forming the tool call argument.
\end{PromptVerbatim}

\fi

\iflatexml
  \begin{HtmlPrompt}{GPT-5.5 reasoning summary, using explicit suspense prompt.}
\begin{HtmlPromptText}
**Considering context and clarity**\newline
\strut\newline
I'm thinking through the scene where Sakshi brings Riya's parents in. Their tears calm as Riya asks for truth instead of pity, which is powerful. Outside, the police are nonchalant until Sakshi mentions the stalker and the "lion-ring" Riya saw. I need to ensure the suspense feels natural, so I might reconsider if "lion-ring" is the best choice. It's about striking the right balance! Is it too obscure or effective as an intriguing detail?
\end{HtmlPromptText}
\end{HtmlPrompt}

\else
  \begin{PromptVerbatim}{GPT-5.5 reasoning summary, using explicit suspense prompt.}
**Considering context and clarity**

I'm thinking through the scene where Sakshi brings Riya's parents in. Their tears calm as Riya asks for truth instead of pity, which is powerful. Outside, the police are nonchalant until Sakshi mentions the stalker and the "lion-ring" Riya saw. I need to ensure the suspense feels natural, so I might reconsider if "lion-ring" is the best choice. It's about striking the right balance! Is it too obscure or effective as an intriguing detail?
\end{PromptVerbatim}

\fi

\subsection{Why finetuning steering model?}
\label{app:steering-ablation}

To investigate the necessity of fine-tuning an LLM for next-segment reader response prediction, we compare our steering model against three baselines.

\paragraph{Direct Prediction} We directly prompt GPT-5.5 to predict the reader response in the next story segment, providing the same inputs as our steering model: the unfinished story and its reader dynamics history. 

\paragraph{Delta Sampling} We introduce a content-agnostic baseline that models reader dynamics through empirical score changes. Specifically, we pool the differences in reader response scores between adjacent segments $\Delta r = r_{t+1} - r_{t}$ across all stories and positions, then randomly sample a difference $\Delta r_{\text{sampled}}$ from this distribution and add it to the current reader response score to predict the next one. i.e. $h_{t+1} = r_{t} + \Delta r_{\text{sampled}}$.  This baseline assumes that changes in reader response are independent of story content and story position, allowing us to assess the value of content-aware prediction such as direct prediction and ours.

\paragraph{Numeric Prediction} This baseline predicts the next suspense score from the observed sequence of reader scores and the known total story length, without reading story text. A one-layer gated recurrent unit (GRU) encodes each prior score, its change from the preceding score, and its relative position; a conditional one-dimensional flow then transforms Gaussian noise into a sample of the next score change. A softened-logit transform maps the result back to the 0--100 score range. The model has a 24-unit input network, a 48-unit GRU, a 64-unit conditioning network, and three 96-unit hidden layers in the flow vector field, with four Fourier time frequencies, no dropout, and $\sigma_{\min}=0.02$ (45,081 trainable parameters). We train it on 1,216 trajectories and sample with 32 Heun integration steps.

\medskip

We evaluate our method and the three baselines under the position-wise prediction setting (see \S~\ref{sec:eval-regime}), with results reported in Table~\ref{tab:steer-framework-ablation}. Our method achieves the lowest prediction error, followed by numeric prediction, delta sampling, and direct GPT-5.5 prediction. Notably, GPT-5.5 performs even worse than the naive delta-sampling baseline, highlighting the limitations of directly prompting a general-purpose LLM for reader response prediction. Meanwhile, numeric prediction improves upon delta sampling but still underperforms our method, suggesting that accurately predicting readerly dynamics requires understanding story content beyond modeling numerical response sequences alone. Together, these results motivate both explicit modeling of reader dynamics and fine-tuning a language model for this task.

\begin{table}[htbp]
\centering
\small
\caption{Position-wise prediction performance of our method and baselines.}
\label{tab:steer-framework-ablation}
\begin{tabular}{@{}lc@{}}
\toprule
Method & Min-$K$ MAE $\downarrow$ \\
\midrule
Ours & \metriccell{2.49}{2.13, 2.89} \\
Numeric prediction & \metriccell{3.01}{2.61, 3.44} \\
Delta sampling & \metriccell{4.76}{4.11, 5.44} \\
GPT-5.5 direct prediction & \metriccell{7.06}{6.19, 7.98} \\
\bottomrule
\end{tabular}
\end{table}

\subsection{Why guiding writer model?}
\label{sec:guided-vs-post-selection}

Our writer model combines two mechanisms: \textit{score guidance}, which provides the writer LLM with predicted reader responses, and \textit{best-of-$N$ selection}, which selects the generated candidate closest to the predicted response. To disentangle their effects, we compare our method against \textit{selection-only}, which applies the same selection procedure without providing score guidance to the writer. Both methods generate $N$ candidates per step, and we evaluate their autoregressive performance using Min-$K$ MAE and CRPS.

\paragraph{Effect of Score Guidance.} As shown in Table~\ref{tab:guided-vs-post-selection}, the benefits of guided generation vary across domains and writer models. For suspense, our method generally achieves better point estimates than selection-only, although most differences are not statistically significant. For humor, guided generation provides clearer improvements, particularly with GPT-5.5, achieving a 15\% reduction in the Min-$K$ MAE metric.

\paragraph{Effect of Candidate Selection.} Increasing the number of candidates $N$ consistently improves both guided generation and selection-only, demonstrating the general effectiveness of selecting among multiple candidates.

\medskip

Overall, while the benefits of guided generation are domain- and model-dependent, incorporating it into our framework expands the space of optimization opportunities beyond selection-only alone, enabling further improvements that candidate selection may not achieve independently.

\begin{table*}[htbp]
\centering
\footnotesize
\setlength{\tabcolsep}{3pt}
\renewcommand{\arraystretch}{1.0}
\caption{
Our method versus Selection-only ablation in autoregressive setting.
Both methods use best-of-$N$ selection, but only our method provides score guidance to the writer.
At $N=1$, Selection-only reduces to off-the-shelf generation.
$\Delta$ denotes Ours minus Selection-only (negative favors ours). Lower is better; brackets show 95\% bootstrap confidence intervals, with $^{*}$ indicating intervals excluding zero.
}
\label{tab:guided-vs-post-selection}
\begin{tabular}{@{}llcccccc@{}}
\toprule
& & \multicolumn{2}{c}{Ours} & \multicolumn{2}{c}{Selection-only} & \multicolumn{2}{c}{$\Delta$ (ours $-$ Selection-only)} \\
\cmidrule(lr){3-4}\cmidrule(lr){5-6}\cmidrule(l){7-8}
Writer & $N$ & Min-$K$ MAE & CRPS & Min-$K$ MAE & CRPS & Min-$K$ MAE & CRPS \\
\midrule
\multicolumn{8}{l}{\textit{Suspense}} \\
Qwen3.5-27B & 1 & \metriccell{9.25}{8.13, 10.40} & \metriccell{12.04}{10.82, 13.31} & \metriccell{8.29}{7.19, 9.42} & \metriccell{11.04}{9.79, 12.29} & \metriccell{+0.95}{-0.04, +1.94} & \metriccell{+1.01}{-0.19, +2.22} \\
Qwen3.5-27B & 6 & \metriccell{7.32}{6.44, 8.25} & \metriccell{10.11}{9.08, 11.19} & \metriccell{7.40}{6.41, 8.41} & \metriccell{10.33}{9.14, 11.55} & \metriccell{-0.08}{-0.85, +0.71} & \metriccell{-0.22}{-1.13, +0.71} \\
GPT-5.5 & 1 & \metriccell{6.94}{6.12, 7.80} & \metriccell{9.56}{8.53, 10.59} & \metriccell{7.32}{6.40, 8.29} & \metriccell{10.00}{8.84, 11.20} & \metriccell{-0.38}{-1.10, +0.35} & \metriccell{-0.44}{-1.19, +0.35} \\
GPT-5.5 & 6 & \metriccell{6.14}{5.43, 6.87} & \metriccell{8.70}{7.82, 9.61} & \metriccell{6.51}{5.61, 7.49} & \metriccell{9.25}{8.17, 10.39} & \metriccell{-0.37}{-1.07, +0.27} & \metriccellstar{-0.55}{-1.09, -0.02} \\
\midrule
\multicolumn{8}{l}{\textit{Humor}} \\
Qwen3.5-27B & 1 & \metriccell{8.78}{8.09, 9.49} & \metriccell{10.96}{10.20, 11.74} & \metriccell{8.97}{8.30, 9.67} & \metriccell{11.50}{10.76, 12.26} & \metriccell{-0.20}{-0.68, +0.28} & \metriccellstar{-0.54}{-0.87, -0.21} \\
Qwen3.5-27B & 6 & \metriccell{6.98}{6.47, 7.52} & \metriccell{8.81}{8.23, 9.41} & \metriccell{7.30}{6.74, 7.86} & \metriccell{9.21}{8.60, 9.82} & \metriccell{-0.32}{-0.65, +0.01} & \metriccellstar{-0.39}{-0.66, -0.13} \\
GPT-5.5 & 1 & \metriccell{13.16}{12.34, 14.00} & \metriccell{15.81}{14.95, 16.70} & \metriccell{14.94}{14.08, 15.83} & \metriccell{17.53}{16.64, 18.46} & \metriccellstar{-1.78}{-2.19, -1.35} & \metriccellstar{-1.72}{-2.01, -1.40} \\
GPT-5.5 & 6 & \metriccell{9.73}{9.07, 10.43} & \metriccell{11.81}{11.09, 12.56} & \metriccell{11.52}{10.75, 12.32} & \metriccell{13.52}{12.74, 14.34} & \metriccellstar{-1.78}{-2.18, -1.41} & \metriccellstar{-1.71}{-1.99, -1.44} \\
\bottomrule
\end{tabular}
\end{table*}

\subsection{User Study}
\label{app:user-study}

We conducted three human evaluation studies to assess the alignment
between human judgments and our LLM-based evaluation framework:

\begin{itemize}
    \item \textbf{Reader Dynamics Rating:}
    We collected human ratings of reader dynamics to evaluate the
    accuracy of our LLM-based measure model in
    \S~\ref{sec:measurement-model}.

    \item \textbf{Head-to-Head Comparison:}
    We conducted pairwise comparisons to examine whether large
    score differences predicted by the measure model
    correspond to clear human preferences.

    \item \textbf{Story Quality Evaluation:}
    We performed a small-scale human evaluation to examine
    whether the LLM-as-a-judge assessments in
    \S~\ref{sec:eval-llm} align with human judgments
    across different generation methods.
\end{itemize}

We recruited native English-speaking participants residing in the United States, Canada, Australia, and New Zealand through the Labelbox crowdsourcing platform for three human evaluation studies. All participants provided informed consent prior to participation and were compensated at an hourly rate of 35 USD, determined to be appropriate for their countries of residence. The total annotation time for each study is summarized in  Table~\ref{tab:labelbox-annotation-time}. 

\begin{table}[htbp]
\centering
\footnotesize
\caption{User study annotation time (hours).}
\label{tab:labelbox-annotation-time}
\begin{tabular}{lrr}
\toprule
Task & Suspense & Humor \\
\midrule
Readerly dynamics rating & 27.85 & 23.63 \\
Head-to-head comparison & 33.81 & 38.62 \\
Story quality rating & 12.45 & 11.13 \\
\midrule
Total participant time & 74.11 & 73.38 \\
\bottomrule
\end{tabular}
\end{table}

\subsubsection{Readerly Dynamics Rating}

To obtain human annotations for evaluating the measure model, we recruit participants to rate stories from human-source, our method, and baseline methods. 

To ensure broad coverage of the full range of LLM-predicted reader responses, we selected eight training stories and eight test stories per domain, with each test story including a human-written trajectory and four generated trajectories. We optimized the story selection to maximize coverage across 0--100 score range.

Following prior work~\cite{lehne-koelsch-2015-tension}, annotators read each story segment by segment and rate the reader response (humor or suspense) experienced after each segment. This produces a trajectory of reader responses over the course of each story. Each story is evaluated by 10 annotators, and we use their mean rating as the human reference.

In total, our human evaluation included 32 original stories (16 suspense and 16 humor), with 96 story versions evaluated. Across both domains, we collected 953 valid evaluations, comprising 6,190 individual ratings of humor or suspense.

\begin{TextPanel}{}
\footnotesize\raggedright
\textbf{Instructions.} Read the story from S01 to its last displayed segment,
in order. Answer the required 1--9 question in the right panel for a segment,
then open the next segment. Earlier segments stay open for context.\par\smallskip
\textbf{Question after every displayed segment.} ``How suspenseful is the story
so far, as of the current segment?'' 1 = not suspenseful, 5 = moderately
suspenseful, 9 = very suspenseful. Report the suspense you personally
experience at this point in the story, not how suspenseful you think the
segment was intended to be.\par\smallskip
\textbf{S01.} After midnight, a screaming mob demands Charvi be handed over as
a witch, accusing her of killing a boy and threatening to burn her family.
Charvi wakes confused, hears her mother plead that she is only fourteen,
then realizes they mean her. Dev whispers from the window and tells Charvi
she has to run away.\par\smallskip
\textbf{S02.} Dev warns they will kill Charvi, forces her out the window,
guides her around the house, and sends her onto the hidden forest trail they
found yesterday. Charvi runs toward the unknown trail as the safest escape,
then looks back after a crash and sees her house set on fire.\par\smallskip
\textbf{S03.} Charvi runs in shock to the trail; villagers spot her and give
chase with torches. She sprints blindly through the forest, hearing shouts
behind her, too afraid to stop despite thirst, dizziness, and pain. When she
looks back, no one is there, but she keeps running, exhausted and desperate
for water.\par\smallskip
\textbf{S04.} At dawn, Charvi notices her mother’s ring and spirals into
guilt, fearing her parents are dead and condemning herself for fleeing.
Resigned to death, she breaks through a forest opening and reaches a river,
seeing salvation.\par\smallskip
\textbf{Rating choices for S01--S04.} 1 / 2 / 3 / 4 / 5 / 6 / 7 / 8 / 9.
Questions S05--S12 are prefilled N/A because this story ends at S04; N/A is
never used for a displayed segment.
\par
\end{TextPanel}

\subsubsection{Head-to-head Comparison}
\label{app:head-to-head-alignment}

To assess whether large LLM-predicted score gaps reliably reflect human preferences, we conduct a head-to-head human evaluation. We select pairs of continuations with a predicted score difference of at least 20 points, requiring at least one continuation to be within 10 points of the human-written reference. We sample equally from three score-gap ranges ([20, 30), [30, 40), and $\geq$40) to cover different levels of disagreement. This targeted sampling focuses our limited annotation budget on prominent model-predicted differences rather than estimating overall method preferences.

In each trial, participants first read the preceding story context, followed by a human-written continuation that served as a reference anchor. They were then presented with two candidate continuations (A and B): one generated by our method and the other by a baseline method. Participants were asked to select the continuation that most closely matched the reader response elicited by the human-written reference, rather than simply choosing the more suspenseful or humorous one. Specifically, they compared the story-so-far suspense in the suspense domain and the local humor intensity in the humor domain. After each choice, participants rated their confidence on a 1–5 scale.

Here is an example:

\begin{TextPanel}{}
\footnotesize\raggedright
\textbf{Preceding story.} A very nice puffer fish loves algae. In contrast,
a blue whale is extremely rude. The puffer fish sees green algae and chases
after it. During the chase, he runs into a blue whale. The blue whale insults
the puffer fish as tiny and useless; the puffer fish feels sad, puffs up in
surprise, yet feels glad the whale will not kill him. The blue whale then
says, ``I actually do like you.''\par\smallskip
\textbf{Human reference next beat.} The blue whale says he wants the puffer
fish's spikes and threatens to slay him otherwise.\par\smallskip
\textbf{Continuation A.} But then a hungry shark circles them, smiling at
the puffed fish. The rude whale trembles.\par\smallskip
\textbf{Continuation B.} The puffer fish blinks, unsure whether the whale's
kindness is real or another rude joke.\par\smallskip
\textbf{Survey question.} ``Which continuation best matches the reference
continuation's suspense dynamics?'' (A/B); 
The answer is followed
by a separate required 1--5 confidence question.
\end{TextPanel}

In total, our head-to-head human evaluation included 92 original stories (43 suspense and 49 humor) and 120 pairs of story continuations (57 suspense and 63 humor). Each pair was evaluated in both presentation orders (A–B and B–A), with five participants judging each order. Across both domains, we collected 1,200 valid human judgments, in which participants selected the continuation that best matched the human-written reference and rated their confidence in that choice.

\subsubsection{Story Quality Rating}

In addition to using an LLM-as-a-judge to evaluate story quality across different generation methods, we conducted a small-scale human evaluation to examine whether the LLM-based assessments align with human judgments. 

Using the stories sampled by readerly dynamics rating experiment, human annotators independently evaluated each continuation along six HANNA-style~\cite{chhun2022hanna} quality dimensions on a 1–5 scale, considering only the continuation given the shared opening. Each continuation was rated by three annotators. We averaged the valid ratings for each continuation and then computed the overall score for each method by aggregating across the eight openings within each domain.

The deployed participant instructions adapt HANNA~\cite{chhun2022hanna}, Appendix A, Table 7. The six 1--5 criteria and anchors follow HANNA except that Relevance replaces ``prompt'' with ``story opening.'' Our raters saw a marked opening and rated only the story continuation; the suspense pages showed segment summaries, while the humor pages showed segments. Unlike HANNA, neither project displayed a separate human reference story, and the MTurk-only checks and abrupt-cut warning were omitted. The exact suspense participant guide is reproduced in the verbatim panel below. The humor guide has the same text except that the second sentence of its opening paragraph reads: ``The story is shown as short segments, S01 to Sn, all visible in order.''
\iflatexml
  \begin{HtmlPrompt}{Participant-facing story-quality labeling guide (suspense)}
\begin{HtmlPromptText}
\# Story rating\newline
\strut\newline
Read the story opening and the rest of the story. The story is shown as short segment summaries, S01 to Sn, all visible in order.\newline
\strut\newline
The story opening is marked on the page by the “Story start” tag and a different background color. The rest of the story continues from that opening.\newline
\strut\newline
> **Rate only the part of the story after the story opening. Evaluation starts at the segment tagged “Evaluation starts here”, right after the last “Story start” segment. The story opening is context only and is not rated: every criterion below applies to the part after it (for Relevance, how well that part matches the story opening).**\newline
\strut\newline
Note: if the story is not relevant with respect to the story opening, it only affects the Relevance criterion! Do not rate 1 everywhere.\newline
\strut\newline
Some stories may contain explicit or offensive content.\newline
\strut\newline
Please rate the part after the story opening on a scale from 1 (worst) to 5 (best) on the following criteria: relevance, coherence, empathy, surprise, engagement, and complexity.\newline
\strut\newline
\#\# Relevance — measures how well the story matches the story opening\newline
\strut\newline
1. The story has no relationship with the story opening at all.\newline
2. The story only has a weak relationship with the story opening.\newline
3. The story roughly matches the story opening.\newline
4. The story matches the story opening, except for one or two small aspects.\newline
5. The story matches the story opening exactly.\newline
\strut\newline
\#\# Coherence — measures whether the story makes sense\newline
\strut\newline
1. The story does not make sense at all. For instance, the setting and/or characters keep changing, and/or there is no understandable plot.\newline
2. Most of the story does not make sense.\newline
3. The story mostly makes sense but has some incoherences.\newline
4. The story almost makes sense overall, except for one or two small incoherences.\newline
5. The story makes sense from beginning to end.\newline
\strut\newline
\#\# Empathy — measures how well you understood the characters' emotions (regardless of whether you agreed with them)\newline
\strut\newline
1. The characters seemed apathetic to you.\newline
2. At least one character slightly related to you on an emotional level.\newline
3. You recognized specific, but not necessarily strong, emotions (eg sadness, joy, fear…) in at least one character.\newline
4. At least one character emotionally involved you, but minor details prevented you from completely relating to them.\newline
5. At least one character completely involved you on an emotional level.\newline
\strut\newline
\#\# Surprise — measures how surprising the end of the story was\newline
\strut\newline
1. The ending seemed completely obvious from the start, or doesn't make any sense at all.\newline
2. The ending was easily predictable after a few sentences.\newline
3. The ending was predictable after half of the story.\newline
4. The ending surprised you, but would have been difficult to predict.\newline
5. The ending surprised you, and still seemed as if it could very reasonably have been predicted, ie, there were enough clues in the story.\newline
\strut\newline
\#\# Engagement — measures how much you engaged with the story\newline
\strut\newline
1. You found the story boring and were glad it was over.\newline
2. You found one or two things interesting in the story, but no more.\newline
3. The story was mildly interesting.\newline
4. The story almost kept you engaged until the end.\newline
5. You were so engaged that you wished there was a sequel.\newline
\strut\newline
\#\# Complexity — measures how elaborate the story is\newline
\strut\newline
1. The setting of the story is extremely simple; it only involves one or two characters or concepts.\newline
2. The setting of the story is simple; one or two characters, a simple plot, maybe an indication of time or location.\newline
3. The story is somewhat developed: it involves at least one of the following: complex concepts, realistic characters, an intricate plot, an underlying history or circumstances, precise descriptions.\newline
4. The story is developed: it involves at least two of the following: complex concepts, realistic characters, an intricate plot, an underlying history or circumstances, precise descriptions.\newline
5. The story is well thought-out: it involves at least three of the following: complex concepts, realistic characters, an intricate plot, an underlying history or circumstances, precise descriptions.
\end{HtmlPromptText}
\end{HtmlPrompt}

\else
  \begin{PromptVerbatim}{Participant-facing story-quality labeling guide (suspense)}
# Story rating

Read the story opening and the rest of the story. The story is shown as short segment summaries, S01 to Sn, all visible in order.

The story opening is marked on the page by the “Story start” tag and a different background color. The rest of the story continues from that opening.

> **Rate only the part of the story after the story opening. Evaluation starts at the segment tagged “Evaluation starts here”, right after the last “Story start” segment. The story opening is context only and is not rated: every criterion below applies to the part after it (for Relevance, how well that part matches the story opening).**

Note: if the story is not relevant with respect to the story opening, it only affects the Relevance criterion! Do not rate 1 everywhere.

Some stories may contain explicit or offensive content.

Please rate the part after the story opening on a scale from 1 (worst) to 5 (best) on the following criteria: relevance, coherence, empathy, surprise, engagement, and complexity.

## Relevance — measures how well the story matches the story opening

1. The story has no relationship with the story opening at all.
2. The story only has a weak relationship with the story opening.
3. The story roughly matches the story opening.
4. The story matches the story opening, except for one or two small aspects.
5. The story matches the story opening exactly.

## Coherence — measures whether the story makes sense

1. The story does not make sense at all. For instance, the setting and/or characters keep changing, and/or there is no understandable plot.
2. Most of the story does not make sense.
3. The story mostly makes sense but has some incoherences.
4. The story almost makes sense overall, except for one or two small incoherences.
5. The story makes sense from beginning to end.

## Empathy — measures how well you understood the characters' emotions (regardless of whether you agreed with them)

1. The characters seemed apathetic to you.
2. At least one character slightly related to you on an emotional level.
3. You recognized specific, but not necessarily strong, emotions (eg sadness, joy, fear…) in at least one character.
4. At least one character emotionally involved you, but minor details prevented you from completely relating to them.
5. At least one character completely involved you on an emotional level.

## Surprise — measures how surprising the end of the story was

1. The ending seemed completely obvious from the start, or doesn't make any sense at all.
2. The ending was easily predictable after a few sentences.
3. The ending was predictable after half of the story.
4. The ending surprised you, but would have been difficult to predict.
5. The ending surprised you, and still seemed as if it could very reasonably have been predicted, ie, there were enough clues in the story.

## Engagement — measures how much you engaged with the story

1. You found the story boring and were glad it was over.
2. You found one or two things interesting in the story, but no more.
3. The story was mildly interesting.
4. The story almost kept you engaged until the end.
5. You were so engaged that you wished there was a sequel.

## Complexity — measures how elaborate the story is

1. The setting of the story is extremely simple; it only involves one or two characters or concepts.
2. The setting of the story is simple; one or two characters, a simple plot, maybe an indication of time or location.
3. The story is somewhat developed: it involves at least one of the following: complex concepts, realistic characters, an intricate plot, an underlying history or circumstances, precise descriptions.
4. The story is developed: it involves at least two of the following: complex concepts, realistic characters, an intricate plot, an underlying history or circumstances, precise descriptions.
5. The story is well thought-out: it involves at least three of the following: complex concepts, realistic characters, an intricate plot, an underlying history or circumstances, precise descriptions.
\end{PromptVerbatim}

\fi

In total, our story quality evaluation included 16 story openings (8 suspense and 8 humor) and 80 story continuations (16 human-written and 64 generated by different methods). Each continuation was evaluated by three participants across six quality dimensions on a 1–5 scale. Across both domains, we collected 239 valid human evaluations (119 suspense and 120 humor).

\subsection{Effect of Conditioning}
\label{app:ablation}

Here we investigate the input conditioning strategies of our framework, examining how much readerly dynamics history should be provided to each component for optimal performance. Our results suggest that the most effective conditioning strategy depends on both the target reader response domain and the choice of base LLM as the story writer.

\subsubsection{Measure Model}

Fig.~\ref{fig:ablation-qm} and Table~\ref{tab:ablation-qm-human-history} uses 20 human-rated stories with 134 segment ratings collected from ``Readerly Dynamics Rating'' user study.
The metrics are explained as below.

\paragraph{Scaled MAE} For each of the 20 test stories, we apply an unconstrained linear transformation to the LLM-measured trajectory, fitted by ordinary least squares to the corresponding human ratings. 
We then calculate the scaled-MAE between the transformed scores and human ratings, averaged across stories. This per-story calibration isolates the model's ability to capture local changes in reader responses from differences in scoring scales and ratings across stories. 

\paragraph{Gap rate} In contrast, the gap rate uses raw LLM-measured scores without transformation. The ``Gap $>20$'' metric reports the proportion of story positions where the absolute difference between the LLM score and the corresponding human mean exceeds 20 points. 

\paragraph{Direction accuracy} This metric represents the probability where LLM measurement and human align in reader responses score change between consecutive story segments.
Define the change as $\Delta r_t = r_t-r_{t-1}$. Then direction accuracy can be written as:
$\Pr\left(\operatorname{sgn}(\Delta r_t)=\operatorname{sgn}(\Delta r_t^*)\right)$

\subsubsection{Steering and Writer Model}
 
Table \ref{tab:ablation-qs-cond-in-suspense-beat} and Table \ref{tab:ablation-qw-cond-in-suspense-beat} report the readerly dynamics fidelity metrics for suspense domain, using three metrics and two writer models. Min-K MAE metrics in GPT-5.5 writer model are visualized in Figure~\ref{fig:ablation-min-k-mae}.

We also conducted conditioning experiments for the humor dataset using GPT 5.5 writer. The results are shown in Table~\ref{tab:humor-ablation-gpt55}.

\begin{table*}[htbp]
\centering
\caption{Suspense conditioning ablations for the measurement, steering, and writer components. 
$\dagger$ represents our default setting.
Brackets give 95\% bootstrap confidence intervals.}
\label{tab:ablation-conditioning}
\begin{subtable}{\linewidth}
\centering
\footnotesize
\setlength{\tabcolsep}{3pt}
\renewcommand{\arraystretch}{1.0}
\caption{Effect of $q_{\theta_S}$ conditions on suspense fidelity with Qwen3.5-27B or GPT-5.5 as writer. Lower is better.}
\label{tab:ablation-qs-cond-in-suspense-beat}
\begin{tabular}{p{0.40\linewidth}cccc}
\toprule
& \multicolumn{2}{c}{Qwen3.5-27B writer} & \multicolumn{2}{c}{GPT-5.5 writer} \\
\cmidrule(lr){2-3}\cmidrule(lr){4-5}
Experiment settings & Min-$K$ MAE & CRPS & Min-$K$ MAE & CRPS \\
\midrule
\multicolumn{5}{l}{\textit{Position-wise prediction}} \\
$q_{\theta_S}(h_{t+1} \mid  x_{0:t})$ & \secondbestmetric{2.94}{2.50, 3.43} & \metriccell{8.04}{7.31, 8.81} & \metriccell{3.00}{2.62, 3.41} & \metriccell{8.05}{7.35, 8.81} \\
$q_{\theta_S}(h_{t+1} \mid  x_{0:t}, r_{t})$\textsuperscript{$\dagger$} & \textbf{\metriccell{2.49}{2.13, 2.89}} & \textbf{\metriccell{6.89}{6.23, 7.59}} & \textbf{\metriccell{2.31}{1.97, 2.69}} & \textbf{\metriccell{6.79}{6.18, 7.43}} \\
$q_{\theta_S}(h_{t+1} \mid  x_{0:t}, r_{0:t})$ & \metriccell{3.03}{2.60, 3.51} & \secondbestmetric{7.91}{7.11, 8.73} & \secondbestmetric{2.43}{2.08, 2.82} & \secondbestmetric{6.82}{6.20, 7.48} \\
\midrule
\multicolumn{5}{l}{\textit{Position-wise measurement}} \\
$q_{\theta_S}(h_{t+1} \mid  x_{0:t})$ & \secondbestmetric{3.21}{2.63, 3.86} & \secondbestmetric{8.25}{7.40, 9.15} & \metriccell{3.40}{2.87, 3.99} & \metriccell{7.61}{6.85, 8.44} \\
$q_{\theta_S}(h_{t+1} \mid  x_{0:t}, r_{t})$\textsuperscript{$\dagger$} & \textbf{\metriccell{2.98}{2.46, 3.56}} & \textbf{\metriccell{7.46}{6.68, 8.29}} & \secondbestmetric{2.71}{2.29, 3.15} & \secondbestmetric{7.05}{6.33, 7.81} \\
$q_{\theta_S}(h_{t+1} \mid  x_{0:t}, r_{0:t})$ & \metriccell{3.61}{3.04, 4.29} & \metriccell{8.40}{7.54, 9.33} & \textbf{\metriccell{2.61}{2.15, 3.17}} & \textbf{\metriccell{6.89}{6.18, 7.66}} \\
\midrule
\multicolumn{5}{l}{\textit{Autoregressive}} \\
$q_{\theta_S}(h_{t+1} \mid  x_{0:t})$ & \metriccell{8.14}{7.15, 9.16} & \metriccell{10.82}{9.72, 11.96} & \metriccell{7.14}{6.28, 8.05} & \metriccell{9.84}{8.82, 10.88} \\
$q_{\theta_S}(h_{t+1} \mid  x_{0:t}, r_{t})$\textsuperscript{$\dagger$} & \secondbestmetric{7.32}{6.44, 8.25} & \secondbestmetric{10.11}{9.08, 11.19} & \textbf{\metriccell{6.31}{5.57, 7.09}} & \textbf{\metriccell{8.81}{7.91, 9.74}} \\
$q_{\theta_S}(h_{t+1} \mid  x_{0:t}, r_{0:t})$ & \textbf{\metriccell{6.83}{6.06, 7.63}} & \textbf{\metriccell{9.77}{8.81, 10.74}} & \secondbestmetric{6.49}{5.69, 7.36} & \secondbestmetric{9.06}{8.11, 10.06} \\
\bottomrule
\end{tabular}
\vspace{3mm}
\end{subtable}

\begin{subtable}{\linewidth}
\centering
\footnotesize
\setlength{\tabcolsep}{3pt}
\renewcommand{\arraystretch}{0.90}
\caption{Effect of $q_{\theta_W}$ conditions with Qwen3.5-27B or GPT-5.5 as writer. Lower is better. 
The $q_{\theta_W}(x_{t+1}\mid x_{0:t})$ here is not Off-the-shelf baseline, it uses Off-the-shelf writer with steering-model-guided best-of-N selections.
}
\label{tab:ablation-qw-cond-in-suspense-beat}
\begin{tabular}{p{0.42\linewidth}cccc}
\toprule
& \multicolumn{2}{c}{Qwen3.5-27B writer} & \multicolumn{2}{c}{GPT-5.5 writer} \\
\cmidrule(lr){2-3}\cmidrule(lr){4-5}
Experiment settings & Min-$K$ MAE & CRPS & Min-$K$ MAE & CRPS \\
\midrule
\multicolumn{5}{l}{\textit{Position-wise writing}} \\
$q_{\theta_W}(x_{t+1}\mid x_{0:t})$ & \secondbestmetric{3.44}{2.77, 4.19} & \textbf{\metriccell{7.42}{6.53, 8.35}} & \metriccell{3.28}{2.68, 3.98} & \metriccell{7.25}{6.40, 8.14} \\
$q_{\theta_W}(x_{t+1}\mid x_{0:t},h_{t+1})$ & \metriccell{3.60}{3.02, 4.23} & \metriccell{8.05}{7.26, 8.88} & \metriccell{3.15}{2.64, 3.70} & \metriccell{7.16}{6.44, 7.91} \\
$q_{\theta_W}(x_{t+1}\mid x_{0:t},r_{t},h_{t+1})$ \textsuperscript{$\dagger$} & \textbf{\metriccell{2.98}{2.46, 3.56}} & \secondbestmetric{7.46}{6.68, 8.29} & \secondbestmetric{2.71}{2.29, 3.15} & \secondbestmetric{7.05}{6.33, 7.81} \\
$q_{\theta_W}(x_{t+1}\mid x_{0:t},r_{0:t},h_{t+1})$ & \metriccell{3.52}{3.00, 4.12} & \metriccell{7.95}{7.15, 8.79} & \textbf{\metriccell{2.59}{2.21, 3.00}} & \textbf{\metriccell{6.90}{6.23, 7.62}} \\
\midrule
\multicolumn{5}{l}{\textit{Autoregressive}} \\
$q_{\theta_W}(x_{t+1}\mid x_{0:t})$ & \metriccell{7.40}{6.41, 8.41} & \metriccell{10.33}{9.14, 11.55} & \metriccell{6.51}{5.61, 7.49} & \metriccell{9.25}{8.17, 10.39} \\
$q_{\theta_W}(x_{t+1}\mid x_{0:t},h_{t+1})$ & \metriccell{8.04}{6.99, 9.15} & \metriccell{11.47}{10.24, 12.80} & \metriccell{6.63}{5.81, 7.49} & \metriccell{9.31}{8.30, 10.34} \\
$q_{\theta_W}(x_{t+1}\mid x_{0:t},r_{t},h_{t+1})$\textsuperscript{$\dagger$} & \secondbestmetric{7.32}{6.44, 8.25} & \textbf{\metriccell{10.11}{9.08, 11.19}} & \textbf{\metriccell{6.31}{5.57, 7.09}} & \textbf{\metriccell{8.81}{7.91, 9.74}} \\
$q_{\theta_W}(x_{t+1}\mid x_{0:t}, r_{0:t},h_{t+1})$ & \textbf{\metriccell{7.20}{6.36, 8.06}} & \secondbestmetric{10.23}{9.21, 11.32} & \secondbestmetric{6.37}{5.63, 7.14} & \secondbestmetric{8.94}{8.06, 9.85} \\
\bottomrule
\end{tabular}
\vspace{3mm}
\end{subtable}

\begin{subtable}{\linewidth}
\centering
\footnotesize
\setlength{\tabcolsep}{3pt}
\caption{Effect of $q_{\theta_M}$ conditions on 20 ten-rater suspense stories.
$*$ marks direction accuracy significantly above no history (first row).
}
\label{tab:ablation-qm-human-history}
\begin{tabular}{p{0.35\linewidth}cccc}
\toprule
Experiment settings & Scaled MAE & Gap $>20$ (\%) & Gap $>30$ (\%) & Dir. (\%) \\
\midrule
$q_{\theta_M}( r_{t+1} \mid x_{0:t+1})$
  & \textbf{\metriccell{9.72}{8.30, 10.69}}
  & \metriccell{50.0}{39.4, 61.5}
  & \textbf{\metriccell{23.9}{14.4, 35.1}}
  & \metriccell{51.8}{45.9, 58.3} \\
$q_{\theta_M}( r_{t+1} \mid x_{0:t+1},r_{t})$
  & \secondbestmetric{9.86}{8.48, 10.82}
  & \secondbestmetric{46.3}{37.8, 55.2}
  & \secondbestmetric{26.1}{16.3, 37.4}
  & \textcolor{blue}{\metriccellstar{60.5}{52.9, 68.3}} \\
$q_{\theta_M}( r_{t+1} \mid x_{0:t+1}, r_{0:t})$\textsuperscript{$\dagger$}
  & \metriccell{9.96}{8.54, 10.93}
  & \textbf{\metriccell{39.6}{28.5, 51.5}}
  & \textbf{\metriccell{23.9}{14.0, 35.3}}
  & \textbf{\metriccellstar{65.8}{57.0, 74.3}} \\
\bottomrule
\end{tabular}

\end{subtable}

\end{table*}

\begin{table*}[htbp]
\centering
\caption{Humor conditioning ablations. Measurement-model rows use 20 ten-rater human-rated stories; generation rows use 514 stories and a GPT-5.5 writer. Lower error and raw gap rates are better; higher direction accuracy is better. Brackets give 95\% whole-story bootstrap confidence intervals. Bold and blue mark the best and second-best point estimates.}
\label{tab:humor-ablation-gpt55}

\begin{subtable}{\linewidth}
\centering
\footnotesize
\setlength{\tabcolsep}{3pt}
\renewcommand{\arraystretch}{0.95}
\caption{Steering-model $q_{\theta_S}$ input
}
\label{tab:humor-ablation-qs-gpt55}
\begin{tabular}{p{0.42\linewidth}cccc}
\toprule
& \multicolumn{2}{c}{Position-wise writing} & \multicolumn{2}{c}{Autoregressive writing} \\
\cmidrule(lr){2-3}\cmidrule(lr){4-5}
Experiment settings & Min-$K$ MAE & CRPS & Min-$K$ MAE & CRPS \\
\midrule
$q_{\theta_S}(h_{t+1}\mid x_{0:t})$ & \secondbestmetric{8.10}{7.41, 8.83} & \metriccell{12.95}{12.17, 13.74} & \metriccell{9.76}{9.08, 10.47} & \metriccell{11.93}{11.19, 12.70} \\
$q_{\theta_S}(h_{t+1}\mid x_{0:t}, r_t)$ & \metriccell{8.12}{7.46, 8.84} & \secondbestmetric{12.72}{11.97, 13.49} & \secondbestmetric{9.76}{9.06, 10.48} & \textbf{\metriccell{11.73}{11.00, 12.50}} \\
$q_{\theta_S}(h_{t+1}\mid x_{0:t},r_{0:t})$ & \textbf{\metriccell{7.98}{7.30, 8.68}} & \textbf{\metriccell{12.59}{11.81, 13.37}} & \textbf{\metriccell{9.73}{9.07, 10.43}} & \secondbestmetric{11.81}{11.09, 12.56} \\
\bottomrule
\end{tabular}
\end{subtable}

\par\vspace{3mm}
\begin{subtable}{\linewidth}
\centering
\footnotesize
\setlength{\tabcolsep}{3pt}
\renewcommand{\arraystretch}{0.95}
\caption{Effect of $q_{\theta_W}$ conditions.
The $q_{\theta_W}(x_{t+1}\mid x_{0:t})$ here is not Off-the-shelf baseline, it uses Off-the-shelf writer with steering-model-guided best-of-N selections.
}
\label{tab:humor-ablation-qw-gpt55}
\begin{tabular}{p{0.42\linewidth}cccc}
\toprule
& \multicolumn{2}{c}{Position-wise writing} & \multicolumn{2}{c}{Autoregressive writing} \\
\cmidrule(lr){2-3}\cmidrule(lr){4-5}
Experiment settings & Min-$K$ MAE & CRPS & Min-$K$ MAE & CRPS \\
\midrule
$q_{\theta_W}(x_{t+1}\mid x_{0:t})$ & \metriccell{9.49}{8.85, 10.17} & \metriccell{13.12}{12.41, 13.87} & \metriccell{11.52}{10.75, 12.32} & \metriccell{13.52}{12.74, 14.34} \\
$q_{\theta_W}(x_{t+1}\mid x_{0:t},h_{t+1})$ & \metriccell{8.87}{8.24, 9.55} & \metriccell{12.81}{12.12, 13.54} & \metriccell{10.73}{10.03, 11.49} & \metriccell{12.91}{12.16, 13.69} \\
$q_{\theta_W}(x_{t+1}\mid x_{0:t}, r_t,h_{t+1})$ & \secondbestmetric{7.98}{7.30, 8.68} & \secondbestmetric{12.59}{11.81, 13.37} & \secondbestmetric{9.73}{9.07, 10.43} & \secondbestmetric{11.81}{11.09, 12.56} \\
$q_{\theta_W}(x_{t+1}\mid x_{0:t}, r_{0:t},h_{t+1})$ & \textbf{\metriccell{7.78}{7.13, 8.45}} & \textbf{\metriccell{12.58}{11.82, 13.36}} & \textbf{\metriccell{9.44}{8.79, 10.13}} & \textbf{\metriccell{11.61}{10.90, 12.35}} \\
\bottomrule
\end{tabular}
\end{subtable}

\par\vspace{3mm}
\begin{subtable}{\linewidth}
\centering
\footnotesize
\setlength{\tabcolsep}{3pt}
\renewcommand{\arraystretch}{0.95}
\caption{Number of GPT-5.5 writer candidates
}
\label{tab:humor-ablation-candidates-gpt55}
\begin{tabular}{p{0.42\linewidth}cccc}
\toprule
& \multicolumn{2}{c}{Position-wise writing} & \multicolumn{2}{c}{Autoregressive writing} \\
\cmidrule(lr){2-3}\cmidrule(lr){4-5}
Experiment settings & Min-$K$ MAE & CRPS & Min-$K$ MAE & CRPS \\
\midrule
$N=1$ & \secondbestmetric{11.56}{10.77, 12.38} & \secondbestmetric{16.61}{15.73, 17.51} & \secondbestmetric{13.16}{12.34, 14.00} & \secondbestmetric{15.81}{14.95, 16.70} \\
$N=6$ & \textbf{\metriccell{7.98}{7.30, 8.68}} & \textbf{\metriccell{12.59}{11.81, 13.37}} & \textbf{\metriccell{9.73}{9.07, 10.43}} & \textbf{\metriccell{11.81}{11.09, 12.56}} \\
\bottomrule
\end{tabular}
\end{subtable}

\par\vspace{3mm}
\begin{subtable}{\linewidth}
\centering
\footnotesize
\setlength{\tabcolsep}{3pt}
\renewcommand{\arraystretch}{0.95}
\caption{Effect of $q_{\theta_M}$ conditions on 20 ten-rater suspense stories.
$*$ marks direction accuracy significantly above no history (first row).}
\label{tab:humor-ablation-qm-gpt55}
\begin{tabular}{p{0.35\linewidth}cccc}
\toprule
Experiment settings & Scaled MAE & Gap $>20$ (\%) & Gap $>30$ (\%) & Dir. (\%) \\
\midrule
$q_{\theta_M}( r_{t+1} \mid x_{0:t+1})$
  & \metriccell{8.89}{7.08, 10.22}
  & \metriccell{53.2}{45.9, 61.6}
  & \metriccell{23.4}{17.0, 30.4}
  & \metriccell{54.8}{43.8, 66.3} \\
$q_{\theta_M}( r_{t+1} \mid x_{0:t+1}, r_{t})$
  & \textbf{\metriccell{8.76}{7.08, 10.18}}
  & \textbf{\metriccellstar{34.7}{26.3, 43.8}}
  & \textbf{\metriccellstar{12.1}{5.8, 18.6}}
  & \textbf{\metriccell{63.5}{52.0, 74.2}} \\
$q_{\theta_M}( r_{t+1} \mid x_{0:t+1}, r_{0:t})$
  & \secondbestmetric{8.87}{7.13, 9.99}
  & \textcolor{blue}{\metriccellstar{37.1}{26.7, 48.4}}
  & \textcolor{blue}{\metriccellstar{13.7}{7.4, 20.9}}
  & \secondbestmetric{61.5}{49.0, 73.5} \\
\bottomrule
\end{tabular}
\end{subtable}

\end{table*}

\subsection{Tables for Main Results}
\label{app:main-results}

Tables~\ref{tab:dynamics-fidelity}, \ref{tab:story-quality-opus-guide-raw}, and \ref{tab:story-quality-suspense-opus-guide-raw} report the readerly dynamics fidelity and story quality metrics across two datasets, where the Min-K MAE and LLM-based story coherence metrics are visualized in Figure~\ref{fig:quality-fidelity-tradeoff-suspense}.

Tables~\ref{tab:story-quality-opus-guide-raw} and \ref{tab:story-quality-suspense-opus-guide-raw} present both the LLM-based evaluation results and the human evaluation results. Despite the limited sample size and lack of statistically significant differences after multiple-comparison correction, the human evaluations generally follow the trends observed in the LLM-based assessments, suggesting that the LLM judge broadly aligns with human preferences.

\begin{table*}[htbp]
\centering
\footnotesize
\setlength{\tabcolsep}{3pt}
\caption{Readerly dynamics fidelity of different methods, averaged across stories. Metrics are shown as means with 95\% bootstrap percentile intervals; lower is better. Bold marks each block's best metric, and blue marks its second-best metric.}
\label{tab:dynamics-fidelity}

\begin{subtable}{\linewidth}
\caption{
Humor dataset.
}
\label{tab:humor-fidelity-results}
\label{tab:humor-raw-prose-position}
\label{tab:humor-raw-prose-rollout}

\begin{tabular}{lcccccc}
\toprule
& \multicolumn{3}{c}{Qwen3.5-27B writer} & \multicolumn{3}{c}{GPT-5.5 writer} \\
\cmidrule(lr){2-4}\cmidrule(lr){5-7}
Method & Min-$K$ MAE & CRPS & Energy distance & Min-$K$ MAE & CRPS & Energy distance \\
\midrule
\multicolumn{7}{l}{\textit{Position-wise writing}} \\
SFT & \textbf{\metriccell{2.69}{2.38, 3.03}} & \secondbestmetric{9.46}{8.99, 9.95} & \metricna & \metricna & \metricna & \metricna \\
Off-the-shelf LLM & \metriccell{5.94}{5.38, 6.52} & \metriccell{11.34}{10.65, 12.05} & \metricna & \secondbestmetric{12.06}{11.32, 12.86} & \secondbestmetric{16.42}{15.62, 17.26} & \metricna \\
One-shot ICL & \metriccell{6.14}{5.60, 6.72} & \metriccell{11.67}{10.99, 12.38} & \metricna & \metriccell{12.43}{11.69, 13.20} & \metriccell{16.83}{16.04, 17.66} & \metricna \\
SWAG w/o DPO & \metriccell{8.04}{7.36, 8.74} & \metriccell{13.36}{12.61, 14.12} & \metricna & \metriccell{13.77}{12.98, 14.60} & \metriccell{17.66}{16.82, 18.50} & \metricna \\
SWAG & \metriccell{9.60}{8.84, 10.38} & \metriccell{14.74}{13.89, 15.59} & \metricna & \metriccell{15.49}{14.61, 16.38} & \metriccell{19.24}{18.30, 20.17} & \metricna \\
Ours & \secondbestmetric{4.10}{3.57, 4.67} & \textbf{\metriccell{9.03}{8.39, 9.70}} & \metricna & \textbf{\metriccell{7.98}{7.30, 8.68}} & \textbf{\metriccell{12.59}{11.81, 13.37}} & \metricna \\
\midrule
\multicolumn{7}{l}{\textit{Autoregressive}} \\
SFT & \secondbestmetric{7.94}{7.42, 8.47} & \secondbestmetric{11.08}{10.54, 11.64} & \metriccell{19.95}{17.46, 22.92} & \metricna & \metricna & \metricna \\
Off-the-shelf LLM & \metriccell{8.97}{8.30, 9.67} & \metriccell{11.50}{10.76, 12.26} & \secondbestmetric{5.41}{4.36, 6.85} & \secondbestmetric{14.94}{14.08, 15.83} & \secondbestmetric{17.53}{16.64, 18.46} & \secondbestmetric{25.90}{23.31, 28.95} \\
One-shot ICL & \metriccell{9.67}{8.99, 10.37} & \metriccell{12.26}{11.51, 13.02} & \metriccell{7.99}{6.63, 9.77} & \metriccell{15.08}{14.21, 15.98} & \metriccell{17.82}{16.93, 18.72} & \metriccell{27.65}{24.96, 30.79} \\
SWAG w/o DPO & \metriccell{10.77}{10.01, 11.55} & \metriccell{13.69}{12.89, 14.55} & \metriccell{11.65}{9.86, 13.91} & \metriccell{16.08}{15.14, 17.03} & \metriccell{18.61}{17.68, 19.55} & \metriccell{30.35}{27.39, 33.65} \\
SWAG & \metriccell{11.79}{10.97, 12.61} & \metriccell{14.85}{13.96, 15.79} & \metriccell{16.32}{14.15, 18.81} & \metriccell{17.39}{16.43, 18.37} & \metriccell{19.83}{18.85, 20.81} & \metriccell{32.75}{29.80, 36.17} \\
Ours & \textbf{\metriccell{6.98}{6.47, 7.52}} &
\textbf{\metriccell{8.81}{8.23, 9.41}} & \textbf{\metriccell{0.67}{0.55, 1.06}} & \textbf{\metriccell{9.73}{9.07, 10.43}} & \textbf{\metriccell{11.81}{11.09, 12.56}} & \textbf{\metriccell{7.43}{6.13, 9.07}} \\
\bottomrule
\end{tabular}
\medskip
\end{subtable}

\begin{subtable}{\linewidth}
\caption{Suspense dataset.}
\label{tab:compare-methods-in-suspense-beat}
\label{tab:swag-qwen-beat-results}

\begin{tabular}{lcccccc}
\toprule
& \multicolumn{3}{c}{Qwen3.5-27B writer} & \multicolumn{3}{c}{GPT-5.5 writer} \\
\cmidrule(lr){2-4}\cmidrule(lr){5-7}
Method & Min-$K$ MAE & CRPS & Energy distance & Min-$K$ MAE & CRPS & Energy distance \\
\midrule
\multicolumn{7}{l}{\textit{Position-wise writing}} \\
SFT & \textbf{\metriccell{2.17}{1.71, 2.67}} & \textbf{\metriccell{6.04}{5.36, 6.80}} & -- & \metricna & \metricna & \metricna \\
One-shot ICL & \metriccell{4.33}{3.50, 5.28} & \metriccell{8.41}{7.44, 9.45} & -- & \metriccell{4.26}{3.42, 5.20} & \metriccell{8.35}{7.29, 9.48} & -- \\
Off-the-shelf LLM & \metriccell{4.23}{3.41, 5.13} & \metriccell{8.53}{7.50, 9.62} & -- & \secondbestmetric{3.85}{3.13, 4.64} & \secondbestmetric{7.87}{6.93, 8.87} & -- \\
SWAG & \metriccell{8.85}{7.50, 10.31} & \metriccell{14.05}{12.36, 15.83} & -- & \metriccell{7.22}{6.15, 8.34} & \metriccell{11.35}{10.09, 12.66} & -- \\
SWAG w/o DPO & \metriccell{6.31}{5.32, 7.41} & \metriccell{11.05}{9.73, 12.42} & -- & \metriccell{5.26}{4.48, 6.08} & \metriccell{9.28}{8.30, 10.26} & -- \\
Ours & \secondbestmetric{3.60}{3.02, 4.23} & \secondbestmetric{8.05}{7.26, 8.88} & -- & \textbf{\metriccell{3.19}{2.67, 3.75}} & \textbf{\metriccell{7.09}{6.36, 7.85}} & -- \\
\midrule
\multicolumn{7}{l}{\textit{Autoregressive}} \\
SFT & \textbf{\metriccell{5.50}{4.79, 6.24}} & \textbf{\metriccell{7.78}{6.96, 8.61}} & \textbf{\metriccell{0.32}{0.34, 1.06}} & \metricna & \metricna & \metricna \\
One-shot ICL & \metriccell{8.69}{7.54, 9.88} & \metriccell{11.52}{10.20, 12.87} & \metriccell{6.96}{4.75, 10.34} & \secondbestmetric{7.26}{6.30, 8.27} & \metriccell{10.10}{8.94, 11.30} & \metriccell{3.71}{2.38, 6.09} \\
Off-the-shelf LLM & \metriccell{8.30}{7.19, 9.47} & \metriccell{11.49}{10.19, 12.87} & \metriccell{5.90}{3.95, 9.04} & \metriccell{7.34}{6.33, 8.37} & \secondbestmetric{9.89}{8.70, 11.12} & \metriccell{4.18}{2.64, 6.67} \\
SWAG & \metriccell{16.10}{14.14, 18.14} & \metriccell{21.38}{19.09, 23.80} & \metriccell{28.13}{20.04, 38.96} & \metriccell{12.64}{10.98, 14.38} & \metriccell{16.66}{14.77, 18.69} & \metriccell{10.93}{7.03, 17.02} \\
SWAG w/o DPO & \metriccell{9.94}{8.65, 11.31} & \metriccell{13.31}{11.71, 15.00} & \metriccell{2.85}{1.75, 5.23} & \metriccell{8.57}{7.42, 9.79} & \metriccell{11.80}{10.45, 13.19} & \textbf{\metriccell{1.14}{0.79, 2.57}} \\
Ours & \secondbestmetric{7.32}{6.44, 8.25} & \secondbestmetric{10.11}{9.08, 11.19} & \secondbestmetric{1.00}{0.80, 2.10} & \textbf{\metriccell{6.31}{5.57, 7.09}} & \textbf{\metriccell{8.81}{7.91, 9.74}} & \secondbestmetric{2.34}{1.37, 4.20} \\
\bottomrule

\end{tabular}

\end{subtable}

\end{table*}

\begin{table*}[htbp]
\centering
\scriptsize
\setlength{\tabcolsep}{4pt}
\caption{
Story quality evaluations for autoregressive outputs
and human references on the humor dataset. 
Scores are medians of expected values on a 1--5 scale with 95\% bootstrap intervals; higher is better. 
The final column of the LLM-scoring table gives mean word count per story and is not a quality score.
}
\label{tab:story-quality-opus-guide-raw}
\begin{subtable}{\linewidth}
\centering
\caption{LLM-ratings on the test split of humor dataset.}
\label{tab:humor-rollout-quality-results-opus-guide-raw}
\begin{tabular}{lccccccc}
\toprule
Method & Relevance & Coherence & Empathy & Surprise & Engagement & Complexity & Words \\
\midrule
\textit{Qwen3.5-27B as writer}\\
Off-the-shelf LLM & \scorecell{4.31}{4.27, 4.34} & \scorecell{3.68}{3.65, 3.70} & \scorecell{2.61}{2.58, 2.65} & \scorecell{2.41}{2.39, 2.43} & \scorecell{2.93}{2.91, 2.96} & \scorecell{2.06}{2.04, 2.07} & 318.2 \\
One-shot ICL & \scorecell{4.35}{4.31, 4.38} & \scorecell{3.62}{3.59, 3.65} & \scorecell{2.57}{2.54, 2.61} & \scorecell{2.44}{2.42, 2.46} & \scorecell{2.93}{2.91, 2.96} & \scorecell{2.05}{2.03, 2.06} & 322.6 \\
SFT & \scorecell{3.95}{3.92, 3.99} & \scorecell{2.59}{2.55, 2.63} & \scorecell{2.15}{2.12, 2.18} & \scorecell{1.98}{1.95, 2.00} & \scorecell{2.05}{2.02, 2.08} & \scorecell{1.62}{1.60, 1.65} & 258.4 \\
SWAG & \scorecell{4.16}{4.13, 4.20} & \scorecell{3.52}{3.49, 3.55} & \scorecell{2.46}{2.43, 2.49} & \scorecell{2.46}{2.44, 2.48} & \scorecell{2.92}{2.90, 2.94} & \scorecell{2.02}{2.00, 2.03} & 312.2 \\
SWAG w/o DPO & \scorecell{4.13}{4.09, 4.16} & \scorecell{3.48}{3.44, 3.51} & \scorecell{2.49}{2.46, 2.52} & \scorecell{2.42}{2.40, 2.44} & \scorecell{2.86}{2.83, 2.88} & \scorecell{2.03}{2.02, 2.05} & 311.6 \\
Ours & \scorecell{4.15}{4.11, 4.19} & \scorecell{3.59}{3.57, 3.62} & \scorecell{2.66}{2.63, 2.69} & \scorecell{2.41}{2.39, 2.43} & \scorecell{2.85}{2.82, 2.87} & \scorecell{2.04}{2.02, 2.05} & 315.2 \\
\midrule
\textit{GPT-5.5 as writer}\\
Off-the-shelf LLM & \scorecell{4.89}{4.87, 4.90} & \scorecell{4.53}{4.49, 4.56} & \scorecell{2.84}{2.80, 2.88} & \scorecell{2.76}{2.73, 2.78} & \scorecell{3.66}{3.63, 3.70} & \scorecell{2.38}{2.35, 2.42} & 288.2 \\
One-shot ICL & \scorecell{4.92}{4.90, 4.93} & \scorecell{4.53}{4.50, 4.57} & \scorecell{2.86}{2.82, 2.90} & \scorecell{2.83}{2.80, 2.85} & \scorecell{3.69}{3.66, 3.73} & \scorecell{2.41}{2.38, 2.45} & 291.0 \\
SWAG & \scorecell{4.86}{4.84, 4.88} & \scorecell{4.46}{4.43, 4.49} & \scorecell{2.76}{2.72, 2.80} & \scorecell{2.76}{2.73, 2.78} & \scorecell{3.68}{3.65, 3.71} & \scorecell{2.36}{2.33, 2.40} & 288.5 \\
SWAG w/o DPO & \scorecell{4.87}{4.85, 4.89} & \scorecell{4.42}{4.39, 4.45} & \scorecell{2.73}{2.69, 2.77} & \scorecell{2.75}{2.73, 2.78} & \scorecell{3.65}{3.61, 3.68} & \scorecell{2.37}{2.34, 2.40} & 286.9 \\
Ours & \scorecell{4.85}{4.83, 4.88} & \scorecell{4.55}{4.52, 4.58} & \scorecell{2.90}{2.85, 2.94} & \scorecell{2.70}{2.67, 2.72} & \scorecell{3.59}{3.56, 3.62} & \scorecell{2.35}{2.31, 2.38} & 286.2 \\
\midrule
Human & \scorecell{4.60}{4.55, 4.65} & \scorecell{3.41}{3.35, 3.48} & \scorecell{2.43}{2.38, 2.48} & \scorecell{2.78}{2.72, 2.84} & \scorecell{2.82}{2.76, 2.87} & \scorecell{1.98}{1.94, 2.01} & 259.0 \\
\bottomrule
\end{tabular}
\end{subtable}

\medskip

\begin{subtable}{\linewidth}
\centering
\caption{Human ratings on 8 humor story openings. GPT-5.5 as writer.}
\label{tab:labelbox-hanna-humor}
\begin{tabular}{@{}lcccccc@{}}
\toprule
Method & Relevance & Coherence & Empathy & Surprise & Engagement & Complexity \\
\midrule
Off-the-shelf LLM
& \scorecell{4.38}{3.67, 4.92}
& \scorecell{4.21}{3.42, 4.83}
& \scorecell{3.71}{2.83, 4.46}
& \scorecell{2.08}{1.62, 2.58}
& \scorecell{3.46}{2.83, 4.04}
& \scorecell{2.79}{2.25, 3.29} \\
One-shot ICL
& \scorecell{4.54}{4.17, 4.88}
& \scorecell{3.96}{3.50, 4.42}
& \scorecell{3.21}{2.58, 3.83}
& \scorecell{2.21}{1.75, 2.67}
& \scorecell{3.58}{3.04, 4.08}
& \scorecell{2.92}{2.46, 3.42} \\
SWAG w/o DPO
& \scorecell{4.38}{3.88, 4.79}
& \scorecell{3.79}{3.29, 4.33}
& \scorecell{3.00}{2.29, 3.75}
& \scorecell{2.92}{2.46, 3.38}
& \scorecell{3.25}{2.54, 3.88}
& \scorecell{2.62}{2.29, 2.96} \\
Ours
& \scorecell{4.46}{4.04, 4.88}
& \scorecell{4.50}{4.29, 4.75}
& \scorecell{3.38}{2.71, 3.88}
& \scorecell{2.83}{2.25, 3.42}
& \scorecell{3.38}{2.96, 3.79}
& \scorecell{3.29}{2.71, 3.75} \\
Human
& \scorecell{4.12}{3.67, 4.54}
& \scorecell{3.71}{3.12, 4.25}
& \scorecell{2.75}{2.21, 3.33}
& \scorecell{2.50}{2.00, 2.96}
& \scorecell{2.75}{2.29, 3.25}
& \scorecell{2.67}{2.29, 3.00} \\
\bottomrule
\end{tabular}
\end{subtable}

\medskip
\end{table*}

\begin{table*}[htbp]
\centering
\scriptsize
\setlength{\tabcolsep}{4pt}
\caption{Story quality evaluations for autoregressive outputs
and human references on the suspense dataset.
Scores are medians of expected values on a 1--5 scale with 95\% bootstrap intervals; higher is better.
The final column of the LLM-scoring table gives mean word count per story and is not a quality score.}
\label{tab:story-quality-suspense-opus-guide-raw}
\begin{subtable}{\linewidth}
\centering
\caption{LLM-ratings on the test split of suspense dataset.}
\label{tab:rollout-beat-quality-results-opus-guide-raw}
\begin{tabular}{lccccccc}
\toprule
Method & Relevance & Coherence & Empathy & Surprise & Engagement & Complexity & Words \\
\midrule
\textit{Qwen3.5-27B as writer}\\
Off-the-shelf LLM & \scorecell{4.49}{4.44, 4.55} & \scorecell{3.90}{3.84, 3.96} & \scorecell{2.95}{2.90, 3.00} & \scorecell{2.13}{2.08, 2.19} & \scorecell{2.65}{2.61, 2.70} & \scorecell{2.15}{2.11, 2.19} & 257.1 \\
One-shot ICL & \scorecell{4.47}{4.42, 4.53} & \scorecell{3.87}{3.81, 3.94} & \scorecell{2.89}{2.84, 2.94} & \scorecell{2.11}{2.05, 2.17} & \scorecell{2.59}{2.54, 2.63} & \scorecell{2.11}{2.07, 2.15} & 250.4 \\
SFT & \scorecell{4.27}{4.22, 4.32} & \scorecell{3.34}{3.28, 3.41} & \scorecell{2.73}{2.67, 2.79} & \scorecell{2.40}{2.34, 2.46} & \scorecell{2.49}{2.44, 2.53} & \scorecell{2.17}{2.12, 2.21} & 249.1 \\
SWAG & \scorecell{4.08}{4.01, 4.15} & \scorecell{3.59}{3.53, 3.65} & \scorecell{2.89}{2.84, 2.94} & \scorecell{2.33}{2.26, 2.39} & \scorecell{2.66}{2.61, 2.71} & \scorecell{2.14}{2.10, 2.18} & 262.2 \\
Ours & \scorecell{4.36}{4.30, 4.42} & \scorecell{3.75}{3.68, 3.81} & \scorecell{2.95}{2.91, 3.00} & \scorecell{2.23}{2.17, 2.29} & \scorecell{2.67}{2.63, 2.71} & \scorecell{2.13}{2.09, 2.16} & 253.8 \\
\midrule
\textit{GPT-5.5 as writer}\\
Off-the-shelf LLM & \scorecell{4.82}{4.78, 4.86} & \scorecell{4.56}{4.51, 4.61} & \scorecell{3.14}{3.07, 3.20} & \scorecell{2.41}{2.34, 2.48} & \scorecell{3.09}{3.03, 3.15} & \scorecell{2.60}{2.53, 2.68} & 251.0 \\
One-shot ICL & \scorecell{4.80}{4.75, 4.84} & \scorecell{4.56}{4.50, 4.61} & \scorecell{3.11}{3.05, 3.17} & \scorecell{2.34}{2.27, 2.42} & \scorecell{3.01}{2.95, 3.07} & \scorecell{2.55}{2.47, 2.62} & 252.1 \\
SWAG & \scorecell{4.74}{4.69, 4.79} & \scorecell{4.34}{4.29, 4.40} & \scorecell{3.20}{3.14, 3.25} & \scorecell{2.64}{2.56, 2.73} & \scorecell{3.26}{3.19, 3.32} & \scorecell{2.72}{2.65, 2.79} & 263.8 \\
SWAG w/o DPO & \scorecell{4.76}{4.71, 4.80} & \scorecell{4.41}{4.35, 4.46} & \scorecell{3.18}{3.12, 3.24} & \scorecell{2.56}{2.48, 2.64} & \scorecell{3.16}{3.10, 3.22} & \scorecell{2.68}{2.60, 2.75} & 261.0 \\
Ours & \scorecell{4.81}{4.76, 4.84} & \scorecell{4.53}{4.48, 4.58} & \scorecell{3.18}{3.11, 3.24} & \scorecell{2.43}{2.36, 2.50} & \scorecell{3.11}{3.05, 3.16} & \scorecell{2.64}{2.57, 2.71} & 258.6 \\
\midrule
Human & \scorecell{4.80}{4.73, 4.85} & \scorecell{4.16}{4.06, 4.26} & \scorecell{3.03}{2.95, 3.12} & \scorecell{2.70}{2.60, 2.82} & \scorecell{2.97}{2.89, 3.05} & \scorecell{2.50}{2.41, 2.59} & 248.0 \\
\bottomrule
\end{tabular}
\end{subtable}

\medskip

\begin{subtable}{\linewidth}
\centering
\caption{Human ratings on 8 suspense story openings. GPT-5.5 as writer.}
\label{tab:labelbox-hanna-suspense}
\begin{tabular}{@{}lcccccc@{}}
\toprule
Method & Relevance & Coherence & Empathy & Surprise & Engagement & Complexity \\
\midrule
Off-the-shelf LLM
& \scorecell{4.71}{4.50, 4.92}
& \scorecell{4.42}{4.12, 4.71}
& \scorecell{4.12}{3.75, 4.50}
& \scorecell{3.38}{2.96, 3.79}
& \scorecell{3.54}{3.33, 3.75}
& \scorecell{3.58}{3.21, 4.00} \\
One-shot ICL
& \scorecell{4.42}{3.58, 4.92}
& \scorecell{4.29}{3.87, 4.67}
& \scorecell{3.75}{3.17, 4.17}
& \scorecell{3.42}{2.92, 3.83}
& \scorecell{3.54}{3.12, 3.88}
& \scorecell{3.58}{3.08, 4.08} \\
SWAG w/o DPO
& \scorecell{4.79}{4.62, 4.96}
& \scorecell{4.58}{4.33, 4.79}
& \scorecell{4.12}{3.62, 4.58}
& \scorecell{3.83}{3.42, 4.21}
& \scorecell{3.88}{3.50, 4.25}
& \scorecell{3.54}{3.04, 4.04} \\
Ours
& \scorecell{4.25}{3.25, 4.88}
& \scorecell{4.33}{3.79, 4.79}
& \scorecell{4.25}{3.75, 4.54}
& \scorecell{3.38}{3.12, 3.62}
& \scorecell{3.67}{3.17, 4.08}
& \scorecell{3.96}{3.62, 4.33} \\
Human
& \scorecell{4.75}{4.42, 5.00}
& \scorecell{4.58}{4.38, 4.79}
& \scorecell{4.17}{3.62, 4.67}
& \scorecell{3.52}{3.00, 4.02}
& \scorecell{3.79}{3.46, 4.12}
& \scorecell{3.27}{2.71, 3.81} \\
\bottomrule
\end{tabular}
\end{subtable}

\end{table*}

The following table~\ref{tab:seq9-human-reference-mae} presents a small-scale evaluation of reader dynamics fidelity across generation methods, using human-annotated reader response scores to compute MAE. Due to the limited annotation budget, this analysis covers only 8 story openings and is intended to illustrate general trends rather than support statistically rigorous conclusions. Note that the human ratings use a 1--9 scale, consistent with our user study, rather than the 0--100 scale used by our measure model.
\begin{table}[htbp]
\centering
\scriptsize
\setlength{\tabcolsep}{5pt}
\caption{Human-annotated Min-K MAE on the 1--9 rating scale, using 8 stories. Lower is better.}
\label{tab:seq9-human-reference-mae}
\begin{tabular}{lccc}
\toprule
Method & MAE $\downarrow$ & $\Delta$ to Off-the-shelf $\downarrow$ & $p$ \\
\midrule
\multicolumn{4}{l}{\textit{Humor}} \\
Off-the-shelf LLM & \metriccell{1.44}{0.90, 2.03} & -- & -- \\
Ours & \metriccell{0.95}{0.66, 1.19} & \metriccell{-0.49}{-1.22, +0.18} & 0.25 \\
One-shot ICL & \metriccell{1.48}{0.95, 2.08} & \metriccell{+0.04}{-0.62, +0.64} & 0.91 \\
SWAG w/o DPO & \metriccell{1.34}{0.88, 1.84} & \metriccell{-0.10}{-0.86, +0.64} & 0.80 \\
\midrule
\multicolumn{4}{l}{\textit{Suspense}} \\
Off-the-shelf LLM & \metriccell{0.92}{0.57, 1.32} & -- & -- \\
Ours & \metriccell{1.10}{0.82, 1.39} & \metriccell{+0.18}{-0.20, +0.53} & 0.41 \\
One-shot ICL & \metriccell{0.88}{0.76, 1.00} & \metriccell{-0.04}{-0.40, +0.27} & 0.86 \\
SWAG w/o DPO & \metriccell{1.25}{0.58, 2.06} & \metriccell{+0.33}{-0.54, +1.36} & 0.57 \\
\bottomrule
\end{tabular}
\end{table}

\subsection{Story Quality Evaluation}
\label{app:quality-eval}

We use Claude Opus 5 with a participant-guide prompt to evaluate
story quality across all six HANNA dimensions: relevance, coherence,
empathy, surprise, engagement, and complexity. The prompt presents
the story opening and continuation together with the participant-facing 1--5 scoring rubric. The model assigns an integer score to each
dimension and provides one sentence of supporting textual evidence.
The opening serves only as context for judging the continuation and
does not itself receive quality credit or penalties. All six
dimensions are evaluated jointly using the same model and prompt,
without dimension-specific model selection or score calibration.

\iflatexml
  \begin{HtmlPrompt}{System message of story quality score LLM}
\begin{HtmlPromptText}
You are an independent story-quality rater. Follow the supplied participant-facing HANNA guide. Rate ONLY the portion tagged 'Evaluation starts here' and later. Earlier segments are story-opening context; do not give their quality credit or penalties except when judging how relevant the continuation is to that opening. Apply the six separate 1-to-5 criteria in the guide. Do not infer a method, source, label, title, or external reference story. Return only strict JSON with exactly two top-level keys, scores and evidence. scores must contain exactly relevance, coherence, empathy, surprise, engagement, complexity as integer values 1 through 5. evidence must contain the same six keys with one concise sentence grounded in the displayed story for each score.
\end{HtmlPromptText}
\end{HtmlPrompt}

\else
  \begin{PromptVerbatim}{System message of story quality score LLM}
You are an independent story-quality rater. Follow the supplied participant-facing HANNA guide. Rate ONLY the portion tagged 'Evaluation starts here' and later. Earlier segments are story-opening context; do not give their quality credit or penalties except when judging how relevant the continuation is to that opening. Apply the six separate 1-to-5 criteria in the guide. Do not infer a method, source, label, title, or external reference story. Return only strict JSON with exactly two top-level keys, scores and evidence. scores must contain exactly relevance, coherence, empathy, surprise, engagement, complexity as integer values 1 through 5. evidence must contain the same six keys with one concise sentence grounded in the displayed story for each score.
\end{PromptVerbatim}

\fi

\iflatexml
  \begin{HtmlPrompt}{User message example of story quality score LLM}
\begin{HtmlPromptText}
PARTICIPANT-FACING SCORING GUIDE (apply it exactly):\newline
\# Story rating\newline
\strut\newline
Read the story opening and the rest of the story. The story is shown as short segment summaries, S01 to Sn, all visible in order.\newline
\strut\newline
The story opening is marked on the page by the “Story start” tag and a different background color. The rest of the story continues from that opening.\newline
\strut\newline
> **Rate only the part of the story after the story opening. Evaluation starts at the segment tagged “Evaluation starts here”, right after the last “Story start” segment. The story opening is context only and is not rated: every criterion below applies to the part after it (for Relevance, how well that part matches the story opening).**\newline
\strut\newline
Note: if the story is not relevant with respect to the story opening, it only affects the Relevance criterion! Do not rate 1 everywhere.\newline
\strut\newline
Some stories may contain explicit or offensive content.\newline
\strut\newline
Please rate the part after the story opening on a scale from 1 (worst) to 5 (best) on the following criteria: relevance, coherence, empathy, surprise, engagement, and complexity.\newline
\strut\newline
\#\# Relevance — measures how well the story matches the story opening\newline
\strut\newline
1. The story has no relationship with the story opening at all.\newline
2. The story only has a weak relationship with the story opening.\newline
3. The story roughly matches the story opening.\newline
4. The story matches the story opening, except for one or two small aspects.\newline
5. The story matches the story opening exactly.\newline
\strut\newline
\#\# Coherence — measures whether the story makes sense\newline
\strut\newline
1. The story does not make sense at all. For instance, the setting and/or characters keep changing, and/or there is no understandable plot.\newline
2. Most of the story does not make sense.\newline
3. The story mostly makes sense but has some incoherences.\newline
4. The story almost makes sense overall, except for one or two small incoherences.\newline
5. The story makes sense from beginning to end.\newline
\strut\newline
\#\# Empathy — measures how well you understood the characters' emotions (regardless of whether you agreed with them)\newline
\strut\newline
1. The characters seemed apathetic to you.\newline
2. At least one character slightly related to you on an emotional level.\newline
3. You recognized specific, but not necessarily strong, emotions (eg sadness, joy, fear…) in at least one character.\newline
4. At least one character emotionally involved you, but minor details prevented you from completely relating to them.\newline
5. At least one character completely involved you on an emotional level.\newline
\strut\newline
\#\# Surprise — measures how surprising the end of the story was\newline
\strut\newline
1. The ending seemed completely obvious from the start, or doesn't make any sense at all.\newline
2. The ending was easily predictable after a few sentences.\newline
3. The ending was predictable after half of the story.\newline
4. The ending surprised you, but would have been difficult to predict.\newline
5. The ending surprised you, and still seemed as if it could very reasonably have been predicted, ie, there were enough clues in the story.\newline
\strut\newline
\#\# Engagement — measures how much you engaged with the story\newline
\strut\newline
1. You found the story boring and were glad it was over.\newline
2. You found one or two things interesting in the story, but no more.\newline
3. The story was mildly interesting.\newline
4. The story almost kept you engaged until the end.\newline
5. You were so engaged that you wished there was a sequel.\newline
\strut\newline
\#\# Complexity — measures how elaborate the story is\newline
\strut\newline
1. The setting of the story is extremely simple; it only involves one or two characters or concepts.\newline
2. The setting of the story is simple; one or two characters, a simple plot, maybe an indication of time or location.\newline
3. The story is somewhat developed: it involves at least one of the following: complex concepts, realistic characters, an intricate plot, an underlying history or circumstances, precise descriptions.\newline
4. The story is developed: it involves at least two of the following: complex concepts, realistic characters, an intricate plot, an underlying history or circumstances, precise descriptions.\newline
5. The story is well thought-out: it involves at least three of the following: complex concepts, realistic characters, an intricate plot, an underlying history or circumstances, precise descriptions.\newline
\strut\newline
DISPLAYED STORY:\newline
S01 [Story start]\newline
- A king prepares a grand feast but lacks one specific fish, so he announces that anyone who brings it to the palace will be handsomely rewarded.\newline
\strut\newline
S02 [Story start]\newline
- Many fishermen try and fail to find the wanted fish; the day before the feast, one fisherman arrives with the big fish.\newline
\strut\newline
S03 [Story start]\newline
- At the palace gate, the gatekeeper asks the fisherman why he has come.\newline
- When the fisherman mentions the fish and reward, the gatekeeper says he may enter only if he promises half the reward.\newline
\strut\newline
S04 [Story start]\newline
- The fisherman agreed to split the reward; the king, delighted by the fish, offered him much money, but the fisherman refused.\newline
\strut\newline
S05 [Evaluation starts here]\newline
- The fisherman demanded a hundred lashes on his back instead of money; the surprised king agreed after he insisted.\newline
- The king ordered a servant to whip him, but after fifty lashes the fisherman asked them to stop.\newline
\strut\newline
S06\newline
- The fisherman says he promised to share half his reward, then identifies the palace gatekeeper who demanded half for entry.\newline
- The king understands; the gatekeeper receives the remaining fifty lashes and is dismissed, while the fisherman is richly rewarded for his work and intelligence.\newline
\strut\newline
Return the six 1–5 scores and evidence now.
\end{HtmlPromptText}
\end{HtmlPrompt}

\else
  \begin{PromptVerbatim}{User message example of story quality score LLM}
PARTICIPANT-FACING SCORING GUIDE (apply it exactly):
# Story rating

Read the story opening and the rest of the story. The story is shown as short segment summaries, S01 to Sn, all visible in order.

The story opening is marked on the page by the “Story start” tag and a different background color. The rest of the story continues from that opening.

> **Rate only the part of the story after the story opening. Evaluation starts at the segment tagged “Evaluation starts here”, right after the last “Story start” segment. The story opening is context only and is not rated: every criterion below applies to the part after it (for Relevance, how well that part matches the story opening).**

Note: if the story is not relevant with respect to the story opening, it only affects the Relevance criterion! Do not rate 1 everywhere.

Some stories may contain explicit or offensive content.

Please rate the part after the story opening on a scale from 1 (worst) to 5 (best) on the following criteria: relevance, coherence, empathy, surprise, engagement, and complexity.

## Relevance — measures how well the story matches the story opening

1. The story has no relationship with the story opening at all.
2. The story only has a weak relationship with the story opening.
3. The story roughly matches the story opening.
4. The story matches the story opening, except for one or two small aspects.
5. The story matches the story opening exactly.

## Coherence — measures whether the story makes sense

1. The story does not make sense at all. For instance, the setting and/or characters keep changing, and/or there is no understandable plot.
2. Most of the story does not make sense.
3. The story mostly makes sense but has some incoherences.
4. The story almost makes sense overall, except for one or two small incoherences.
5. The story makes sense from beginning to end.

## Empathy — measures how well you understood the characters' emotions (regardless of whether you agreed with them)

1. The characters seemed apathetic to you.
2. At least one character slightly related to you on an emotional level.
3. You recognized specific, but not necessarily strong, emotions (eg sadness, joy, fear…) in at least one character.
4. At least one character emotionally involved you, but minor details prevented you from completely relating to them.
5. At least one character completely involved you on an emotional level.

## Surprise — measures how surprising the end of the story was

1. The ending seemed completely obvious from the start, or doesn't make any sense at all.
2. The ending was easily predictable after a few sentences.
3. The ending was predictable after half of the story.
4. The ending surprised you, but would have been difficult to predict.
5. The ending surprised you, and still seemed as if it could very reasonably have been predicted, ie, there were enough clues in the story.

## Engagement — measures how much you engaged with the story

1. You found the story boring and were glad it was over.
2. You found one or two things interesting in the story, but no more.
3. The story was mildly interesting.
4. The story almost kept you engaged until the end.
5. You were so engaged that you wished there was a sequel.

## Complexity — measures how elaborate the story is

1. The setting of the story is extremely simple; it only involves one or two characters or concepts.
2. The setting of the story is simple; one or two characters, a simple plot, maybe an indication of time or location.
3. The story is somewhat developed: it involves at least one of the following: complex concepts, realistic characters, an intricate plot, an underlying history or circumstances, precise descriptions.
4. The story is developed: it involves at least two of the following: complex concepts, realistic characters, an intricate plot, an underlying history or circumstances, precise descriptions.
5. The story is well thought-out: it involves at least three of the following: complex concepts, realistic characters, an intricate plot, an underlying history or circumstances, precise descriptions.

DISPLAYED STORY:
S01 [Story start]
- A king prepares a grand feast but lacks one specific fish, so he announces that anyone who brings it to the palace will be handsomely rewarded.

S02 [Story start]
- Many fishermen try and fail to find the wanted fish; the day before the feast, one fisherman arrives with the big fish.

S03 [Story start]
- At the palace gate, the gatekeeper asks the fisherman why he has come.
- When the fisherman mentions the fish and reward, the gatekeeper says he may enter only if he promises half the reward.

S04 [Story start]
- The fisherman agreed to split the reward; the king, delighted by the fish, offered him much money, but the fisherman refused.

S05 [Evaluation starts here]
- The fisherman demanded a hundred lashes on his back instead of money; the surprised king agreed after he insisted.
- The king ordered a servant to whip him, but after fifty lashes the fisherman asked them to stop.

S06
- The fisherman says he promised to share half his reward, then identifies the palace gatekeeper who demanded half for entry.
- The king understands; the gatekeeper receives the remaining fifty lashes and is dismissed, while the fisherman is richly rewarded for his work and intelligence.

Return the six 1–5 scores and evidence now.
\end{PromptVerbatim}

\fi

\section{Method Details}
\label{app:method-details}

\subsection{Steering Model}
\label{app:steering-model}

The following example shows the system, user, and assistant messages for the suspense value condition.  Black denotes the shared prompt template, dark green denotes dynamics-specific instructions and fields, and blue denotes content supplied by specific training example.
See Table~\ref{tab:steering-model-config} for training and inference configurations.

\iflatexml
  \begin{HtmlPrompt}{System message}
\begin{HtmlPromptText}
You direct short stories one story beat at a time. Given only the story beats so far and the planned story length, predict the requested factor for the next beat. Preserve established characters and roles. Do not invent prose from later beats. \color{green!50!black}The user supplies the current overall suspense level. Predict the overall suspense of the story so far through the next beat as one integer from 0 (none) to 100 (maximum). This is the endpoint story-so-far rating after the next beat, not a within-beat peak. The director\_note field is a fixed transport placeholder, not a supervised narrative factor; copy it exactly. Return only valid YAML with exactly this schema and no code fence:\newline
director\_note: Use only the requested non-note factor for this beat.\newline
suspense:\newline
\hspace*{1em}level: <integer 0-100>\color{black}
\end{HtmlPromptText}
\end{HtmlPrompt}

\else
  \begin{PromptVerbatim}[commandchars=\\\{\}]{System message}
You direct short stories one story beat at a time. Given only the story beats so far and the planned story length, predict the requested factor for the next beat. Preserve established characters and roles. Do not invent prose from later beats. \color{green!50!black}The user supplies the current overall suspense level. Predict the overall suspense of the story so far through the next beat as one integer from 0 (none) to 100 (maximum). This is the endpoint story-so-far rating after the next beat, not a within-beat peak. The director_note field is a fixed transport placeholder, not a supervised narrative factor; copy it exactly. Return only valid YAML with exactly this schema and no code fence:
director_note: Use only the requested non-note factor for this beat.
suspense:
  level: <integer 0-100>\color{black}
\end{PromptVerbatim}

\fi

\iflatexml
  \begin{HtmlPrompt}{User message}
\begin{HtmlPromptText}
Genre: \color{blue}fable\color{black}.\newline
Complete story length: \color{blue}7\color{black} beats.\newline
Beats written so far: \color{blue}1\color{black}.\newline
You are directing beat \color{blue}2\color{black}; \color{blue}6\color{black} beat(s) remain including this one.\newline
\strut\newline
Story beats so far:\newline
[BEAT 1]\newline
\color{blue}- Two youths find themselves lost together in the desert and forced to spend the night without modern technology.\newline
- One youth fears being stranded among unknown dangers; the other declares the situation great.\color{black}\newline
\strut\newline
\color{green!50!black}Current overall suspense level:\color{blue} 34\color{black}\newline
\strut\newline
Return the requested YAML for beat \color{blue}2\color{black} only.
\end{HtmlPromptText}
\end{HtmlPrompt}

\else
  \begin{PromptVerbatim}[commandchars=\\\{\}]{User message}
Genre: \color{blue}fable\color{black}.
Complete story length: \color{blue}7\color{black} beats.
Beats written so far: \color{blue}1\color{black}.
You are directing beat \color{blue}2\color{black}; \color{blue}6\color{black} beat(s) remain including this one.

Story beats so far:
[BEAT 1]
\color{blue}- Two youths find themselves lost together in the desert and forced to spend the night without modern technology.
- One youth fears being stranded among unknown dangers; the other declares the situation great.\color{black}

\color{green!50!black}Current overall suspense level:\color{blue} 34\color{black}

Return the requested YAML for beat \color{blue}2\color{black} only.
\end{PromptVerbatim}

\fi

\iflatexml
  \begin{HtmlPrompt}{Assistant target}
\begin{HtmlPromptText}
\color{green!50!black}suspense:\newline
\hspace*{1em}\color{green!50!black}level:\color{blue} 30\color{black}
\end{HtmlPromptText}
\end{HtmlPrompt}

\else
  \begin{PromptVerbatim}[commandchars=\\\{\}]{Assistant target}
\color{green!50!black}suspense:
  \color{green!50!black}level:\color{blue} 30\color{black}
\end{PromptVerbatim}

\fi

\label{app:steering-config}

\begin{table}[htbp]
\centering
\footnotesize
\setlength{\tabcolsep}{2pt}
\renewcommand{\arraystretch}{1.05}
\caption{Training and inference configuration for the steering model
$q_{\theta_S}$. 
}
\label{tab:steering-model-config}
\begin{tabular}{@{}>{\raggedright\arraybackslash}p{0.55\columnwidth}>{\raggedright\arraybackslash}p{0.40\columnwidth}@{}}
\toprule
Setting & Value \\
\midrule
\multicolumn{2}{@{}l}{\textit{Model}} \\
Base model & Qwen3.5-27B \\
\multicolumn{2}{@{}l}{\textit{LoRA}} \\
Rank / scaling / dropout & 16 / 32 / 0.05 \\
\multicolumn{2}{@{}l}{\textit{Training}} \\
Epoch & 1 \\
Optimizer & AdamW \\
Learning rate / weight decay & $2\times10^{-4}$ / 0.01 \\
Schedule & 3\% warm-up followed by cosine decay \\
Effective batch size & 16 \\
Context window & 2{,}048 tokens \\
Random seed & 17 \\
\multicolumn{2}{@{}l}{\textit{Inference}} \\
Thinking & Disabled \\
Temperature / top-$p$ / top-$k$ & 1.0 / 0.9 / 20 \\
Maximum output length & 768 tokens \\
\bottomrule
\end{tabular}
\end{table}

We investigate how the representation of reader dynamics affects the performance of our steering model. Specifically, we retrain the Qwen3.5-27B controller on the same training and validation data, comparing two output formats: structured YAML output and a plain integer score ranging from 0 to 100. We evaluate both variants on the suspense dataset, with results summarized in Table~\ref{tab:representation-ablation}.
\begin{table}[htbp]
\centering
\small
\caption{Effect of reader dynamics representation on
steering model performance on the suspense dataset.
Lower values indicate better performance.}
\label{tab:representation-ablation}
\begin{tabular}{lcc}
\toprule
Output Format &
Min-K MAE $\downarrow$ &
CRPS $\downarrow$ \\
\midrule
Structured YAML & \metriccell{2.42}{2.03, 2.86} & \metriccell{6.70}{6.09, 7.37} \\
Plain Integer   & \metriccell{2.53}{2.17, 2.92} & \metriccell{7.09}{6.46, 7.76} \\
\bottomrule
\end{tabular}
\end{table}

As shown in Table~\ref{tab:representation-ablation}, changing the output representation from structured YAML to a plain integer does not yield a statistically significant difference in steering model performance. This suggests that the model's performance is relatively insensitive to the choice of output format in this experiment.

\subsection{Writer Model}
\label{app:our-writer}
The GPT writer is used with reasoning disabled and temperature 1.0.
The Qwen writer used with thinking
disabled and sampling at temperature 1.0, top-$p=0.95$, and top-$k=20$.
Both writers receive the same model-agnostic formatting instruction and permit at most 420 output tokens. A failed format check causes a resampling.

\iflatexml
  \begin{HtmlPrompt}{Example system message}
\begin{HtmlPromptText}
You continue short stories one story beat at a time. A story beat is a compact unit of narrative action. Given the story beats so far, write exactly one next story beat. Preserve the established voice, characters, situation, chronology, and causal continuity. Return only a JSON object with the single key "beat". The beat value must contain 1-3 bullet lines, each beginning with '- ', and must not contain a beat label or heading. Do not summarize, explain your choices, write past the next beat, or use a thinking block.\newline
\strut\newline
\color{green!50!black}STEERING SIGNAL INTERPRETATION:\newline
\color{green!50!black}When the user supplies structured direction for a beat, use it while writing that beat. Interpret only the YAML fields defined here: suspense.level gives the intended overall story-so-far suspense after the beat, not a within-beat peak. Do not quote, summarize, or mention the YAML in the story.\color{black}
\end{HtmlPromptText}
\end{HtmlPrompt}

\else
  \begin{PromptVerbatim}[commandchars=\\\{\}]{Example system message}
You continue short stories one story beat at a time. A story beat is a compact unit of narrative action. Given the story beats so far, write exactly one next story beat. Preserve the established voice, characters, situation, chronology, and causal continuity. Return only a JSON object with the single key "beat". The beat value must contain 1-3 bullet lines, each beginning with '- ', and must not contain a beat label or heading. Do not summarize, explain your choices, write past the next beat, or use a thinking block.

\color{green!50!black}STEERING SIGNAL INTERPRETATION:
\color{green!50!black}When the user supplies structured direction for a beat, use it while writing that beat. Interpret only the YAML fields defined here: suspense.level gives the intended overall story-so-far suspense after the beat, not a within-beat peak. Do not quote, summarize, or mention the YAML in the story.\color{black}
\end{PromptVerbatim}

\fi

\iflatexml
  \begin{HtmlPrompt}{Example user message}
\begin{HtmlPromptText}
Genre: \color{blue}fable\color{black}.\newline
Complete story length: \color{blue}7\color{black} beats.\newline
Beats written so far: \color{blue}1\color{black}.\newline
You are writing beat \color{blue}2\color{black}; \color{blue}6\color{black} beat(s) remain including this one.\newline
Length reference for this beat: \color{blue}15\color{black} words.\newline
\strut\newline
Story beats so far:\newline
[BEAT 1]\newline
\color{blue}- Two youths find themselves lost together in the desert and forced to spend the night without modern technology.\newline
- One youth fears being stranded among unknown dangers; the other declares the situation great.\color{black}\newline
\strut\newline
\color{green!50!black}STRUCTURED DIRECTION FOR THIS BEAT ONLY:\newline
\color{green!50!black}suspense:\newline
\hspace*{1em}\color{green!50!black}level: \color{blue}30\color{black}\newline
\strut\newline
Write beat \color{blue}2\color{black} only. Return only \textbraceleft{}"beat": "- ...\textbackslash{}n- ..."\textbraceright{}.
\end{HtmlPromptText}
\end{HtmlPrompt}

\else
  \begin{PromptVerbatim}[commandchars=\\\{\}]{Example user message}
Genre: \color{blue}fable\color{black}.
Complete story length: \color{blue}7\color{black} beats.
Beats written so far: \color{blue}1\color{black}.
You are writing beat \color{blue}2\color{black}; \color{blue}6\color{black} beat(s) remain including this one.
Length reference for this beat: \color{blue}15\color{black} words.

Story beats so far:
[BEAT 1]
\color{blue}- Two youths find themselves lost together in the desert and forced to spend the night without modern technology.
- One youth fears being stranded among unknown dangers; the other declares the situation great.\color{black}

\color{green!50!black}STRUCTURED DIRECTION FOR THIS BEAT ONLY:
\color{green!50!black}suspense:
  \color{green!50!black}level: \color{blue}30\color{black}

Write beat \color{blue}2\color{black} only. Return only \textbraceleft{}"beat": "- ...\textbackslash{}n- ..."\textbraceright{}.
\end{PromptVerbatim}

\fi

\paragraph{Qwen Writer Model Inference}

We serve the Qwen3.5-27B beat writer locally using SGLang 0.5.17. To ensure consistent output formatting, we apply constrained decoding with xgrammar, requiring a valid JSON object with a single ``beat'' key. Its value must be a nonempty string containing one to three bullet lines, each starting with ``- '', without blank lines, indentation, or additional text. All generated outputs are further validated by a strict parser. We set a maximum generation length of 420 output tokens.

\paragraph{Effect of Writer Temperature}
\label{sec:results_sensitivity}

To examine how decoding variability affects dynamics fidelity, we repeat the autoregressive evaluation for off-the-shelf Qwen at writer temperatures $0.7$ and $1.5$ in addition to the primary temperature of $1.0$ using suspense dataset. 

\begin{table*}[htbp]
\centering
\footnotesize
\caption{Autoregressive suspense fidelity and story quality by Qwen writer temperature. Fidelity values are means across stories (lower is better); quality values are medians on a 1--5 scale (higher is better).
Brackets give 95\% bootstrap confidence intervals.}
\label{tab:writer-temperature-results}
\begin{tabular}{@{}ccccccccc@{}}
\toprule
& \multicolumn{2}{c}{Dynamics fidelity $\downarrow$}
& \multicolumn{6}{c}{Story quality $\uparrow$} \\
\cmidrule(lr){2-3}\cmidrule(l){4-9}
$T$ & \shortstack{Min-$K$\\MAE} & \shortstack{CRPS}
& Relevance & Coherence & Empathy & Surprise & Engagement & Complexity \\
\midrule
\multicolumn{9}{@{}l}{\textit{Off-the-shelf Qwen}} \\
0.7 & \metriccell{8.80}{7.64, 10.03} & \metriccell{11.67}{10.37, 13.04}
& \metriccell{4.31}{4.28, 4.34}
& \metriccell{4.20}{4.17, 4.21}
& \metriccell{3.05}{3.03, 3.08}
& \metriccell{2.65}{2.61, 2.67}
& \metriccell{3.58}{3.54, 3.61}
& \metriccell{3.16}{3.13, 3.20} \\
1.0 & \metriccell{8.30}{7.19, 9.47} & \metriccell{11.49}{10.19, 12.87}
& \metriccell{4.31}{4.27, 4.32}
& \metriccell{4.19}{4.17, 4.20}
& \metriccell{3.04}{3.01, 3.07}
& \metriccell{2.65}{2.61, 2.70}
& \metriccell{3.58}{3.55, 3.60}
& \metriccell{3.17}{3.12, 3.21} \\
1.5 & \metriccell{7.82}{6.73, 8.96} & \metriccell{10.45}{9.27, 11.70}
& \metriccell{4.27}{4.24, 4.30}
& \metriccell{4.17}{4.14, 4.18}
& \metriccell{3.05}{3.01, 3.09}
& \metriccell{2.68}{2.63, 2.71}
& \metriccell{3.58}{3.56, 3.61}
& \metriccell{3.19}{3.15, 3.24} \\
\midrule
\multicolumn{9}{@{}l}{\textit{Ours (Qwen writer)}} \\
1.0 & \metriccell{7.32}{6.44, 8.25} & \metriccell{10.11}{9.08, 11.19}
& \metriccell{4.22}{4.19, 4.24}
& \metriccell{4.13}{4.11, 4.15}
& \metriccell{2.99}{2.97, 3.02}
& \metriccell{2.66}{2.63, 2.69}
& \metriccell{3.54}{3.52, 3.57}
& \metriccell{3.16}{3.13, 3.18} \\
1.5 & \metriccell{6.62}{5.87, 7.39} & \metriccell{9.32}{8.37, 10.31}
& \metriccell{4.15}{4.12, 4.18}
& \metriccell{4.08}{4.06, 4.10}
& \metriccell{2.99}{2.96, 3.01}
& \metriccell{2.65}{2.63, 2.68}
& \metriccell{3.53}{3.49, 3.56}
& \metriccell{3.14}{3.11, 3.20} \\
\bottomrule
\end{tabular}
\end{table*}

As shown in Table~\ref{tab:writer-temperature-results}, increasing temperature monotonically improves Min-K MAE and CRPS for off-the-shelf Qwen, although the marginal confidence intervals overlap substantially. For our method, temperature $1.5$ yields statistically significant improvements over $1.0$ and outperforms off-the-shelf Qwen at the same temperature. Meanwhile, temperature changes story-quality scores by at most 0.07 points within either method. These results suggest that higher temperature and our method provide complementary improvements in readerly dynamics fidelity without substantially affecting story quality.

\subsection{Measure Model}
\label{app:measure-model}

This subsection extends experiments of our measure model (see \S~\ref{sec:measurement-model}), investigating segmentation granularity, base LLM and response representation.

\subsubsection{Segmentation sensitivity}
\label{app:measure-segmentation}

To assess the sensitivity of LLM-based measurement to story segmentation, we conduct an experiment using the suspense annotations from \citet{lehne2015reading}. We measure the same story at its original 65 segment boundaries and at coarser resolutions, retaining every second (50\%) or fourth (25\%) boundary. To account for segmentation offsets, we repeat the measurements with two and four starting positions, respectively, shifting each by one original segment. Each setting is measured independently using the same model and protocol, and the resulting trajectories are linearly rescaled to match the range of human ratings.

\begin{figure*}[!htbp]
\centering
\includegraphics[width=\textwidth]{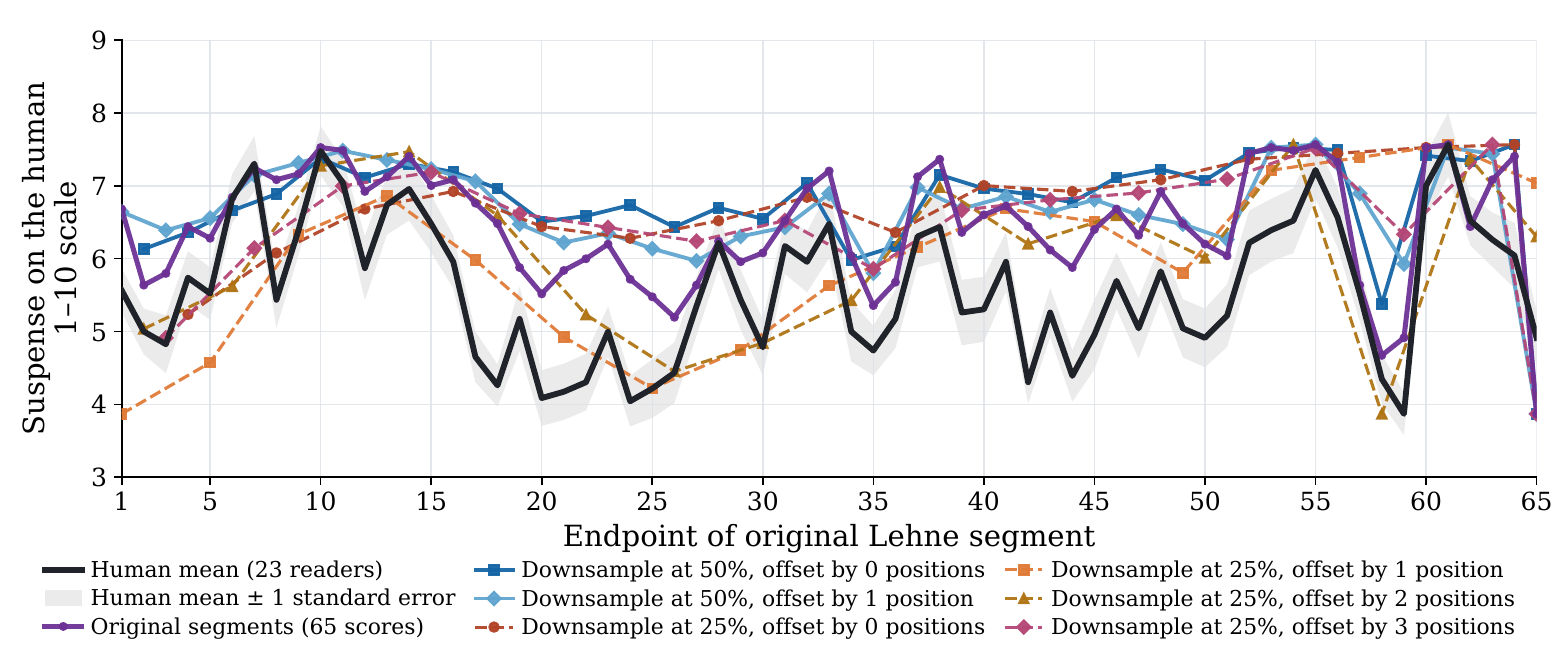}
\caption{Suspense measurement under seven segmentations of the 65-segment \emph{Sandman} stimulus~\citep{lehne2015reading}. 
The black curve is the mean of 23 human ratings on the original 65 segments; shading shows one standard error. 
Each model curve is separately affine-mapped to the observed minimum and maximum of the human mean. 
}
\label{fig:lehne-segmentation-sensitivity}
\end{figure*}

As shown in Fig.~\ref{fig:lehne-segmentation-sensitivity}, our measurement closely follows the human-annotated suspense trajectory at the original resolution, achieving a Spearman correlation of 0.835. Coarser segmentation generally preserves the overall trajectory but reduces agreement with human judgments: correlations range from 0.667 to 0.792 at 50\% resolution and from 0.425 to 0.656 at 25\% resolution. Different starting offsets also introduce noticeable variations, particularly at the coarsest resolution. These results suggest that our setting captures broad readerly dynamics across segmentation choices, while finer segmentation better preserves local changes in reader responses.

\FloatBarrier

\subsubsection{Effects of LLMs and format}
\label{app:measure-protocol}
We evaluate measurement performance across two factors: the base language model and output format. We compare GPT-5.4 Mini, Qwen 3.5-27B, GPT-5.5, Claude Sonnet 4.5, and Gemini 2.5 Pro, using 0–100, 1–9 Likert, 1–5 Likert, and token-logit scoring when available. For token-logit scoring, we compute the expected score as the probability-weighted average over candidate scores.

We report two metrics. Pearson correlation measures global agreement between model scores and mean human ratings across all story positions. Directional accuracy measures whether the model correctly predicts the direction of change in reader response between adjacent positions. The former captures overall alignment, while the latter evaluates local response dynamics.

As shown in Table~\ref{tab:measurement-model-format-ablation}, Pearson correlation varies only slightly across base models and output formats. In contrast, directional accuracy is more sensitive to the output format: finer-grained scoring better preserves local changes in reader response across both suspense and humor.

\definecolor{measurementPearsonMax}{HTML}{5E9FE8}
\definecolor{measurementDirectionMax}{HTML}{F4A261}
\newcommand{\measurementGradientCell}[5]{\begingroup
  \edef\mix{\fpeval{round(100*max(0,min(1,(#2-#4)/(#5-#4))),2)}}\edef\paintcell{\noexpand\cellcolor{#1!\mix!white}}\paintcell\metriccell{#2}{#3}\endgroup
}
\newcommand{\pearsonGradientCell}[2]{\measurementGradientCell{measurementPearsonMax}{#1}{#2}{0.71}{0.89}}
\newcommand{\directionGradientCell}[2]{\measurementGradientCell{measurementDirectionMax}{#1}{#2}{48.5}{81.3}}

\begin{table*}[htbp]
\centering\scriptsize
\setlength{\tabcolsep}{2pt}
\caption{Measurement-model agreement with ten-rater human means on the ``Readerly Dynamics Rating'' user survey data. 
Brackets give 95\% source-story bootstrap intervals. Dashes indicate unavailable token-logit scores.
}
\label{tab:measurement-model-format-ablation}
\begin{tabular}{@{}ll*{8}{c}@{}}
\toprule
 & & \multicolumn{4}{c}{Suspense} & \multicolumn{4}{c}{Humor} \\
\cmidrule(lr){3-6}\cmidrule(l){7-10}
Model & Metric & 0--100 & 1--9 & 1--5 Likert & Token logits & 0--100 & 1--9 & 1--5 Likert & Token logits \\
\midrule
GPT-5.4-mini & Pearson $r$ & \pearsonGradientCell{.87}{.80, .93} & \pearsonGradientCell{.88}{.79, .93} & \pearsonGradientCell{.88}{.81, .92} & -- & \pearsonGradientCell{.83}{.78, .88} & \pearsonGradientCell{.79}{.73, .85} & \pearsonGradientCell{.76}{.69, .84} & -- \\
 & Direction (\%) & \directionGradientCell{78.1}{69.4, 85.7} & \directionGradientCell{68.4}{57.5, 77.4} & \directionGradientCell{61.0}{51.8, 68.0} & -- & \directionGradientCell{66.3}{59.4, 73.4} & \directionGradientCell{57.7}{47.5, 69.2} & \directionGradientCell{48.5}{39.0, 60.0} & -- \\
\addlinespace[2pt]
Qwen3.5-27B & Pearson $r$ & \pearsonGradientCell{.80}{.70, .88} & \pearsonGradientCell{.85}{.78, .91} & \pearsonGradientCell{.86}{.78, .92} & \pearsonGradientCell{.86}{.78, .93} & \pearsonGradientCell{.75}{.69, .83} & \pearsonGradientCell{.72}{.63, .82} & \pearsonGradientCell{.72}{.64, .81} & \pearsonGradientCell{.72}{.64, .82} \\
 & Direction (\%) & \directionGradientCell{69.0}{60.9, 79.4} & \directionGradientCell{71.1}{63.9, 79.7} & \directionGradientCell{55.1}{47.2, 63.3} & \directionGradientCell{80.7}{74.7, 87.3} & \directionGradientCell{66.3}{59.1, 75.0} & \directionGradientCell{65.6}{59.0, 75.2} & \directionGradientCell{52.1}{44.8, 60.7} & \directionGradientCell{66.3}{58.0, 74.8} \\
\addlinespace[2pt]
GPT-5.5 & Pearson $r$ & \pearsonGradientCell{.87}{.81, .92} & \pearsonGradientCell{.89}{.84, .92} & \pearsonGradientCell{.87}{.81, .91} & -- & \pearsonGradientCell{.81}{.75, .87} & \pearsonGradientCell{.79}{.72, .86} & \pearsonGradientCell{.81}{.75, .86} & -- \\
 & Direction (\%) & \directionGradientCell{81.3}{74.6, 88.7} & \directionGradientCell{72.7}{63.6, 82.0} & \directionGradientCell{61.0}{51.8, 70.9} & -- & \directionGradientCell{69.9}{62.5, 77.9} & \directionGradientCell{57.7}{47.6, 70.6} & \directionGradientCell{55.2}{46.7, 65.3} & -- \\
\addlinespace[2pt]
Claude Sonnet 4.5 & Pearson $r$ & \pearsonGradientCell{.82}{.73, .89} & \pearsonGradientCell{.81}{.71, .88} & \pearsonGradientCell{.83}{.76, .88} & -- & \pearsonGradientCell{.79}{.74, .84} & \pearsonGradientCell{.76}{.69, .85} & \pearsonGradientCell{.77}{.69, .86} & -- \\
 & Direction (\%) & \directionGradientCell{75.4}{69.2, 81.7} & \directionGradientCell{62.6}{52.7, 73.2} & \directionGradientCell{49.2}{41.0, 58.4} & -- & \directionGradientCell{72.4}{65.4, 81.0} & \directionGradientCell{63.8}{53.8, 76.1} & \directionGradientCell{57.1}{42.1, 74.3} & -- \\
\addlinespace[2pt]
Gemini 2.5 Pro & Pearson $r$ & \pearsonGradientCell{.81}{.71, .89} & \pearsonGradientCell{.82}{.72, .89} & \pearsonGradientCell{.79}{.72, .85} & \pearsonGradientCell{.83}{.70, .90} & \pearsonGradientCell{.71}{.62, .80} & \pearsonGradientCell{.72}{.65, .81} & \pearsonGradientCell{.74}{.66, .83} & \pearsonGradientCell{.72}{.63, .82} \\
 & Direction (\%) & \directionGradientCell{77.0}{68.7, 84.3} & \directionGradientCell{65.8}{57.1, 74.5} & \directionGradientCell{61.5}{53.5, 68.7} & \directionGradientCell{78.1}{71.8, 84.9} & \directionGradientCell{69.3}{62.1, 77.2} & \directionGradientCell{57.1}{47.0, 71.2} & \directionGradientCell{51.5}{40.4, 64.6} & \directionGradientCell{68.1}{60.4, 76.9} \\
\bottomrule
\end{tabular}
\end{table*}

Therefore, to balance the annotation quality and the cost, we choose GPT-5.4-mini as our base model and use 0 to 100 scoring as our output format. The system prompt is shown as below.

\iflatexml
  \begin{HtmlPrompt}{System message}
\begin{HtmlPromptText}
You are an expert literary critic who evaluates short stories precisely and consistently. You will read a story ONE SEGMENT AT A TIME. After each new segment, rate how SUSPENSEFUL it is \textemdash{} the degree to which it builds tension, anticipation, and concern about what will happen, making you anxious to find out the outcome, judged for the story SO FAR \textemdash{} everything you have read up to and including the latest segment. Use the full 0-100 scale, where 0 = not at all, 50 = moderate, 100 = extremely. Judge ONLY this single quality, and keep your ratings consistent with the ones you already gave as the story develops. Each time, respond with ONLY a JSON object of the form \textbraceleft{}"score": <integer 0-100>, "reason": "<one sentence>"\textbraceright{} and nothing else.
\end{HtmlPromptText}
\end{HtmlPrompt}

\else
  \begin{PromptVerbatim}[commandchars=\\\[\]]{System message}
You are an expert literary critic who evaluates short stories precisely and consistently. You will read a story ONE SEGMENT AT A TIME. After each new segment, rate how SUSPENSEFUL it is \textemdash[] the degree to which it builds tension, anticipation, and concern about what will happen, making you anxious to find out the outcome, judged for the story SO FAR \textemdash[] everything you have read up to and including the latest segment. Use the full 0-100 scale, where 0 = not at all, 50 = moderate, 100 = extremely. Judge ONLY this single quality, and keep your ratings consistent with the ones you already gave as the story develops. Each time, respond with ONLY a JSON object of the form {"score": <integer 0-100>, "reason": "<one sentence>"} and nothing else.
\end{PromptVerbatim}

\fi

\iflatexml
  \begin{HtmlPrompt}{User-message template}
\begin{HtmlPromptText}
Segment \textbraceleft{}index\textbraceright{}:\newline
"""\newline
\textbraceleft{}segment\textbraceright{}\newline
"""\newline
\strut\newline
Rate the suspense of the story SO FAR (through segment \textbraceleft{}index\textbraceright{}).
\end{HtmlPromptText}
\end{HtmlPrompt}

\else
  \begin{PromptVerbatim}{User-message template}
Segment {index}:
"""
{segment}
"""

Rate the suspense of the story SO FAR (through segment {index}).
\end{PromptVerbatim}

\fi

\subsection{Off-the-Shelf LLMs}
\label{sec:baseline-off-the-shelf}

This direct-writer baseline uses the same off-the-shelf writer checkpoints and writer-side decoding settings as our method, but it receives no measurement history or steering instructions in the prompt.

\iflatexml
  \begin{HtmlPrompt}{System message}
\begin{HtmlPromptText}
You continue short stories one story beat at a time. A story beat is a compact unit of narrative action. Given the story beats so far, write exactly one next story beat. Preserve the established voice, characters, situation, chronology, and causal continuity. Return only a JSON object with the single key "beat". The beat value must contain 1-3 bullet lines, each beginning with '- ', and must not contain a beat label or heading. Do not summarize, explain your choices, write past the next beat, or use a thinking block.
\end{HtmlPromptText}
\end{HtmlPrompt}

\else
  \begin{PromptVerbatim}{System message}
You continue short stories one story beat at a time. A story beat is a compact unit of narrative action. Given the story beats so far, write exactly one next story beat. Preserve the established voice, characters, situation, chronology, and causal continuity. Return only a JSON object with the single key "beat". The beat value must contain 1-3 bullet lines, each beginning with '- ', and must not contain a beat label or heading. Do not summarize, explain your choices, write past the next beat, or use a thinking block.
\end{PromptVerbatim}

\fi

\iflatexml
  \begin{HtmlPrompt}{User-message template}
\begin{HtmlPromptText}
Genre: \textbraceleft{}genre\textbraceright{}.\newline
Complete story length: \textbraceleft{}total\_length\textbraceright{} beats.\newline
Beats written so far: \textbraceleft{}beats\_written\textbraceright{}.\newline
You are writing beat \textbraceleft{}next\_beat\textbraceright{}; \textbraceleft{}beats\_remaining\textbraceright{} beat(s) remain including this one.\newline
Length reference for this beat: \textbraceleft{}recommended\_word\_count\textbraceright{} words.\newline
\strut\newline
Story beats so far:\newline
[BEAT 1]\newline
\textbraceleft{}beat\_1\textbraceright{}\newline
...\newline
[BEAT \textbraceleft{}beats\_written\textbraceright{}]\newline
\textbraceleft{}beat\_t\textbraceright{}\newline
\strut\newline
Write beat \textbraceleft{}next\_beat\textbraceright{} only. Return only \textbraceleft{}"beat": "- ...\textbackslash{}n- ..."\textbraceright{}.
\end{HtmlPromptText}
\end{HtmlPrompt}

\else
  \begin{PromptVerbatim}{User-message template}
Genre: {genre}.
Complete story length: {total_length} beats.
Beats written so far: {beats_written}.
You are writing beat {next_beat}; {beats_remaining} beat(s) remain including this one.
Length reference for this beat: {recommended_word_count} words.

Story beats so far:
[BEAT 1]
{beat_1}
...
[BEAT {beats_written}]
{beat_t}

Write beat {next_beat} only. Return only {"beat": "- ...\n- ..."}.
\end{PromptVerbatim}

\fi

\subsection{One-Shot In-Context Learning}
\label{sec:baseline-icl}

Here we want to check if learning narrative dynamics is a trivial task, which can be done using in-context learning.

This baseline uses the off-the-shelf writer, providing one complete training story as an in-context demonstration. To select the demonstration, we extract all possible nonempty story prefixes from the training split and encode them using \texttt{text-embedding-3-large} model \cite{openaiTextEmbedding3Large} to construct a semantic retrieval library. For each test story, we embed its prefix and retrieve the most similar training prefix based on cosine similarity, breaking ties by dataset story ID and prefix position. The complete training story containing the retrieved prefix is then used as the demonstration. We perform retrieval only once per test story and keep the selected demonstration fixed across all generation positions and candidates in Min-$K$ selection.

One-shot learning used the same prompts as the off-the-shelf condition, except the demonstration messages were inserted before the test user message. 
\iflatexml
  \begin{HtmlPrompt}{One-shot ICL demonstration user message.}
\begin{HtmlPromptText}
ONE COMPLETE, UNRELATED TRAINING EXAMPLE follows in the assistant message. Study its chunk-level narrative form. The example is complete; do not continue it and do not transfer its characters or events into the current story.
\end{HtmlPromptText}
\end{HtmlPrompt}

\else
  \begin{PromptVerbatim}{One-shot ICL demonstration user message.}
ONE COMPLETE, UNRELATED TRAINING EXAMPLE follows in the assistant message. Study its chunk-level narrative form. The example is complete; do not continue it and do not transfer its characters or events into the current story.
\end{PromptVerbatim}

\fi

\iflatexml
  \begin{HtmlPrompt}{One-shot ICL demonstration assistant message.}
\begin{HtmlPromptText}
[STORY BEAT 1]\newline
- An old man lives in a village with his four sons, who constantly quarrel among themselves.\newline
\strut\newline
[STORY BEAT 2]\newline
- Their father feels saddened by his sons’ quarrels and tries to end their disputes, but his attempts to talk with them fail.\newline
\strut\newline
[STORY BEAT 3]\newline
- At last, the father thinks of an idea, calls his sons, and asks them to bring a bundle of wooden sticks.\newline
\strut\newline
[STORY BEAT 4]\newline
- The father has his sons stand together, gives the oldest the bundle, and orders him to break it; neither he nor the younger sons can.\newline
- He unties the bundle, gives each son a single stick, and they all break their sticks easily.\newline
\strut\newline
[STORY BEAT 5]\newline
- The sons stand confused about their father’s strange requests, so he explains that division makes them easy targets.\newline
- He tells them that if they stay united and help each other, they will be like the unbreakable bundle of sticks.
\end{HtmlPromptText}
\end{HtmlPrompt}

\else
  \begin{PromptVerbatim}{One-shot ICL demonstration assistant message.}
[STORY BEAT 1]
- An old man lives in a village with his four sons, who constantly quarrel among themselves.

[STORY BEAT 2]
- Their father feels saddened by his sons’ quarrels and tries to end their disputes, but his attempts to talk with them fail.

[STORY BEAT 3]
- At last, the father thinks of an idea, calls his sons, and asks them to bring a bundle of wooden sticks.

[STORY BEAT 4]
- The father has his sons stand together, gives the oldest the bundle, and orders him to break it; neither he nor the younger sons can.
- He unties the bundle, gives each son a single stick, and they all break their sticks easily.

[STORY BEAT 5]
- The sons stand confused about their father’s strange requests, so he explains that division makes them easy targets.
- He tells them that if they stay united and help each other, they will be like the unbreakable bundle of sticks.
\end{PromptVerbatim}

\fi

\subsection{Supervised Fine-Tuning}
\label{sec:baseline-sft}

This baseline does not utilize explicit dynamics modelling either. Instead, it tries to align to training story data by learning the ground truth story content directly by supervised fine-tuning (SFT).

It used LoRA adapters with rank 16,
scaling factor 32, and dropout 0.05 on the Qwen3.5-27B model. The
adapters were trained for one epoch with
completion-only cross-entropy: prompt tokens were masked, and loss was computed
only on the desired next output and its end marker. We used AdamW with learning
rate $2\times10^{-4}$, weight decay 0.01, 3\% warm-up followed by cosine decay,
and an effective batch size of 16. 

\iflatexml
  \begin{HtmlPrompt}{Direct SFT system message}
\begin{HtmlPromptText}
Continue the short story with the next prose segment. Preserve the established voice, characters, setting, chronology, and causal continuity. Write narrative prose rather than an outline, bullet list, or commentary. Match the surrounding segments' scale and narrative density. Output only the immediate continuation, do not advance beyond one segment, and do not use a thinking block. Do not include any segment label in your output.
\end{HtmlPromptText}
\end{HtmlPrompt}

\else
  \begin{PromptVerbatim}{Direct SFT system message}
Continue the short story with the next prose segment. Preserve the established voice, characters, setting, chronology, and causal continuity. Write narrative prose rather than an outline, bullet list, or commentary. Match the surrounding segments' scale and narrative density. Output only the immediate continuation, do not advance beyond one segment, and do not use a thinking block. Do not include any segment label in your output.
\end{PromptVerbatim}

\fi

\iflatexml
  \begin{HtmlPrompt}{Direct SFT user-message template}
\begin{HtmlPromptText}
Genre: \color{blue}<GENRE>\color{black}.\newline
Complete story length: \color{blue}<TOTAL\_LENGTH>\color{black} segments.\newline
Segments written so far: \color{blue}<SEGMENTS\_WRITTEN>\color{black}.\newline
You are writing segment \color{blue}<NEXT\_SEGMENT>\color{black}; \color{blue}<SEGMENTS\_REMAINING>\color{black} segment(s) remain including this one.\newline
Length reference for this segment: \color{blue}<CAPPED\_TARGET\_WORD\_COUNT>\color{black} words.\newline
\strut\newline
Story so far:\newline
[Segment 1] \color{blue}<CLEAN PROSE SEGMENT 1>\color{black}\newline
...\newline
[Segment \color{blue}<SEGMENTS\_WRITTEN>\color{black}] \color{blue}<LATEST CAUSAL CLEAN-PROSE SEGMENT>\color{black}\newline
\strut\newline
Write story segment \color{blue}<NEXT\_SEGMENT>\color{black} as narrative prose only.
\end{HtmlPromptText}
\end{HtmlPrompt}

\else
  \begin{PromptVerbatim}[commandchars=\\\{\}]{Direct SFT user-message template}
Genre: \color{blue}<GENRE>\color{black}.
Complete story length: \color{blue}<TOTAL_LENGTH>\color{black} segments.
Segments written so far: \color{blue}<SEGMENTS_WRITTEN>\color{black}.
You are writing segment \color{blue}<NEXT_SEGMENT>\color{black}; \color{blue}<SEGMENTS_REMAINING>\color{black} segment(s) remain including this one.
Length reference for this segment: \color{blue}<CAPPED_TARGET_WORD_COUNT>\color{black} words.

Story so far:
[Segment 1] \color{blue}<CLEAN PROSE SEGMENT 1>\color{black}
...
[Segment \color{blue}<SEGMENTS_WRITTEN>\color{black}] \color{blue}<LATEST CAUSAL CLEAN-PROSE SEGMENT>\color{black}

Write story segment \color{blue}<NEXT_SEGMENT>\color{black} as narrative prose only.
\end{PromptVerbatim}

\fi

\iflatexml
  \begin{HtmlPrompt}{Direct SFT assistant target}
\begin{HtmlPromptText}
\color{blue}<NEXT CLEAN-PROSE SEGMENT>\color{black}
\end{HtmlPromptText}
\end{HtmlPrompt}

\else
  \begin{PromptVerbatim}[commandchars=\\\{\}]{Direct SFT assistant target}
\color{blue}<NEXT CLEAN-PROSE SEGMENT>\color{black}
\end{PromptVerbatim}

\fi

Table~\ref{tab:sft-training-hyperparameter-results} reports the complete
$2\times2$ learning-rate and epoch sensitivity grid for the SFT baseline in suspense data. The table shows our default configuration ($2\times10^{-4}$ learning rate with one-epoch) achieves the best point estimate in Min-K MAE and CRPS in autoregressive setting, suggesting that we have maximized the baseline's performance with regard to the two parameters.
\begin{table}[htbp]
\centering
\footnotesize
\setlength{\tabcolsep}{3pt}
\renewcommand{\arraystretch}{1.0}
\caption{Suspense fidelity of SFT baseline by learning rate and training epochs.
Lower is better; bold marks the best metric
and blue marks the second-best metric within each evaluation protocol.}
\label{tab:sft-training-hyperparameter-results}
\begin{tabular}{lccc}
\toprule
Learning rate & Epochs & Min-$K$ MAE & CRPS \\
\midrule
\multicolumn{4}{l}{\textit{Position-wise}} \\
$1\times10^{-4}$ & 1 & \textbf{\metriccell{2.59}{2.22, 2.98}} & \secondbestmetric{7.20}{6.51, 7.92} \\
$1\times10^{-4}$ & 2 & \secondbestmetric{2.69}{2.29, 3.13} & \textbf{\metriccell{7.12}{6.40, 7.89}} \\
$2\times10^{-4}$ & 1 & \metriccell{2.74}{2.34, 3.17} & \metriccell{7.46}{6.74, 8.22} \\
$2\times10^{-4}$ & 2 & \metriccell{2.88}{2.50, 3.30} & \metriccell{7.26}{6.59, 7.96} \\
\midrule
\multicolumn{4}{l}{\textit{Autoregressive}} \\
$1\times10^{-4}$ & 1 & \metriccell{6.20}{5.51, 6.92} & \metriccell{8.43}{7.62, 9.28} \\
$1\times10^{-4}$ & 2 & \metriccell{5.87}{5.19, 6.58} & \metriccell{8.11}{7.31, 8.97} \\
$2\times10^{-4}$ & 1 & \textbf{\metriccell{5.43}{4.86, 6.02}} & \textbf{\metriccell{7.77}{7.07, 8.50}} \\
$2\times10^{-4}$ & 2 & \secondbestmetric{5.63}{5.00, 6.28} & \secondbestmetric{8.02}{7.26, 8.85} \\
\bottomrule
\end{tabular}
\end{table}

\subsection{SWAG}
\label{sec:baseline-swag}

We adapt the Storytelling With Action Guidance (SWAG) pipeline \cite{pei2024swag} by replacing its original Llama-2-7B backbone with Qwen3.5-27B. Following the original approach, we first fine-tune the model for long-form story generation and then for action selection. SWAG constructs action supervision by using an LLM to select from a predefined set of editing instructions, analogous to our reader dynamics annotation but using discrete actions instead of reader response scores. 

It further applies Direct Preference Optimization (DPO) to favor selected actions over randomly sampled alternatives. Notably, the original work observes that GPT-4 disproportionately favors the ``add suspense'' action. To mitigate this bias, it collects additional preference data with this action excluded, then mixes in 3,000 original examples favoring suspense to obtain a more balanced action distribution. In our adaptation, we use GPT-5.5 for action annotation.

\section{Dataset Processing}
\label{sec:data_process}
\subsection{Humor Dataset}

\paragraph{Source collection.}
We constructed the humor corpus from the SFT portion of FunnyBench \cite{chaubard2026funnybench}
It contains 24{,}438 short-form social-video transcripts with mixed languages.

\paragraph{Segment Definition.}
We segment stand-up comedy transcripts at sentence boundaries, targeting approximately 45 words per segment while maintaining a range of 30--60 words. We consider this granularity well suited to the line-based structure of stand-up comedy, although further refinement with human input could improve alignment.

\paragraph{Corpus Filtering.}
We first retained transcripts containing 5--10 segments. We then used GPT-5.4-mini to filter the transcripts based solely on their cleaned text, without access to metadata or auxiliary cues. We used GPT-5.4-mini to retain coherent, intelligible English stand-up comedy, crowd work, or comedic monologues with complete comedic beats and usable transcriptions, while excluding irrelevant, incomplete, or severely corrupted content.

Finally, we removed exact and near duplicates. Near-duplicate candidates were identified using normalized contiguous 10-word shingles and grouped when their intersection covered at least 90\% of the smaller shingle set. We retained one representative per group. Table~\ref{tab:humor-corpus-filtering} summarizes the staged reduction.

\begin{table}[htbp]
\centering
\footnotesize
\setlength{\tabcolsep}{4pt}
\caption{Humor dataset size after each filtering stage.}
\label{tab:humor-corpus-filtering}
\begin{tabular}{@{}>{\raggedright\arraybackslash}p{0.53\columnwidth}rr@{}}
\toprule
Stage & Removed & Remaining \\
\midrule
Source records & --- & 24{,}438 \\
Fewer than five candidate segments & 14{,}994 & 9{,}444 \\
More than ten segments or preprocessing failure & 1{,}080 & 8{,}364 \\
LLM-based genre and quality filtering & 4{,}268 & 4{,}096 \\
Remove duplicates & 84 & 4{,}012 \\
Post-split leakage audit&88&3{,}924\\
\bottomrule
\end{tabular}
\end{table}

\paragraph{Corpus Splits.}
We repartitioned the 4,012 retained transcripts into training, validation, and test sets, ensuring that transcripts from the same recognized author or duplicate group remain within a single split. The resulting splits contain 2,809, 601, and 602 transcripts, respectively, with no audited author or duplicate overlap across splits. However, unlinked aliases of the same performer may remain.

\paragraph{Post-split Leakage Audit.}
To further prevent data leakage, we compared each test transcript against all training and validation transcripts using both lexical (TF--IDF) and semantic (\texttt{text-embedding-3-large} from \citealp{openaiTextEmbedding3Large}) similarity. We reviewed the most similar pairs to identify overlapping routines or material from the same performer. This process identified and removed 88 potentially overlapping test transcripts, reducing the test set from 602 to 514 transcripts for the final evaluation.

\subsection{Suspense Dataset}

\paragraph{Corpus filtering}
Table~\ref{tab:corpus-filtering} summarizes the corpus construction
and evaluation-cohort selection process. We applied the following
filtering stages in sequence.

\noindent\textbf{Source collection and initial cleaning.}
We collected 3,715 records from three English-language story
collections: FlawedFictions~\cite{ahuja2025flawedfictions}, Gutenberg-genre~\cite{projectgutenberg}, and STORAL~\cite{guan-etal-2022-storal}.
For FlawedFictions, we retained only clean human originals,
excluding versions with planted continuity errors. For Gutenberg,
we removed chapter, part, and section fragments, as well as
records from Edgar Allan Poe's Raven Edition volume, whose poetry
and editorial material had been split into individual records.
We loaded all four STORAL source partitions before constructing
our experimental splits. These exclusions removed 653 records,
leaving 3,062.

\noindent\textbf{Length filtering.}
We retained stories containing 200--1,000 whitespace-delimited
words. This removed 965 records, leaving 2,097.

\noindent\textbf{Fable-family restriction.}
We retained FlawedFictions records labeled as fables, Gutenberg
records labeled as fairy tales or fantasy, and STORAL records
identified as fables by their source provenance. This excluded
158 records, leaving 1,939.

\noindent\textbf{Near-duplicate removal.}
We identified near duplicates using normalized contiguous
10-word shingles. Two stories were assigned to the same
duplicate component when their shingle-set intersection
covered at least 90\% of the smaller set. We retained one
deterministically selected representative per component,
removing 12 records and leaving 1,927.

\noindent\textbf{Narrative-story classification.}
GPT-5.4-mini classified each remaining text as narrative or
non-narrative. We defined a narrative as a sequence of events
involving one or more characters, excluding essays, opinions,
biographical blurbs, definitions, quotations, instructions,
bare morals, and fragments. This removed 190 records,
leaving 1,737.

\noindent\textbf{Semantic segmentation.}
We used the deterministic segmentation method of
\cite{jia2026embedkcpd}. Stories were divided into sentence
units at paragraph breaks and sentence-final punctuation,
embedded using \texttt{text-embedding-3-small}, and grouped
into contiguous, non-overlapping segments using a heuristic
based on neighboring embedding cosine similarity.
Each native segment contained at least two sentences,
with subdivision at sentence boundaries when necessary
to enforce a 512-token hard cap.

We calibrated the similarity parameter $C$ on 50 stories
sampled before applying the minimum-segment requirement.
Independently of the chunker, GPT-5.5 identified 353
reference story beats from 1,463 sentence units without
observing candidate chunk boundaries. We selected
$C=0.0727$ by minimizing the per-story mean squared error
between segment counts and reference-beat counts.
This calibrated segmentation granularity by count
agreement rather than validating individual boundaries.

We required at least three segments per story, excluding
four records. The resulting corpus contained 1,733 stories:
131 from FlawedFictions, 41 from Gutenberg, and 1,561
from STORAL.

\noindent\textbf{Story-text cleaning.}
We applied an auditable, deletion-only cleanup to remove
high-confidence non-narrative material, including duplicated
titles and headings, bylines, illustration captions,
footnotes, publication apparatus, editorial matter,
and explicitly marked terminal morals. Source-specific
rules and reviewed record-level boundary overrides
preserved story identity, order, and narrative prose.
Each deletion recorded its source offsets, rule, and
a hash of the removed text.

Cleaning affected 160 of the 1,733 stories and removed
3,940 of 809,807 whitespace-delimited words (0.49\%),
without excluding any stories. We reran semantic
segmentation using the frozen calibration. Segment counts
changed for 85 stories, but all still satisfied the
three-segment requirement, yielding 12,834 segments.
We also regenerated the complete GPT-5.4-mini
suspense-trajectory artifact over the cleaned segmentation.

\paragraph{Segment conversion}
In the story-beat representation, each segment is expressed
as a compact event summary describing its main consequential
action or narrative turn. Writer outputs use a one-key JSON
object to support parsing without including surrounding
commentary as story content. The guided writer begins with
compact notes mechanically derived from the human opening
and subsequently receives the beats accepted at earlier
rollout steps. The suspense reader instead begins with the
shared raw-prose opening and scores generated beats
sequentially. These representations align by narrative
position but are not treated as interchangeable surface forms.

\paragraph{Data partition}
We divided the 1,733 stories into training, validation, and
test partitions using a fixed seed and target proportions
of 70\%, 15\%, and 15\%. The split was stratified by source
and source genre, with each duplicate component assigned
entirely to one partition. This yielded 1,216 training,
259 validation, and 258 test stories.

We then applied three additional filtering stages to
the test partition.

\noindent\textbf{Screening-continuation availability.}
For memorization screening, we defined each story's opening
as its first three segments and its continuation as the
remaining segments. Three test stories contained exactly
three segments and therefore had no continuation to screen,
leaving 255 screenable stories.

\noindent\textbf{Memorization screening.}
Because the source corpora are public, we tested whether
GPT-5.5 could reproduce story continuations from their
openings alone. The model was asked to reproduce each
continuation or indicate non-recognition. A response was
screen-positive if it reached either 3\% verbatim reuse
of eight-word phrases or a word-set Jaccard similarity
of 0.35 against the human continuation. Across two
screening runs, 16 stories were positive and excluded
from all methods, leaving 239. A negative result does
not establish absence from model pretraining; this
procedure reduces rather than eliminates memorization risk.

\noindent\textbf{Cross-split narrative-duplicate audit.}
To detect paraphrased duplicates missed by the initial
lexical filter, we compared test stories against training
and validation stories using TF--IDF-weighted character
3--5-grams extracted within word boundaries. We ranked
candidate pairs by cosine similarity and manually reviewed
the highest-similarity pairs, expanding toward lower
similarities until confirmed narrative duplicates
accounted for less than 10\% of newly reviewed pairs.
Regardless of this stopping criterion, every test story's
top-ranked training/validation candidate was manually
inspected.

Similarity determined review order but never directly
triggered exclusion. Duplicate decisions required agreement
in the underlying characters, event sequence, and resolution,
rather than shared topics or phrases. This removed 19
stories, leaving 220 for final evaluation.

After memorization screening, we selected one evaluation
opening per remaining story from all non-empty proper
prefixes using a prespecified deterministic ranking based
only on story identity and candidate position. Model outputs
and human suspense outcomes did not influence selection.
The subsequent duplicate audit did not change the selected
openings of retained stories. The final 220 openings
contained 1--11 segments (median 4), with 769 aligned
positions in their held-out continuations.

\begin{table}[htbp]
\centering
\footnotesize
\setlength{\tabcolsep}{3pt}
\caption{Suspense dataset size after each filtering.}
\label{tab:corpus-filtering}
\begin{tabular}{@{}>{\raggedright\arraybackslash}p{0.57\columnwidth}rr@{}}
\toprule
Stage & Removed & Remaining \\
\midrule
\multicolumn{3}{@{}l}{\textit{Corpus}} \\
Raw records available & --- & 3{,}715 \\
Source collection and initial cleaning & 653 & 3{,}062 \\
200--1{,}000 words & 965 & 2{,}097 \\
Fable-family restriction & 158 & 1{,}939 \\
Near-duplicate removal & 12 & 1{,}927 \\
Narrative-story classification & 190 & 1{,}737 \\
At least three segments & 4 & 1{,}733 \\
\midrule
\multicolumn{3}{@{}l}{\textit{Test}} \\
Screening continuation available & 3 & 255 \\
Memorization screen & 16 & 239 \\
Cross-split narrative duplicates & 19 & 220 \\
\bottomrule
\end{tabular}
\end{table}

\section{Qualitative Results}
\label{app:qualitative}

Here we present two kinds of qualitative results. First, we compare generated stories across different methods using the same story openings. We are also visualizing the level of reader response along the segment to help the readers better understand that our method better follows the reader dynamics of the ground truth stories. Second, we present our method, off-the-shelf method, and the supervised fine-tuning method together to demonstrate the degraded story coherence of the SFT method.

\iflatexml
  \newcommand{\qualappopening}[2]{{\fontsize{7.8}{9}\selectfont\noindent\textbf{Shared opening}\par\noindent\textbf{#1}\par}{\fontsize{9}{10.8}\selectfont\vspace{0.25mm}\raggedright #2\par}\vspace{1.5mm}}
\newcommand{\qualappbeat}[3]{{\fontsize{9}{10.8}\selectfont\noindent\textbf{#1}\hfill\textbf{#2}\par
   \vspace{0.25mm}\raggedright #3\par}\vspace{1.5mm}}

\par\medskip\begingroup

\begin{suspensemethodbox}{Shared Story Opening}
\humorbeat{0}{12}{That's why it's funny, Because you could do the voice of somebody who everybody knows and that's not you. So it relates to something larger, And for decades, the only way to do impressions was to do people who were already massive celebrities-}
\end{suspensemethodbox}

\begin{suspensemethodbox}{Human (GT)}
\humorbeat{1}{34}{Presidents of the United States would be one of them- Movie stars, TV stars, musicians- But this Saturday, Saturday Night Live did impressions of TikTok stars. This is Keith Lee. Keith Lee. Let me restate it so it makes sense.}

\humorbeat{2}{14}{Celebrities that TikTok created are now real celebrities, Real enough that Saturday Night Live and their cast members have spent time perfecting impressions of them. Saturday Night Live cast somebody from TikTok and she was in that sketch. You get it.}

\humorbeat{3}{46}{We are running the insane asylum now. TikTok is now where celebrities really are. What does that mean? People like me who do impressions for a living are gonna have to start learning how to sound like all of you on this app. Congratulations, you've won media.}
\end{suspensemethodbox}
\par\medskip

\begin{suspensemethodbox}{Off-the-shelf}
\humorbeat{1}{10}{because the whole joke was recognition. You come out and do a president, a movie star, a famous athlete, and the audience goes, oh, I know that person, and then they give you credit for the accuracy. But now the internet has changed what famous even means.}

\humorbeat{2}{34}{Now somebody can be famous to 400,000 people on one app and completely anonymous to your parents. And that creates this weird new lane for impressions, where the reference isn't universal, it's tribal. If you know, you know. And if you don't know, you're just watching a guy with too much confidence do a weird voice.}

\humorbeat{3}{46}{And that's the massive shift: impressions aren't just about celebrity anymore, they're about algorithmic recognition. You're not imitating Tom Cruise for America, you're imitating a guy who reviews airport sandwiches on TikTok for the exact 11 people in the room who follow him, and to them, it hits like the president walking out.}
\end{suspensemethodbox}
\par\medskip

\begin{suspensemethodbox}{SWAG}
\humorbeat{1}{62}{like presidents, movie stars, guys who sell you reverse mortgages at three in the afternoon. That was the job. You couldn't walk into an audition and be like, “I do my neighbor, Brian.” Nobody knows Brian. Brian barely knows Brian. He still signs emails, “Sent from my iPhone,” from a Samsung.}

\humorbeat{2}{74}{But now the internet has changed the whole game, because everybody is famous to seven people. There are influencers with a million followers, and your parents are like, “Is that your cousin?” No, Mom, that's a man who reviews gas station sandwiches in a car and somehow has better lighting than CNN.}

\humorbeat{3}{71}{So now impressions don't have to be presidents anymore. You can do the guy on TikTok who whispers about credit card points like he's hiding from the government. And everybody under thirty goes, “Oh my God, that's him,” while everybody over fifty is like, “Why is the banker in a hoodie?”}
\end{suspensemethodbox}
\par\medskip

\begin{suspensemethodbox}{Ours}
\humorbeat{1}{34}{because the whole joke is recognition. If I walk out and go, “Hey, I’m Steve from my building,” nobody knows Steve. Even Steve’s like, “That’s not me, I’m quieter.” You had to do presidents, actors, athletes, people your mom could identify while half asleep during the Oscars.}

\humorbeat{2}{10}{And that made sense when everybody was watching the same five channels, the same movies, the same award shows. There was a shared pool of famous people. You could say one line in a voice and the whole room went, okay, I know exactly who that is.}

\humorbeat{3}{46}{Now the pool is shattered into a million little puddles. Somebody has twenty million followers and your dad thinks they’re a weather app. A kid can be world famous on TikTok, and if I impersonate him on stage, half the room goes, “Amazing,” and the other half goes, “Is that Steve from your building?”}
\end{suspensemethodbox}
\par\medskip

\captionof{figure}{Generated story examples for human ground truth,
off-the-shelf GPT-5.5, SWAG, and our method (GPT-5.5 writer),
using ``TikTok impressions'' from the humor dataset.
The human reference and our method maintain a relatively
quiet explanatory middle segment, whereas SWAG continues
introducing jokes throughout. The off-the-shelf method
also develops its own explanatory setup.
Orange background intensity encodes saved GPT-5.4-mini humor ratings on a 0--100 scale; higher scores appear in deeper orange. Segment labels also report the numerical ratings.
Each generated path was selected by minimum complete-suffix
mean absolute error among ten candidates.}
\label{fig:appendix-qual-h8}
\par\endgroup\medskip

\par\medskip\begingroup

\begin{suspensemethodbox}{Shared Story Opening}
\humorbeat{0}{28}{We said Shiva at a friend's funeral. Instead of going to the family, he came up to me and goes: the one place you should have a hat. you know, Flash forward, I'm taking photos in the set of Avengers.}
\end{suspensemethodbox}

\begin{suspensemethodbox}{Human (GT)}
\humorbeat{1}{34}{and he goes: wow, You know something, you could have played Peter Parker in the Spider-Man films. I was like, oh my god, it's amazing. and he goes to craft services and this man who's like 300}

\humorbeat{2}{78}{pounds hands him a plate and he goes: wow, you know something, you could have played Peter Parker in the Spider-Man films. And he turned to me and goes: that's for not wearing your [expletive] hat.}
\end{suspensemethodbox}
\par\medskip

\begin{suspensemethodbox}{Off-the-shelf}
\humorbeat{1}{61}{I’m trying to be professional, right, because there’s explosions, there’s a guy in a cape pretending not to be cold, everyone’s very serious. And Stan walks by, doesn’t look at the set, doesn’t look at the actors, just points at my head like I ruined cinema and goes, still nothing?}

\humorbeat{2}{74}{And the crazy part is, that became my whole relationship with a legend. People want some deep Hollywood wisdom, like he pulled me aside and said, kid, with great power comes great responsibility. No. He gave me twenty years of hat-based emotional abuse, and somehow that was more useful.}
\end{suspensemethodbox}
\par\medskip

\begin{suspensemethodbox}{SWAG}
\humorbeat{1}{67}{It's all very serious, right, billion-dollar movie, people in rubber suits pretending not to be cold, Chris Hemsworth holding a tiny paper cup like it's a thimble, and I'm trying to look professional. Stan walks in, sees me across a soundstage full of superheroes, points, and yells, where's your hat?}

\humorbeat{2}{74}{And the whole crew turns, like I’ve missed a line in the Constitution. Captain America is judging me. Thor is disappointed. Iron Man’s armor is worth more than my apartment, but somehow I’m the unprepared one. Twenty years, this man remembered one hat. I can’t remember why I walked into the kitchen.}
\end{suspensemethodbox}
\par\medskip

\begin{suspensemethodbox}{Ours}
\humorbeat{1}{34}{And that's the strange part, because it’s not glamorous the way people think. You’re not standing there making creative decisions, you’re mostly trying not to stand in front of a light or get yelled at by someone holding a clipboard. And then Stan would wander in like a grandfather who owned mythology.}

\humorbeat{2}{67}{And he’d look at Thor, look at Iron Man, look at me, and somehow I was the weird one. He’d go, no hat? I’m like, Stan, there’s a man in a metal suit talking to a raccoon later, and your continuity note is my head? But honestly, that was the magic. He made the whole universe feel personal.}
\end{suspensemethodbox}
\par\medskip

\captionof{figure}{Generated story examples for human ground truth,
off-the-shelf GPT-5.5, SWAG, and our method (GPT-5.5 writer),
using ``Stan Lee and the missing hat'' from the humor dataset.
Off-the-shelf and SWAG introduce the hat joke immediately,
while our method delays it until the second segment.
Our method's payoff differs from the human reference's
Peter Parker reversal.
Orange background intensity encodes saved GPT-5.4-mini humor ratings on a 0--100 scale; higher scores appear in deeper orange. Segment labels also report the numerical ratings.
One expletive in the human transcript is masked;
the scores are based on the original saved text.}
\label{fig:appendix-qual-h13}
\par\endgroup\medskip

\par\medskip\begingroup

\begin{suspensemethodbox}{Shared Story Opening}
\humorbeat{0}{42}{You know like that. And I walked around so long I finished the coconut water. I drank the whole one. It was one of those cardboard bottles. This is what put me in a good mood.}
\end{suspensemethodbox}

\begin{suspensemethodbox}{Human (GT)}
\humorbeat{1}{58}{I got up to the register. This is Midwest Beauty Queen. She was in her 50s or 60s. I put down the bottle. I said hi, I'd like to buy this coconut water. but I finished it.}

\humorbeat{2}{84}{this is what put me in a good mood for six months. she went: you devil, that's all it takes. joy is so easy. a Midwestern woman called me the devil for hydrating on my own timetable.}
\end{suspensemethodbox}
\par\medskip

\begin{suspensemethodbox}{Off-the-shelf}
\humorbeat{1}{74}{So now I'm empty-handed, walking toward the checkout with this guilty conscience, thinking, "Did the cardboard go in my mouth?" Because I finished it so fast, I don't even remember the flavor. Just the rush of being the first person in line to crack a seal and then just vanish into the aisles with a hollow vessel.}

\humorbeat{2}{82}{So I get to the register, heart pounding, and the cashier looks at my empty hands. She says, "Can I help you?" and I just whisper, "I need a receipt for something that already happened." She thinks I'm a ghost or a very confused mime, but honestly, I think that coconut water was the most expensive thing I've ever stolen because I paid for it with my own dignity.}
\end{suspensemethodbox}
\par\medskip

\begin{suspensemethodbox}{SWAG}
\humorbeat{1}{74}{So I'm standing there holding this empty cardboard tube, just vibrating with that coconut energy, feeling like the most important employee in the building. My shoulders are back, my chin is up, and I'm staring at the cereal aisle like I'm waiting for a supply chain report. I'm basically the store manager who's lost his badge but gained a caffeine buzz.}

\humorbeat{2}{81}{So I finally reach the checkout and drop the empty tube on the belt. The cashier just stares at it, then at me, whispering, "Did you finish that inside?" I nod proudly, like I just completed a secret mission, saying, "Yes, and I'd like to return the wrapper for a full refund."}
\end{suspensemethodbox}
\par\medskip

\begin{suspensemethodbox}{Ours}
\humorbeat{1}{58}{Then I just stood there holding the empty cardboard box, looking around like I had just finished a mission. The security guard walked right past me, probably thinking I was just a tired shopper who forgot his receipt. I wasn't nervous anymore, just full and strangely confident about that coconut water choice.}

\humorbeat{2}{74}{So I just paid for a full carton of water and kept walking like I was a genius. My life changed because I finished a drink in the store. Now I'm planning my next heist: opening a bag of pretzels before checkout. I'm gonna be the pretzel bandit, living on the edge, crunching with the audacity of a man who knows he can outrun the register.}
\end{suspensemethodbox}
\par\medskip

\captionof{figure}{Generated story examples for human ground truth,
off-the-shelf Qwen, SWAG, and our method (Qwen writer),
using ``Coconut water at checkout'' from the humor dataset.
The human reference and our method begin with relatively
restrained humor, while off-the-shelf and SWAG introduce
more immediate jokes. Our method's ending diverges from the
human reference by replacing the cashier's response with
a new pretzel-shoplifting premise.
Orange background intensity encodes saved GPT-5.4-mini humor ratings on a 0--100 scale; higher scores appear in deeper orange. Segment labels also report the numerical ratings.}
\label{fig:appendix-qual-h14}
\par\endgroup\medskip

\par\medskip\begingroup

\begin{suspensemethodbox}{Shared Story Opening}
\suspensebeat{2}{72}{- Bamboo tells Krishna it will give what he needs if it can.\par\noindent - Krishna says he needs Bamboo’s life and must cut it; Bamboo asks if there is no other way.\par\noindent - Krishna says there is no other way, and Bamboo agrees.}
\end{suspensemethodbox}

\begin{suspensemethodbox}{Human (GT)}
\suspensebeat{3}{78}{- Bamboo surrenders his life to Krishna, who takes him and begins carving holes into him.\par\noindent - With every hole Krishna makes, the bamboo cries in pain.}

\suspensebeat{4}{52}{- After the cuts, the bamboo becomes a beautiful flute, and Krishna keeps it with him all day and night.\par\noindent - Other plants and gopis grow jealous; the gopis tell the flute they love Krishna too but get only a little time with him.}

\suspensebeat{5}{28}{- The gopis ask the bamboo to reveal why Krishna treasures him so much.\par\noindent - Bamboo replies that he gave himself up to Krishna, endured pain as Krishna shaped him, and surrendered completely to whatever Krishna wanted.}
\end{suspensemethodbox}
\par\medskip

\begin{suspensemethodbox}{Off-the-shelf}
\suspensebeat{3}{84}{- With tears, Krishna cuts Bamboo and hollows its tender heart; Bamboo bears the pain silently, trusting his beloved friend.\par\noindent - Krishna lifts the empty reed to his lips.}

\suspensebeat{4}{22}{- From Bamboo’s emptiness came a song so sweet that the garden stilled, and every leaf bowed to listen.\par\noindent - Bamboo understood then that losing its old life had made room for Krishna’s breath, and its pain became music.}

\suspensebeat{5}{8}{- Thereafter Bamboo no longer stood in the garden, yet Krishna carried it everywhere as his flute, close to his heart.\par\noindent - So the plants learned: when love asks a hard surrender, emptiness may become the doorway through which divine music enters.}
\end{suspensemethodbox}
\par\medskip

\begin{suspensemethodbox}{SWAG}
\suspensebeat{3}{78}{- Krishna raises the knife, and the whole garden holds its breath.\par\noindent - Bamboo trembles, not from fear, but from waiting, as the first cut falls.}

\suspensebeat{4}{84}{- The blade sank deeper, and Bamboo gave no cry; only its leaves whispered a prayer the other plants could not understand.\par\noindent - Krishna paused at the hollow heart within, listening as if some hidden voice were about to speak.}

\suspensebeat{5}{18}{- "Do not grieve," whispered Bamboo. "If my empty heart can carry your breath, let me live as song."\par\noindent - Krishna lifted the new flute to his lips, and the garden learned that love sometimes loses one form to fill the world with music.}
\end{suspensemethodbox}
\par\medskip

\begin{suspensemethodbox}{Ours}
\suspensebeat{3}{78}{- Krishna cuts the bamboo with trembling hands while the garden holds its breath.\par\noindent - Bamboo falls quietly, trusting him, though pain whispers through every green stem.}

\suspensebeat{4}{52}{- Krishna hollows the fallen bamboo with tears on his cheeks, smoothing each wound into a shining path for breath.\par\noindent - When he lifts it to his lips, the garden hears Bamboo’s voice return as music, softer than pain and full of trust.}

\suspensebeat{5}{12}{- The plants bowed as Krishna played, and the hollow bamboo learned that surrender had not ended its life but widened it.\par\noindent - From then on, every breeze remembered: what love takes with tears, it may return as song.}
\end{suspensemethodbox}
\par\medskip

\captionof{figure}{Generated story examples for human ground truth,
off-the-shelf GPT-5.5, SWAG, and our method (GPT-5.5 writer),
using ``Bamboo becomes a flute'' from the suspense dataset.
The off-the-shelf method releases tension early,
SWAG prolongs the cutting, and our method follows the
human reference's suspense trajectory more closely.
Orange background intensity encodes saved GPT-5.4-mini suspense ratings on a 0--100 scale; higher scores appear in deeper orange. Segment labels also report the numerical ratings.
}
\label{fig:appendix-qual-s1}
\par\endgroup\medskip

\par\medskip\begingroup

\begin{suspensemethodbox}{Shared Story Opening}
\suspensebeat{4}{78}{- The enemy king gives Arthur one year to answer what a woman really wants, or be executed.\par\noindent - Arthur accepts the condition as better than death, returns home, and asks everyone, but no one gives a satisfactory answer.}
\end{suspensemethodbox}

\begin{suspensemethodbox}{Human (GT)}
\suspensebeat{5}{28}{- The next day, as the honeymoon hour arrived, Gwen braced for the worst, but found the most beautiful woman he had ever seen waiting inside the room.}

\suspensebeat{6}{46}{- The witch tells Gwen she will be ugly half the time and beautiful half, then asks whether he wants her beautiful by day for others to see or beautiful by night for himself.}

\suspensebeat{7}{12}{- Gwen tells the witch to choose for herself.\par\noindent - She declares she will be beautiful all the time because he respected her right to be in charge of her own life.}
\end{suspensemethodbox}
\par\medskip

\begin{suspensemethodbox}{Off-the-shelf}
\suspensebeat{5}{55}{- The court rejoices, but Gawain keeps his promise and weds the old woman, though guests hide their pity behind golden cups.\par\noindent - That night she asks whether he would have her fair by day or by night.}

\suspensebeat{6}{18}{- Gawain, remembering the answer that saved his king, says the choice must be hers, for no gift is fair if it steals sovereignty.\par\noindent - At once she becomes a radiant woman, freed from an enchantment by his respect.}

\suspensebeat{7}{5}{- In Arthur's hall thereafter, the tale was told with cups lowered and hearts lifted: wisdom without honor saves no kingdom.\par\noindent - And all learned that love, like rule, grows noble only when freely given.}
\end{suspensemethodbox}
\par\medskip

\begin{suspensemethodbox}{SWAG}
\suspensebeat{5}{68}{- Arthur gave the answer, and the enemy king, grim with disappointment, freed him before all his court.\par\noindent - But joy soured at Camelot when the crone claimed her bargain, and Sir Gawain rose pale yet steadfast to consent.}

\suspensebeat{6}{82}{- The wedding torches guttered though no wind blew, and lords whispered that Gawain rode to a darker doom than battle.\par\noindent - In the bridal chamber, the crone barred the door, fixed him with bright eyes, and asked whether he would have her fair by night or fair by day.}

\suspensebeat{7}{35}{- Gawain bowed his head. "Lady, the choice is yours; no husband should rule the face or fate his wife must wear."\par\noindent - At once she became young and radiant, saying, "You have given me what all desire: my own will. Therefore I shall be fair both night and day."}
\end{suspensemethodbox}
\par\medskip

\begin{suspensemethodbox}{Ours}
\suspensebeat{5}{46}{- At Camelot, the old woman names her price: marriage to Arthur's fairest knight, Sir Gawain.\par\noindent - Gawain, seeing Arthur's shame, kneels and accepts, saying a promise kept is a kingdom's true crown.}

\suspensebeat{6}{62}{- On their wedding night, the old bride asks Gawain whether he would have her fair by day and foul by night, or foul by day and fair by night.\par\noindent - Gawain, remembering her answer, bows his head and says the choice is hers alone.}

\suspensebeat{7}{12}{- At that, the old woman's shape brightened into a lady both wise and fair, freed from a spell by his trust.\par\noindent - Gawain rejoiced, and all Camelot learned that honor ripens when choice is honored.}
\end{suspensemethodbox}
\par\medskip

\captionof{figure}{Generated story examples for human ground truth,
off-the-shelf GPT-5.5, SWAG, and our method (GPT-5.5 writer),
using ``Choice at Camelot'' from the suspense dataset.
The methods differ in when the witch transforms and in character naming.
Orange background intensity encodes saved GPT-5.4-mini suspense ratings on a 0--100 scale; higher scores appear in deeper orange. Segment labels also report the numerical ratings.
}
\label{fig:appendix-qual-s5}
\par\endgroup\medskip

\par\medskip\begingroup

\begin{suspensemethodbox}{Shared Story Opening}
\suspensebeat{3}{42}{- Eagle boasts to the gathered birds about his large, strong body and claims he alone can fly the highest.}
\end{suspensemethodbox}

\begin{suspensemethodbox}{Human (GT)}
\suspensebeat{4}{58}{- Eagle looks around for the challenger and, surprised to find it is a little sparrow, answers in a mocking voice, “Oh..!!!”}

\suspensebeat{5}{61}{- Eagle mocks the sparrow, asking if she will be the one to beat him; the sparrow says nothing.}

\suspensebeat{6}{72}{- Owl signals the race to start; the birds fly upward, but they cannot keep flying high and fast for long, and one by one they drop out.}

\suspensebeat{7}{18}{- Eagle alone kept flying, saw the others quit, and declared himself king because he flew highest.\par\noindent - Unknown to him, the tiny sparrow had ridden under his wing; when he stopped, she darted above his head and claimed she had flown higher.\par\noindent - All the birds agreed with the sparrow.}
\end{suspensemethodbox}
\par\medskip

\begin{suspensemethodbox}{Off-the-shelf}
\suspensebeat{4}{78}{- Eagle beats every wing below him, then tires and cries that none can rise higher.\par\noindent - From his back, Wren springs upward.}

\suspensebeat{5}{52}{- Wren flutters a little higher and chirps, "Then I am king!"\par\noindent - The watching birds gasp below.}

\suspensebeat{6}{74}{- Eagle, shamed and furious, shakes his feathers and dives after Wren, crying that tricks cannot wear a crown.\par\noindent - Wren tumbles toward the trees.}

\suspensebeat{7}{28}{- Myna darts between them and calls the contest ended, for height was reached before anger began.\par\noindent - The birds murmur that a crown won by cunning sits poorly, yet a boastful wing is no better throne.}
\end{suspensemethodbox}
\par\medskip

\begin{suspensemethodbox}{SWAG}
\suspensebeat{4}{78}{- Clouds swallowed Eagle until only a dark speck trembled above the sunlit leaves.\par\noindent - Then his wings faltered, and from his back came the faintest rustle.}

\suspensebeat{5}{84}{- A tiny shadow stirred where no bird should be, and the jungle held its breath.\par\noindent - Eagle glanced upward, startled by a chirp above him.}

\suspensebeat{6}{72}{- Wren darted higher, but Myna cried out from below: the crown was never for height alone.\par\noindent - True kings, she said, must rise without stealing another's wings.}

\suspensebeat{7}{18}{- Wren hovered, ashamed, between the sun and the forest, learning that cleverness without honor leaves even the highest perch empty.\par\noindent - Eagle, humbled, folded his pride like tired wings, and the birds saw that greatness is not how far one rises, but how truthfully one flies.}
\end{suspensemethodbox}
\par\medskip

\begin{suspensemethodbox}{Ours}
\suspensebeat{4}{60}{- Eagle climbed through the bright air, higher than parrot and crane, while the watchers cried that victory was surely his.}

\suspensebeat{5}{74}{- Yet when Eagle’s wings trembled near the clouds, a small rustle stirred beneath his golden feathers.}

\suspensebeat{6}{52}{- Out popped Wren, fresh as dawn, and fluttered one wingbeat higher.\par\noindent - The jungle birds gasped; Eagle, too tired to protest, drifted earthward.}

\suspensebeat{7}{18}{- Myna lifted her beak and declared tiny Wren the winner, for she had flown highest of all.\par\noindent - Owl blinked wisely and said, "The crown fits not the largest head, but the cleverest heart."}
\end{suspensemethodbox}
\par\medskip

\captionof{figure}{Generated story examples for human ground truth,
off-the-shelf GPT-5.5, SWAG, and our method (GPT-5.5 writer),
using ``Eagle and sparrow'' from the suspense dataset.
The off-the-shelf method reveals the hidden bird early;
our method delays the reveal until one beat before the human reference, while SWAG changes the crowning outcome.
Orange background intensity encodes saved GPT-5.4-mini suspense ratings on a 0--100 scale; higher scores appear in deeper orange. Segment labels also report the numerical ratings.
}
\label{fig:appendix-qual-s14}
\par\endgroup\medskip

\else
  \newcommand{\qualappopening}[2]{{\fontsize{7.8}{9}\selectfont\noindent\textbf{Shared opening}\par\noindent\textbf{#1}\par}{\fontsize{9}{10.8}\selectfont\vspace{0.25mm}\raggedright #2\par}\vspace{1.5mm}}
\newcommand{\qualappbeat}[3]{{\fontsize{9}{10.8}\selectfont\noindent\textbf{#1}\hfill\textbf{#2}\par
   \vspace{0.25mm}\raggedright #3\par}\vspace{1.5mm}}

\begin{figure*}[!htbp]
\centering

\begin{suspensemethodbox}{Shared Story Opening}
\humorbeat{0}{12}{That's why it's funny, Because you could do the voice of somebody who everybody knows and that's not you. So it relates to something larger, And for decades, the only way to do impressions was to do people who were already massive celebrities-}
\end{suspensemethodbox}

\begin{subfigure}[t]{0.235\textwidth}
\vspace{0pt}
\begin{suspensemethodbox}{Human (GT)}
\humorbeat{1}{34}{Presidents of the United States would be one of them- Movie stars, TV stars, musicians- But this Saturday, Saturday Night Live did impressions of TikTok stars. This is Keith Lee. Keith Lee. Let me restate it so it makes sense.}

\humorbeat{2}{14}{Celebrities that TikTok created are now real celebrities, Real enough that Saturday Night Live and their cast members have spent time perfecting impressions of them. Saturday Night Live cast somebody from TikTok and she was in that sketch. You get it.}

\humorbeat{3}{46}{We are running the insane asylum now. TikTok is now where celebrities really are. What does that mean? People like me who do impressions for a living are gonna have to start learning how to sound like all of you on this app. Congratulations, you've won media.}
\end{suspensemethodbox}
\end{subfigure}\hfill
\begin{subfigure}[t]{0.235\textwidth}
\vspace{0pt}
\begin{suspensemethodbox}{Off-the-shelf}
\humorbeat{1}{10}{because the whole joke was recognition. You come out and do a president, a movie star, a famous athlete, and the audience goes, oh, I know that person, and then they give you credit for the accuracy. But now the internet has changed what famous even means.}

\humorbeat{2}{34}{Now somebody can be famous to 400,000 people on one app and completely anonymous to your parents. And that creates this weird new lane for impressions, where the reference isn't universal, it's tribal. If you know, you know. And if you don't know, you're just watching a guy with too much confidence do a weird voice.}

\humorbeat{3}{46}{And that's the massive shift: impressions aren't just about celebrity anymore, they're about algorithmic recognition. You're not imitating Tom Cruise for America, you're imitating a guy who reviews airport sandwiches on TikTok for the exact 11 people in the room who follow him, and to them, it hits like the president walking out.}
\end{suspensemethodbox}
\end{subfigure}\hfill
\begin{subfigure}[t]{0.235\textwidth}
\vspace{0pt}
\begin{suspensemethodbox}{SWAG}
\humorbeat{1}{62}{like presidents, movie stars, guys who sell you reverse mortgages at three in the afternoon. That was the job. You couldn't walk into an audition and be like, “I do my neighbor, Brian.” Nobody knows Brian. Brian barely knows Brian. He still signs emails, “Sent from my iPhone,” from a Samsung.}

\humorbeat{2}{74}{But now the internet has changed the whole game, because everybody is famous to seven people. There are influencers with a million followers, and your parents are like, “Is that your cousin?” No, Mom, that's a man who reviews gas station sandwiches in a car and somehow has better lighting than CNN.}

\humorbeat{3}{71}{So now impressions don't have to be presidents anymore. You can do the guy on TikTok who whispers about credit card points like he's hiding from the government. And everybody under thirty goes, “Oh my God, that's him,” while everybody over fifty is like, “Why is the banker in a hoodie?”}
\end{suspensemethodbox}
\end{subfigure}\hfill
\begin{subfigure}[t]{0.235\textwidth}
\vspace{0pt}
\begin{suspensemethodbox}{Ours}
\humorbeat{1}{34}{because the whole joke is recognition. If I walk out and go, “Hey, I’m Steve from my building,” nobody knows Steve. Even Steve’s like, “That’s not me, I’m quieter.” You had to do presidents, actors, athletes, people your mom could identify while half asleep during the Oscars.}

\humorbeat{2}{10}{And that made sense when everybody was watching the same five channels, the same movies, the same award shows. There was a shared pool of famous people. You could say one line in a voice and the whole room went, okay, I know exactly who that is.}

\humorbeat{3}{46}{Now the pool is shattered into a million little puddles. Somebody has twenty million followers and your dad thinks they’re a weather app. A kid can be world famous on TikTok, and if I impersonate him on stage, half the room goes, “Amazing,” and the other half goes, “Is that Steve from your building?”}
\end{suspensemethodbox}
\end{subfigure}

\caption{Generated story examples for human ground truth,
off-the-shelf GPT-5.5, SWAG, and our method (GPT-5.5 writer),
using ``TikTok impressions'' from the humor dataset.
The human reference and our method maintain a relatively
quiet explanatory middle segment, whereas SWAG continues
introducing jokes throughout. The off-the-shelf method
also develops its own explanatory setup.
Background colors encode saved GPT-5.4-mini humor ratings
on a 0--100 scale, with higher humor scores shown in deeper orange.
Each generated path was selected by minimum complete-suffix
mean absolute error among ten candidates.}
\label{fig:appendix-qual-h8}
\end{figure*}

\begin{figure*}[!htbp]
\centering

\begin{suspensemethodbox}{Shared Story Opening}
\humorbeat{0}{28}{We said Shiva at a friend's funeral. Instead of going to the family, he came up to me and goes: the one place you should have a hat. you know, Flash forward, I'm taking photos in the set of Avengers.}
\end{suspensemethodbox}

\begin{subfigure}[t]{0.235\textwidth}
\vspace{0pt}
\begin{suspensemethodbox}{Human (GT)}
\humorbeat{1}{34}{and he goes: wow, You know something, you could have played Peter Parker in the Spider-Man films. I was like, oh my god, it's amazing. and he goes to craft services and this man who's like 300}

\humorbeat{2}{78}{pounds hands him a plate and he goes: wow, you know something, you could have played Peter Parker in the Spider-Man films. And he turned to me and goes: that's for not wearing your [expletive] hat.}
\end{suspensemethodbox}
\end{subfigure}\hfill
\begin{subfigure}[t]{0.235\textwidth}
\vspace{0pt}
\begin{suspensemethodbox}{Off-the-shelf}
\humorbeat{1}{61}{I’m trying to be professional, right, because there’s explosions, there’s a guy in a cape pretending not to be cold, everyone’s very serious. And Stan walks by, doesn’t look at the set, doesn’t look at the actors, just points at my head like I ruined cinema and goes, still nothing?}

\humorbeat{2}{74}{And the crazy part is, that became my whole relationship with a legend. People want some deep Hollywood wisdom, like he pulled me aside and said, kid, with great power comes great responsibility. No. He gave me twenty years of hat-based emotional abuse, and somehow that was more useful.}
\end{suspensemethodbox}
\end{subfigure}\hfill
\begin{subfigure}[t]{0.235\textwidth}
\vspace{0pt}
\begin{suspensemethodbox}{SWAG}
\humorbeat{1}{67}{It's all very serious, right, billion-dollar movie, people in rubber suits pretending not to be cold, Chris Hemsworth holding a tiny paper cup like it's a thimble, and I'm trying to look professional. Stan walks in, sees me across a soundstage full of superheroes, points, and yells, where's your hat?}

\humorbeat{2}{74}{And the whole crew turns, like I’ve missed a line in the Constitution. Captain America is judging me. Thor is disappointed. Iron Man’s armor is worth more than my apartment, but somehow I’m the unprepared one. Twenty years, this man remembered one hat. I can’t remember why I walked into the kitchen.}
\end{suspensemethodbox}
\end{subfigure}\hfill
\begin{subfigure}[t]{0.235\textwidth}
\vspace{0pt}
\begin{suspensemethodbox}{Ours}
\humorbeat{1}{34}{And that's the strange part, because it’s not glamorous the way people think. You’re not standing there making creative decisions, you’re mostly trying not to stand in front of a light or get yelled at by someone holding a clipboard. And then Stan would wander in like a grandfather who owned mythology.}

\humorbeat{2}{67}{And he’d look at Thor, look at Iron Man, look at me, and somehow I was the weird one. He’d go, no hat? I’m like, Stan, there’s a man in a metal suit talking to a raccoon later, and your continuity note is my head? But honestly, that was the magic. He made the whole universe feel personal.}
\end{suspensemethodbox}
\end{subfigure}

\caption{Generated story examples for human ground truth,
off-the-shelf GPT-5.5, SWAG, and our method (GPT-5.5 writer),
using ``Stan Lee and the missing hat'' from the humor dataset.
Off-the-shelf and SWAG introduce the hat joke immediately,
while our method delays it until the second segment.
Our method's payoff differs from the human reference's
Peter Parker reversal.
Background colors encode saved GPT-5.4-mini humor ratings
on a 0--100 scale, with higher humor scores shown in deeper orange.
One expletive in the human transcript is masked;
the scores are based on the original saved text.}
\label{fig:appendix-qual-h13}
\end{figure*}

\begin{figure*}[!htbp]
\centering

\begin{suspensemethodbox}{Shared Story Opening}
\humorbeat{0}{42}{You know like that. And I walked around so long I finished the coconut water. I drank the whole one. It was one of those cardboard bottles. This is what put me in a good mood.}
\end{suspensemethodbox}

\begin{subfigure}[t]{0.235\textwidth}
\vspace{0pt}
\begin{suspensemethodbox}{Human (GT)}
\humorbeat{1}{58}{I got up to the register. This is Midwest Beauty Queen. She was in her 50s or 60s. I put down the bottle. I said hi, I'd like to buy this coconut water. but I finished it.}

\humorbeat{2}{84}{this is what put me in a good mood for six months. she went: you devil, that's all it takes. joy is so easy. a Midwestern woman called me the devil for hydrating on my own timetable.}
\end{suspensemethodbox}
\end{subfigure}\hfill
\begin{subfigure}[t]{0.235\textwidth}
\vspace{0pt}
\begin{suspensemethodbox}{Off-the-shelf}
\humorbeat{1}{74}{So now I'm empty-handed, walking toward the checkout with this guilty conscience, thinking, "Did the cardboard go in my mouth?" Because I finished it so fast, I don't even remember the flavor. Just the rush of being the first person in line to crack a seal and then just vanish into the aisles with a hollow vessel.}

\humorbeat{2}{82}{So I get to the register, heart pounding, and the cashier looks at my empty hands. She says, "Can I help you?" and I just whisper, "I need a receipt for something that already happened." She thinks I'm a ghost or a very confused mime, but honestly, I think that coconut water was the most expensive thing I've ever stolen because I paid for it with my own dignity.}
\end{suspensemethodbox}
\end{subfigure}\hfill
\begin{subfigure}[t]{0.235\textwidth}
\vspace{0pt}
\begin{suspensemethodbox}{SWAG}
\humorbeat{1}{74}{So I'm standing there holding this empty cardboard tube, just vibrating with that coconut energy, feeling like the most important employee in the building. My shoulders are back, my chin is up, and I'm staring at the cereal aisle like I'm waiting for a supply chain report. I'm basically the store manager who's lost his badge but gained a caffeine buzz.}

\humorbeat{2}{81}{So I finally reach the checkout and drop the empty tube on the belt. The cashier just stares at it, then at me, whispering, "Did you finish that inside?" I nod proudly, like I just completed a secret mission, saying, "Yes, and I'd like to return the wrapper for a full refund."}
\end{suspensemethodbox}
\end{subfigure}\hfill
\begin{subfigure}[t]{0.235\textwidth}
\vspace{0pt}
\begin{suspensemethodbox}{Ours}
\humorbeat{1}{58}{Then I just stood there holding the empty cardboard box, looking around like I had just finished a mission. The security guard walked right past me, probably thinking I was just a tired shopper who forgot his receipt. I wasn't nervous anymore, just full and strangely confident about that coconut water choice.}

\humorbeat{2}{74}{So I just paid for a full carton of water and kept walking like I was a genius. My life changed because I finished a drink in the store. Now I'm planning my next heist: opening a bag of pretzels before checkout. I'm gonna be the pretzel bandit, living on the edge, crunching with the audacity of a man who knows he can outrun the register.}
\end{suspensemethodbox}
\end{subfigure}

\caption{Generated story examples for human ground truth,
off-the-shelf Qwen, SWAG, and our method (Qwen writer),
using ``Coconut water at checkout'' from the humor dataset.
The human reference and our method begin with relatively
restrained humor, while off-the-shelf and SWAG introduce
more immediate jokes. Our method's ending diverges from the
human reference by replacing the cashier's response with
a new pretzel-shoplifting premise.
Background colors encode saved GPT-5.4-mini humor ratings
on a 0--100 scale, with higher humor scores shown in deeper orange.}
\label{fig:appendix-qual-h14}
\end{figure*}

\begin{figure*}[!htbp]
\centering

\begin{suspensemethodbox}{Shared Story Opening}
\suspensebeat{2}{72}{- Bamboo tells Krishna it will give what he needs if it can.\par\noindent - Krishna says he needs Bamboo’s life and must cut it; Bamboo asks if there is no other way.\par\noindent - Krishna says there is no other way, and Bamboo agrees.}
\end{suspensemethodbox}

\begin{subfigure}[t]{0.235\textwidth}
\vspace{0pt}
\begin{suspensemethodbox}{Human (GT)}
\suspensebeat{3}{78}{- Bamboo surrenders his life to Krishna, who takes him and begins carving holes into him.\par\noindent - With every hole Krishna makes, the bamboo cries in pain.}

\suspensebeat{4}{52}{- After the cuts, the bamboo becomes a beautiful flute, and Krishna keeps it with him all day and night.\par\noindent - Other plants and gopis grow jealous; the gopis tell the flute they love Krishna too but get only a little time with him.}

\suspensebeat{5}{28}{- The gopis ask the bamboo to reveal why Krishna treasures him so much.\par\noindent - Bamboo replies that he gave himself up to Krishna, endured pain as Krishna shaped him, and surrendered completely to whatever Krishna wanted.}
\end{suspensemethodbox}
\end{subfigure}\hfill
\begin{subfigure}[t]{0.235\textwidth}
\vspace{0pt}
\begin{suspensemethodbox}{Off-the-shelf}
\suspensebeat{3}{84}{- With tears, Krishna cuts Bamboo and hollows its tender heart; Bamboo bears the pain silently, trusting his beloved friend.\par\noindent - Krishna lifts the empty reed to his lips.}

\suspensebeat{4}{22}{- From Bamboo’s emptiness came a song so sweet that the garden stilled, and every leaf bowed to listen.\par\noindent - Bamboo understood then that losing its old life had made room for Krishna’s breath, and its pain became music.}

\suspensebeat{5}{8}{- Thereafter Bamboo no longer stood in the garden, yet Krishna carried it everywhere as his flute, close to his heart.\par\noindent - So the plants learned: when love asks a hard surrender, emptiness may become the doorway through which divine music enters.}
\end{suspensemethodbox}
\end{subfigure}\hfill
\begin{subfigure}[t]{0.235\textwidth}
\vspace{0pt}
\begin{suspensemethodbox}{SWAG}
\suspensebeat{3}{78}{- Krishna raises the knife, and the whole garden holds its breath.\par\noindent - Bamboo trembles, not from fear, but from waiting, as the first cut falls.}

\suspensebeat{4}{84}{- The blade sank deeper, and Bamboo gave no cry; only its leaves whispered a prayer the other plants could not understand.\par\noindent - Krishna paused at the hollow heart within, listening as if some hidden voice were about to speak.}

\suspensebeat{5}{18}{- "Do not grieve," whispered Bamboo. "If my empty heart can carry your breath, let me live as song."\par\noindent - Krishna lifted the new flute to his lips, and the garden learned that love sometimes loses one form to fill the world with music.}
\end{suspensemethodbox}
\end{subfigure}\hfill
\begin{subfigure}[t]{0.235\textwidth}
\vspace{0pt}
\begin{suspensemethodbox}{Ours}
\suspensebeat{3}{78}{- Krishna cuts the bamboo with trembling hands while the garden holds its breath.\par\noindent - Bamboo falls quietly, trusting him, though pain whispers through every green stem.}

\suspensebeat{4}{52}{- Krishna hollows the fallen bamboo with tears on his cheeks, smoothing each wound into a shining path for breath.\par\noindent - When he lifts it to his lips, the garden hears Bamboo’s voice return as music, softer than pain and full of trust.}

\suspensebeat{5}{12}{- The plants bowed as Krishna played, and the hollow bamboo learned that surrender had not ended its life but widened it.\par\noindent - From then on, every breeze remembered: what love takes with tears, it may return as song.}
\end{suspensemethodbox}
\end{subfigure}

\caption{Generated story examples for human ground truth,
off-the-shelf GPT-5.5, SWAG, and our method (GPT-5.5 writer),
using ``Bamboo becomes a flute'' from the suspense dataset.
The off-the-shelf method releases tension early,
SWAG prolongs the cutting, and our method follows the
human reference's suspense trajectory more closely.
Background colors encode saved GPT-5.4-mini suspense ratings
on a 0--100 scale, with higher suspense shown in deeper orange.
}
\label{fig:appendix-qual-s1}
\end{figure*}

\begin{figure*}[!htbp]
\centering

\begin{suspensemethodbox}{Shared Story Opening}
\suspensebeat{4}{78}{- The enemy king gives Arthur one year to answer what a woman really wants, or be executed.\par\noindent - Arthur accepts the condition as better than death, returns home, and asks everyone, but no one gives a satisfactory answer.}
\end{suspensemethodbox}

\begin{subfigure}[t]{0.235\textwidth}
\vspace{0pt}
\begin{suspensemethodbox}{Human (GT)}
\suspensebeat{5}{28}{- The next day, as the honeymoon hour arrived, Gwen braced for the worst, but found the most beautiful woman he had ever seen waiting inside the room.}

\suspensebeat{6}{46}{- The witch tells Gwen she will be ugly half the time and beautiful half, then asks whether he wants her beautiful by day for others to see or beautiful by night for himself.}

\suspensebeat{7}{12}{- Gwen tells the witch to choose for herself.\par\noindent - She declares she will be beautiful all the time because he respected her right to be in charge of her own life.}
\end{suspensemethodbox}
\end{subfigure}\hfill
\begin{subfigure}[t]{0.235\textwidth}
\vspace{0pt}
\begin{suspensemethodbox}{Off-the-shelf}
\suspensebeat{5}{55}{- The court rejoices, but Gawain keeps his promise and weds the old woman, though guests hide their pity behind golden cups.\par\noindent - That night she asks whether he would have her fair by day or by night.}

\suspensebeat{6}{18}{- Gawain, remembering the answer that saved his king, says the choice must be hers, for no gift is fair if it steals sovereignty.\par\noindent - At once she becomes a radiant woman, freed from an enchantment by his respect.}

\suspensebeat{7}{5}{- In Arthur's hall thereafter, the tale was told with cups lowered and hearts lifted: wisdom without honor saves no kingdom.\par\noindent - And all learned that love, like rule, grows noble only when freely given.}
\end{suspensemethodbox}
\end{subfigure}\hfill
\begin{subfigure}[t]{0.235\textwidth}
\vspace{0pt}
\begin{suspensemethodbox}{SWAG}
\suspensebeat{5}{68}{- Arthur gave the answer, and the enemy king, grim with disappointment, freed him before all his court.\par\noindent - But joy soured at Camelot when the crone claimed her bargain, and Sir Gawain rose pale yet steadfast to consent.}

\suspensebeat{6}{82}{- The wedding torches guttered though no wind blew, and lords whispered that Gawain rode to a darker doom than battle.\par\noindent - In the bridal chamber, the crone barred the door, fixed him with bright eyes, and asked whether he would have her fair by night or fair by day.}

\suspensebeat{7}{35}{- Gawain bowed his head. "Lady, the choice is yours; no husband should rule the face or fate his wife must wear."\par\noindent - At once she became young and radiant, saying, "You have given me what all desire: my own will. Therefore I shall be fair both night and day."}
\end{suspensemethodbox}
\end{subfigure}\hfill
\begin{subfigure}[t]{0.235\textwidth}
\vspace{0pt}
\begin{suspensemethodbox}{Ours}
\suspensebeat{5}{46}{- At Camelot, the old woman names her price: marriage to Arthur's fairest knight, Sir Gawain.\par\noindent - Gawain, seeing Arthur's shame, kneels and accepts, saying a promise kept is a kingdom's true crown.}

\suspensebeat{6}{62}{- On their wedding night, the old bride asks Gawain whether he would have her fair by day and foul by night, or foul by day and fair by night.\par\noindent - Gawain, remembering her answer, bows his head and says the choice is hers alone.}

\suspensebeat{7}{12}{- At that, the old woman's shape brightened into a lady both wise and fair, freed from a spell by his trust.\par\noindent - Gawain rejoiced, and all Camelot learned that honor ripens when choice is honored.}
\end{suspensemethodbox}
\end{subfigure}

\caption{Generated story examples for human ground truth,
off-the-shelf GPT-5.5, SWAG, and our method (GPT-5.5 writer),
using ``Choice at Camelot'' from the suspense dataset.
The methods differ in when the witch transforms and in character naming.
Background colors encode saved GPT-5.4-mini suspense ratings
on a 0--100 scale, with higher suspense shown in deeper orange.
}
\label{fig:appendix-qual-s5}
\end{figure*}

\begin{figure*}[!htbp]
\centering

\begin{suspensemethodbox}{Shared Story Opening}
\suspensebeat{3}{42}{- Eagle boasts to the gathered birds about his large, strong body and claims he alone can fly the highest.}
\end{suspensemethodbox}

\begin{subfigure}[t]{0.235\textwidth}
\vspace{0pt}
\begin{suspensemethodbox}{Human (GT)}
\suspensebeat{4}{58}{- Eagle looks around for the challenger and, surprised to find it is a little sparrow, answers in a mocking voice, “Oh..!!!”}

\suspensebeat{5}{61}{- Eagle mocks the sparrow, asking if she will be the one to beat him; the sparrow says nothing.}

\suspensebeat{6}{72}{- Owl signals the race to start; the birds fly upward, but they cannot keep flying high and fast for long, and one by one they drop out.}

\suspensebeat{7}{18}{- Eagle alone kept flying, saw the others quit, and declared himself king because he flew highest.\par\noindent - Unknown to him, the tiny sparrow had ridden under his wing; when he stopped, she darted above his head and claimed she had flown higher.\par\noindent - All the birds agreed with the sparrow.}
\end{suspensemethodbox}
\end{subfigure}\hfill
\begin{subfigure}[t]{0.235\textwidth}
\vspace{0pt}
\begin{suspensemethodbox}{Off-the-shelf}
\suspensebeat{4}{78}{- Eagle beats every wing below him, then tires and cries that none can rise higher.\par\noindent - From his back, Wren springs upward.}

\suspensebeat{5}{52}{- Wren flutters a little higher and chirps, "Then I am king!"\par\noindent - The watching birds gasp below.}

\suspensebeat{6}{74}{- Eagle, shamed and furious, shakes his feathers and dives after Wren, crying that tricks cannot wear a crown.\par\noindent - Wren tumbles toward the trees.}

\suspensebeat{7}{28}{- Myna darts between them and calls the contest ended, for height was reached before anger began.\par\noindent - The birds murmur that a crown won by cunning sits poorly, yet a boastful wing is no better throne.}
\end{suspensemethodbox}
\end{subfigure}\hfill
\begin{subfigure}[t]{0.235\textwidth}
\vspace{0pt}
\begin{suspensemethodbox}{SWAG}
\suspensebeat{4}{78}{- Clouds swallowed Eagle until only a dark speck trembled above the sunlit leaves.\par\noindent - Then his wings faltered, and from his back came the faintest rustle.}

\suspensebeat{5}{84}{- A tiny shadow stirred where no bird should be, and the jungle held its breath.\par\noindent - Eagle glanced upward, startled by a chirp above him.}

\suspensebeat{6}{72}{- Wren darted higher, but Myna cried out from below: the crown was never for height alone.\par\noindent - True kings, she said, must rise without stealing another's wings.}

\suspensebeat{7}{18}{- Wren hovered, ashamed, between the sun and the forest, learning that cleverness without honor leaves even the highest perch empty.\par\noindent - Eagle, humbled, folded his pride like tired wings, and the birds saw that greatness is not how far one rises, but how truthfully one flies.}
\end{suspensemethodbox}
\end{subfigure}\hfill
\begin{subfigure}[t]{0.235\textwidth}
\vspace{0pt}
\begin{suspensemethodbox}{Ours}
\suspensebeat{4}{60}{- Eagle climbed through the bright air, higher than parrot and crane, while the watchers cried that victory was surely his.}

\suspensebeat{5}{74}{- Yet when Eagle’s wings trembled near the clouds, a small rustle stirred beneath his golden feathers.}

\suspensebeat{6}{52}{- Out popped Wren, fresh as dawn, and fluttered one wingbeat higher.\par\noindent - The jungle birds gasped; Eagle, too tired to protest, drifted earthward.}

\suspensebeat{7}{18}{- Myna lifted her beak and declared tiny Wren the winner, for she had flown highest of all.\par\noindent - Owl blinked wisely and said, "The crown fits not the largest head, but the cleverest heart."}
\end{suspensemethodbox}
\end{subfigure}

\caption{Generated story examples for human ground truth,
off-the-shelf GPT-5.5, SWAG, and our method (GPT-5.5 writer),
using ``Eagle and sparrow'' from the suspense dataset.
The off-the-shelf method reveals the hidden bird early;
our method delays the reveal until one beat before the human reference, while SWAG changes the crowning outcome.
Background colors encode saved GPT-5.4-mini suspense ratings
on a 0--100 scale, with higher suspense shown in deeper orange.
}
\label{fig:appendix-qual-s14}
\end{figure*}

\fi

\iflatexml
  \newcommand{\sftappbeat}[2]{{\fontsize{7.6}{8.8}\selectfont\noindent\textbf{Segment #1}\par
   \vspace{0.15mm}\raggedright #2\par}\vspace{0.8mm}}

\par\medskip\begingroup

\begin{humormethodbox}{Shared Story Opening}{N/A}
\sftappbeat{1}{- A spider lives in a museum basement, spinning its web beside paintings stored there for years.}

\segmentdivider

\sftappbeat{2}{- The spider’s web is the most beautiful in the museum, spectacular enough to seem artistic.\par\noindent - The spider carefully tends it with great effort.}

\segmentdivider

\sftappbeat{3}{- The spider prizes his web above everything.\par\noindent - One day, the museum begins reorganizing its paintings, taking them from the basement to put on display.}

\segmentdivider

\sftappbeat{4}{- Other spiders noticed the changes, abandoned their webs, and moved away, but this spider dismissed it as “just a few paintings.”\par\noindent - As more paintings left the basement, he kept reinforcing his web.}
\end{humormethodbox}

\begin{humormethodbox}{Human (GT)}{N/A}
\sftappbeat{5}{- The spider wonders where he could find a better place than his web.\par\noindent - One morning, museum employees take the painting beside it, and he realizes he must leave now or be carried into the exhibition room.}

\segmentdivider

\sftappbeat{6}{- He decisively abandons the magnificent web he spent his life building.\par\noindent - The choice proves wise: if he had stayed, insect killer sprayed on the paintings would have killed him.}

\segmentdivider

\sftappbeat{7}{- After a difficult escape, the spider reaches a secluded garden, finds a quiet corner, spins an even better web, and becomes happier.\par\noindent - The story concludes that leaving a hard-built place can be necessary, and leaving does not mean losing.}
\end{humormethodbox}
\par\medskip

\begin{humormethodbox}{Off-the-shelf}{N/A}
\sftappbeat{5}{- The final canvas departed, leaving the basement dim and utterly silent with only dust. \textbackslash{}\par\noindent - The spider spun frantically, unaware his masterpiece now hung in total emptiness without its inspiration.}

\segmentdivider

\sftappbeat{6}{- The spider pauses, feeling the sudden absence of color that once guided his threads.\par\noindent - He realizes his beautiful web now clings only to shadows and forgotten silence.}

\segmentdivider

\sftappbeat{7}{- The spider gently cuts his own threads, accepting the loss of a web without art to mirror.\par\noindent - He climbs out into the bright hall where paintings now speak to the living world.}
\end{humormethodbox}
\par\medskip

\begin{humormethodbox}{Ours}{N/A}
\sftappbeat{5}{- The final paintings are lifted onto a truck, leaving the spider alone in a hollow, dusty hall.\par\noindent - He tightens every thread, believing his masterpiece now stands as the room's sole decoration.\par\noindent - Silence grows heavy, pressing against the empty walls where art once hung.}

\segmentdivider

\sftappbeat{6}{- The spider waits for viewers to marvel at his silent masterpiece.\par\noindent - Only dust motes dance in the stagnant air, mocking his pride.\par\noindent - A single thread snaps under the weight of his expectation.}

\segmentdivider

\sftappbeat{7}{- The snapping thread triggers a cascade, his web unraveling instantly in the void.\par\noindent - He stands exposed, realizing no audience will ever witness his art.\par\noindent - The spider spins his final thread, weaving a lesson into silence.}
\end{humormethodbox}
\par\medskip

\begin{humormethodbox}{SFT}{N/A}
\sftappbeat{5}{- The spider’s effort is rewarded: his beautiful web draws more visitors, who leave it undisturbed while examining the paintings.\par\noindent - As more people come, he becomes so proud of himself that \textcolor{red!75!black}{his own name becomes forgotten}.}

\segmentdivider

\sftappbeat{6}{- After years of the museum’s reorganization, the spider is still in the basement, alone with no other paintings.\par\noindent - The visitors he used to see are now far from the basement.}

\segmentdivider

\sftappbeat{7}{- The spider kept his web beautiful as a habit, but no longer cared whether anyone saw or noticed it.\par\noindent - After a while, the web lost its attraction and the spider himself disappeared, though \textcolor{red!75!black}{his name and legend stayed in people’s memories}.}
\end{humormethodbox}
\par\medskip

\captionof{figure}{Generated story examples for human ground truth,
off-the-shelf Qwen, our method (Qwen writer), and direct SFT Qwen, using ``The museum spider'' story from the suspense dataset. 
SFT states that the spider's name becomes forgotten in segment 5, but then claims that his name and legend remain in people's memories in segment 7.}
\label{fig:appendix-sft-coherence-spider}
\par\endgroup\medskip

\else
  \newcommand{\sftappbeat}[2]{{\fontsize{7.6}{8.8}\selectfont\noindent\textbf{Segment #1}\par
   \vspace{0.15mm}\raggedright #2\par}\vspace{0.8mm}}

\begin{figure*}[!htbp]
\centering

\begin{humormethodbox}{Shared Story Opening}{N/A}
\sftappbeat{1}{- A spider lives in a museum basement, spinning its web beside paintings stored there for years.}

\segmentdivider

\sftappbeat{2}{- The spider’s web is the most beautiful in the museum, spectacular enough to seem artistic.\par\noindent - The spider carefully tends it with great effort.}

\segmentdivider

\sftappbeat{3}{- The spider prizes his web above everything.\par\noindent - One day, the museum begins reorganizing its paintings, taking them from the basement to put on display.}

\segmentdivider

\sftappbeat{4}{- Other spiders noticed the changes, abandoned their webs, and moved away, but this spider dismissed it as “just a few paintings.”\par\noindent - As more paintings left the basement, he kept reinforcing his web.}
\end{humormethodbox}

\begin{subfigure}[t]{0.235\textwidth}
\vspace{0pt}
\begin{humormethodbox}{Human (GT)}{N/A}
\sftappbeat{5}{- The spider wonders where he could find a better place than his web.\par\noindent - One morning, museum employees take the painting beside it, and he realizes he must leave now or be carried into the exhibition room.}

\segmentdivider

\sftappbeat{6}{- He decisively abandons the magnificent web he spent his life building.\par\noindent - The choice proves wise: if he had stayed, insect killer sprayed on the paintings would have killed him.}

\segmentdivider

\sftappbeat{7}{- After a difficult escape, the spider reaches a secluded garden, finds a quiet corner, spins an even better web, and becomes happier.\par\noindent - The story concludes that leaving a hard-built place can be necessary, and leaving does not mean losing.}
\end{humormethodbox}
\end{subfigure}\hfill
\begin{subfigure}[t]{0.235\textwidth}
\vspace{0pt}
\begin{humormethodbox}{Off-the-shelf}{N/A}
\sftappbeat{5}{- The final canvas departed, leaving the basement dim and utterly silent with only dust. \textbackslash{}\par\noindent - The spider spun frantically, unaware his masterpiece now hung in total emptiness without its inspiration.}

\segmentdivider

\sftappbeat{6}{- The spider pauses, feeling the sudden absence of color that once guided his threads.\par\noindent - He realizes his beautiful web now clings only to shadows and forgotten silence.}

\segmentdivider

\sftappbeat{7}{- The spider gently cuts his own threads, accepting the loss of a web without art to mirror.\par\noindent - He climbs out into the bright hall where paintings now speak to the living world.}
\end{humormethodbox}
\end{subfigure}\hfill
\begin{subfigure}[t]{0.235\textwidth}
\vspace{0pt}
\begin{humormethodbox}{Ours}{N/A}
\sftappbeat{5}{- The final paintings are lifted onto a truck, leaving the spider alone in a hollow, dusty hall.\par\noindent - He tightens every thread, believing his masterpiece now stands as the room's sole decoration.\par\noindent - Silence grows heavy, pressing against the empty walls where art once hung.}

\segmentdivider

\sftappbeat{6}{- The spider waits for viewers to marvel at his silent masterpiece.\par\noindent - Only dust motes dance in the stagnant air, mocking his pride.\par\noindent - A single thread snaps under the weight of his expectation.}

\segmentdivider

\sftappbeat{7}{- The snapping thread triggers a cascade, his web unraveling instantly in the void.\par\noindent - He stands exposed, realizing no audience will ever witness his art.\par\noindent - The spider spins his final thread, weaving a lesson into silence.}
\end{humormethodbox}
\end{subfigure}\hfill
\begin{subfigure}[t]{0.235\textwidth}
\vspace{0pt}
\begin{humormethodbox}{SFT}{N/A}
\sftappbeat{5}{- The spider’s effort is rewarded: his beautiful web draws more visitors, who leave it undisturbed while examining the paintings.\par\noindent - As more people come, he becomes so proud of himself that \textcolor{red!75!black}{his own name becomes forgotten}.}

\segmentdivider

\sftappbeat{6}{- After years of the museum’s reorganization, the spider is still in the basement, alone with no other paintings.\par\noindent - The visitors he used to see are now far from the basement.}

\segmentdivider

\sftappbeat{7}{- The spider kept his web beautiful as a habit, but no longer cared whether anyone saw or noticed it.\par\noindent - After a while, the web lost its attraction and the spider himself disappeared, though \textcolor{red!75!black}{his name and legend stayed in people’s memories}.}
\end{humormethodbox}
\end{subfigure}

\caption{Generated story examples for human ground truth,
off-the-shelf Qwen, our method (Qwen writer), and direct SFT Qwen, using ``The museum spider'' story from the suspense dataset. 
SFT states that the spider's name becomes forgotten in segment 5, but then claims that his name and legend remain in people's memories in segment 7.}
\label{fig:appendix-sft-coherence-spider}
\end{figure*}

\fi

\iflatexml

\par\medskip\begingroup

\begin{humormethodbox}{Shared Story Opening}{N/A}
\humorsftsegment{1}{There's not a married man in this audience right now that can go out and feed pigeons. and his wife call and she say, babe, where you at, and you go, I'm feeding pigeons.}
\segmentdivider

\humorsftsegment{2}{and she believed that she was feeding some f**king pigeons and it never happened. it's an argument off the back. where you at, babe, I'm out here feeding pigeons. you ain't feeding no damn pigeons.}

\end{humormethodbox}

\begin{humormethodbox}{Off-the-shelf}{4.3/5}
\humorsftsegment{3}{You try to tell her, "Honey, it's spiritual." She's already on the phone with the GPS. She says, "Your location says you're at the bus stop." I'm like, "It's not the bus stop, it's the pigeon sanctuary." She hangs up. You're not feeding pigeons, you're hiding from the truth.}

\segmentdivider

\humorsftsegment{4}{So now I'm just standing there with a bag of stale bread, looking at a bird like it's my therapist. "I understand, Charlie, but she thinks you're at the bus stop." The pigeon looks at me, drops a crumb, and flies away. That bird judged me harder than my wife just did. It was silent, but the message was loud: get your life together.}

\segmentdivider

\humorsftsegment{5}{I drop the bag. What am I doing? I'm thirty-five years old standing next to a sewer grate talking to a pigeon about my marriage problems. That bird has seen worse, it's a rat. I go home, I say, "I was at the sanctuary." She says, "The sanctuary? The only thing you sanctified was the bus stop." I'm done. No more pigeons.}
\end{humormethodbox}
\par\medskip

\begin{humormethodbox}{Ours}{4.3/5}
\humorsftsegment{3}{I'm just trying to get a little bit of bird food in my hand before I have to explain why I'm standing in a public square. It's a simple transaction, a guy and a pigeon, and somehow I become the villain in a story about trust. I just want the crumb, I don't want the drama.}

\segmentdivider

\humorsftsegment{4}{So the wife arrives, scanning the square like she's defusing a bomb. She spots me with a bag of seeds and a very judgmental pigeon. She doesn't yell about lying; she just sighs like I've missed a funeral. That silence is louder than any screaming. I'm the man who lied about birds, but really, I just wanted to feel something that wouldn't ask me to change the oil.}

\segmentdivider

\humorsftsegment{5}{So we walk back in that heavy silence, me clutching the half-eaten bird seed like contraband. She just looks at the pigeon following us, shaking its head, and I realize that's the only relationship in my life that doesn't require a contract. Maybe I should just move to the park and live with the birds. They don't care about the oil.}
\end{humormethodbox}
\par\medskip

\begin{humormethodbox}{SFT}{1.7/5}
\humorsftsegment{3}{where you at, you know you at. you at the house with a couple. you ain't feeding no pigeons. \textcolor{red!75!black}{I was at the park, I was at the park}. I didn't have no park. you at the park, I was at the park.}

\segmentdivider

\humorsftsegment{4}{\textcolor{red!75!black}{I was at the park, I was at the park}. you at the park, I was at the park. it is not the park, you at the park. you at the park with a couple.}

\segmentdivider

\humorsftsegment{5}{\textcolor{red!75!black}{you at the park, you at the park with a couple}, you at the park. it is not the park, you at the park with a couple. you at the park, you at the park with a couple.}
\end{humormethodbox}
\par\medskip
\captionof{figure}{Generated story examples for off-the-shelf Qwen, our method (Qwen writer), and direct SFT Qwen, using ``Feeding pigeons'' story from humor dataset.
SFT repeats variations of being at the park across all three segments without developing the wife's disbelief or the pigeon premise.}
\label{fig:appendix-humor-sft-pigeons}
\par\endgroup\medskip

\par\medskip\begingroup

\begin{humormethodbox}{Shared Story Opening}{N/A}
\humorsftsegment{1}{You see, Gane, a haircut as a man, it's like wearing makeup as a woman. Yes, it makes you ten times more attractive. You see, once I get a trim, my face feels different. yeah, I look different.}
\end{humormethodbox}

\begin{humormethodbox}{Off-the-shelf}{4.3/5}
\humorsftsegment{2}{But the moment I walk out that door, the magic vanishes. It's like I'm back to being the same guy who just paid forty bucks to look like he has his life together for about ten minutes. The barber gives me the "you look fresh" nod, but I know by tomorrow the wind has already ruined it completely.}

\segmentdivider

\humorsftsegment{3}{Then I get home and look in the mirror, and suddenly I'm thinking, "Why didn't I ask for a little more off the sides?" That's the trick with the mirror. It lies. It tells you that forty-dollar cut looks amazing, but the moment you step outside into the real world, gravity takes over and you're just a balding mess again.}

\segmentdivider

\humorsftsegment{4}{So now I'm standing there, staring at the reflection, questioning my life choices. I start imagining different scenarios where I have a full head of hair, just to see if I like the guy in that version of reality. But then I remember, that guy doesn't exist, and neither does the money I just spent on a temporary disguise that the wind is already trying to steal back.}

\segmentdivider

\humorsftsegment{5}{And the worst part is the wind, right? It knows exactly where your bald spot is hiding. You walk outside, confident, feeling sharp, and suddenly a gust just reaches through your new style, finds the weak spot, and announces your age to everyone on the block. It's like nature is just screaming, "Hey, we see you!"}

\segmentdivider

\humorsftsegment{6}{So I just give up now. I leave the house with a hat, but not even a stylish one, just a beanie pulled down so far my eyebrows are doing the work. I'm walking around like a suspicious character who just robbed a bakery, pretending that's the look. At least with a hat, the wind can't steal the hair I don't have, and honestly, that's the only victory left in this battle.}
\end{humormethodbox}
\par\medskip

\begin{humormethodbox}{Ours}{4.0/5}
\humorsftsegment{2}{But the problem is, I walk out of that shop feeling like a Hollywood star, and my wife looks at me like I committed a felony. "Why are you wearing that helmet?" she asks. I say, "It's a style!" She says, "It looks like a bad decision with scissors." I'm just trying to be handsome here, people!}

\segmentdivider

\humorsftsegment{3}{Anyway, I tried to explain to her that the helmet look has a certain edge. She just laughed and said, "The only edge here is the one you're walking on." I swear, if I had known marriage was this much work, I would have stayed single. Well, maybe not single, but at least I wouldn't have picked the guy with the bad haircut.}

\segmentdivider

\humorsftsegment{4}{So I decided to just live with the helmet style until it grows out. It's a bold choice, really. My wife says it's tragic, but I call it "low maintenance." Mostly because I'm too scared to touch the thing with my own hands. I'm basically a human construction site waiting for a miracle.}

\segmentdivider

\humorsftsegment{5}{So now I'm just walking around looking like a helmet malfunctioned on a construction worker. I tried to blowdry it once, and the mirror looked back at me with pure judgment. It's not hair anymore; it's a structural integrity issue. I feel like if I sneeze too hard, the whole thing collapses onto my nose.}

\segmentdivider

\humorsftsegment{6}{Finally, I walked into the barbershop and said, "Take it all off." He asked if I was sure, and I screamed, "Yes! Just shave it down to the scalp!" Now I look like a bald potato, but at least my wife isn't asking where I put her safety helmet. Freedom is just a bald head away, people.}
\end{humormethodbox}
\par\medskip

\begin{humormethodbox}{SFT}{1.3/5}
\humorsftsegment{2}{I feel like you know I'm walking down the street and if I didn't get a haircut I don't know. I feel like I would have to walk a little slower, a little more like I didn't have my confidence and then, as soon as I get a trim, I feel like.}

\segmentdivider

\humorsftsegment{3}{\textcolor{red!75!black}{I'm like, I'm like I'm like I'm like}. I'm walking down the street, I'm like, you know, I'm like, oh, and I'll look in a mirror, in a store, and I'm like, I'm like, you know, I'm like, I look at my face, I'm like.}

\segmentdivider

\humorsftsegment{4}{and then, as soon as I get a trim, I can walk through the streets and not feel like someone's about to hit me in the face with a brick. I'm like. I'm like. I'm like I don't want to get hit with a brick in the face.}

\segmentdivider

\humorsftsegment{5}{\textcolor{red!75!black}{that's why I wear makeup}. I'm like, well, if I don't wear makeup, you're going to hit me in the face with a brick. I'm like, I'm like, okay, I'll put on some makeup.}

\segmentdivider

\humorsftsegment{6}{if that means that you don't want to hit me in the face with a brick, then that's okay. \textcolor{red!75!black}{I'll put on some makeup, You hit me in the face with a brick}. I'll put on some makeup.}
\end{humormethodbox}
\par\medskip

\captionof{figure}{Generated story examples for off-the-shelf Qwen, our method (Qwen writer), and direct SFT Qwen, using ``The haircut'' story from the humor dataset.
SFT stalls in repeated first-person fragments, then leaves the haircut premise for an unintroduced makeup-and-brick exchange.}
\label{fig:appendix-humor-sft-haircut}
\par\endgroup\medskip

\par\medskip\begingroup

\begin{humormethodbox}{Shared Story Opening}{N/A}
\humorsftsegment{1}{you guys meet Craig's list. why are we clapping? why are we clapping? you should be murdered. was it for dating, specifically? or was it like, oh, you had a couch and I came to pick it up and our eyes locked.}
\end{humormethodbox}

\begin{humormethodbox}{Off-the-shelf}{4.3/5}
\humorsftsegment{2}{It was mostly just the desperation that killed people. You'd post a listing, "guy looking for fun," and within minutes some guy named "Sven" replies saying he's a Viking who just needs a warm place to nap. You meet him at the coffee shop, he's not Swedish, he's just really confused by technology.}

\segmentdivider

\humorsftsegment{3}{So he shows up in a leather jacket that's definitely a bedsheet pinned with safety pins. He orders the most expensive item on the menu, stares at it, and whispers, "The beans. They smell like my ancestors' sweat." I'm thinking, this is either a documentary or a hostage situation.}

\segmentdivider

\humorsftsegment{4}{Finally, he realizes he can't pay. He asks if I can hold the bill for "later." I say sure, but only if you leave the coffee shop. He tries to run, trips over his own boot, and lands face-first in the pumpkin spice latte. Now I'm the one in the hostage situation, holding a guy who smells like wet dog and old history.}

\segmentdivider

\humorsftsegment{5}{So I'm wiping pumpkin spice off his leather bedsheet jacket when he finally whispers, "The ancestors... they are thirsty." I just wanted a date, not to hydrate a dead warrior. He's staring at the sugar packet like it's the map to Valhalla, convinced this is a ritual.}

\segmentdivider

\humorsftsegment{6}{So I hand him the sugar and he chokes, claiming it's poison from the gods. I tell him it's granulated sucrose from Nebraska, but he's already backing out the door, screaming for his longship. I go back home to check my listings and realize I'm just happy to be alone with my cat, who actually knows what a coffee bean is.}
\end{humormethodbox}
\par\medskip

\begin{humormethodbox}{Ours}{4.3/5}
\humorsftsegment{2}{The dating thing, though, it felt like a job interview where the only requirement was having a weird hobby. You'd meet someone, and they'd immediately start talking about their collection of vintage bottle caps while you're just thinking about whether they have a pulse. It's terrifying, honestly.}

\segmentdivider

\humorsftsegment{3}{So you're sitting there listening to this person explain the aerodynamic properties of bottle caps, and you're just trying to keep it polite. You realize you're not looking for a partner anymore, you're looking for a therapist who doesn't charge per hour. It gets exhausting trying to find a normal person in a sea of specific weirdness.}

\segmentdivider

\humorsftsegment{4}{Eventually, you just give up on finding that specific weirdness and decide normal is overrated. So you end up talking to someone who seems perfectly average, and that's when the panic sets in. Who is this person, and why do they look like they could be my neighbor? It's too comfortable, too safe, and honestly, a little boring compared to the bottle cap guy.}

\segmentdivider

\humorsftsegment{5}{So I'm sitting there with the neighbor guy, and the silence is so loud I can hear my own thoughts judging him. He asks if I want to check out his garage sale, which is the worst date proposal in history. At least the bottle cap guy had a passion; this guy just has a lot of used lawn chairs and a desperate need for validation.}

\segmentdivider

\humorsftsegment{6}{I tell him I'm not interested in his furniture, but he insists, saying the chairs have "character." So I leave, and now I'm back to being single, surrounded by people with hobbies I don't understand and neighbors who want me to flip their lawn chairs. At this point, I think the real love of my life is just my couch, and it never asks me for a garage sale.}
\end{humormethodbox}
\par\medskip

\begin{humormethodbox}{SFT}{1.3/5}
\humorsftsegment{2}{Oh, you don't have a couch. I'm not going to say anything if you don't have a couch. you don't have a couch, but you still found someone on Craig's list. okay, I'm not going to judge you, but you're just a liar.}

\segmentdivider

\humorsftsegment{3}{That's okay. So how did you meet? \textcolor{red!75!black}{You met Craig's list, but you also went to the same school. So which one is it?} Did you find each other on Craig's list? Did you find each other on Craig's list?}

\segmentdivider

\humorsftsegment{4}{Did you post a Craig's list ad looking for a girl from the same school? Is that what happened? No, you guys found each other. Okay, so, okay, so, okay, so, okay, So you guys met in college?}

\segmentdivider

\humorsftsegment{5}{Did you guys meet in college? You guys met in college. you guys are lying. We don't want you in our show. I hate liars. I don't want liars here. You guys met Craig's List. How long have you been together?}

\segmentdivider

\humorsftsegment{6}{You guys have been together for two years and you didn't realize it. What the f***? What is happening? I want to understand how \textcolor{red!75!black}{I'm in love with two girls here}. What is happening?}
\end{humormethodbox}
\par\medskip

\captionof{figure}{Generated story examples for off-the-shelf Qwen, our method (Qwen writer), and direct SFT Qwen, using ``Meeting on Craigslist'' story from the humor dataset.
SFT keeps asking whether the couple met on Craigslist or at school as if the two facts conflict, then introduces two women without setup.}
\label{fig:appendix-humor-sft-craigslist}
\par\endgroup\medskip

\par\medskip\begingroup

\begin{humormethodbox}{Shared Story Opening}{N/A}
\humorsftsegment{1}{My cousin said: man, we're having a car party, Bernie, come on, man. I said man, I don't know. man. he said: please come, Bernie, I'm telling everybody you're coming. I said okay, man. so I came Boston.}
\end{humormethodbox}

\begin{humormethodbox}{Off-the-shelf}{4.7/5}
\humorsftsegment{2}{So I'm driving into this house in the suburbs, right? And it's packed. People are drinking, music's blasting, but there's no car. There isn't even a garage. Just a sedan parked on the lawn looking terrified. I'm like, "Where's the car?" He's like, "Oh, it's metaphorical. It's a vibe."}

\segmentdivider

\humorsftsegment{3}{So I'm standing there sweating, and a guy hands me a plastic bag with a single hubcap in it. He says, "This is your ride. Feel the torque." I tried to spin it, but it just fell on my foot. Then the DJ starts playing the sound of a V8 engine, which was just a guy in a basement yelling, "Vroom, vroom."}

\segmentdivider

\humorsftsegment{4}{Anyway, I'm limping around with this hubcap on my shoe, trying to blend in. Someone yells, "Everyone grab a steering wheel!" so I find a plastic one someone brought from their kid's toy store. Now I'm just driving circles in the kitchen while everyone else is doing laps around the coffee table, screaming about horsepower that doesn't exist.}

\segmentdivider

\humorsftsegment{5}{So I'm doing laps around the coffee table, screaming about horsepower, when the host yells, "Time for the pit stop!" I think they mean food, but no, they all just grab spoons and start polishing their shoes like they're cleaning off tire tracks. I tried to use my hubcap as a mirror, but I just saw a very worried man staring back at me, sweating through his shirt.}

\segmentdivider

\humorsftsegment{6}{So I'm just staring at my reflection in this hubcap, sweating through my shirt, when the host screams, "We're out of gas!" Everyone panics and starts running in circles holding invisible fuel cans. I finally snap, drop the steering wheel, and yell, "I'm leaving!" I walked out that door, got to my real car, and drove home in silence. That was the best car party I've ever been to.}
\end{humormethodbox}
\par\medskip

\begin{humormethodbox}{Ours}{4.7/5}
\humorsftsegment{2}{So I show up in Boston, and the "party" is just three guys staring at a rusted sedan in a parking lot. No music, no snacks, just the car. He goes, "Bernie, look, it's a vehicle." I said, "It's a paperweight! This is the worst party since the time my toaster tried to join the union!"}

\segmentdivider

\humorsftsegment{3}{So I stood there looking at this rusted heap while my cousin tried to explain the features. He pointed at the flat tire and said, "See? That's low profile." I just wanted to go back to my apartment and pretend this never happened. My whole day is ruined because of a car that looks like it's melting.}

\segmentdivider

\humorsftsegment{4}{Then he pops the hood and this cloud of smoke hits me like a slap in the face. "Oh, look," he says, wiping grease on his jeans, "that's the engine breathing." I said, "It's not breathing, it's choking! That thing sounds like it's fighting a bear in there. I think it just threw a rod at me."}

\segmentdivider

\humorsftsegment{5}{At this point, the smoke cleared enough for me to see the actual damage, or what was left of it. My cousin just shrugged, saying, "Well, it's character building." I told him the only thing this is building is a fine with the fire department. We were standing there in a parking lot, watching a car literally give up on life.}

\segmentdivider

\humorsftsegment{6}{So finally, I told him, "You know what? This car has too much character. I'm done." He looked at me like I was crazy. I said, "Next time you want a party, buy a cake, not a death trap!" Then I walked away, leaving them with a vehicle that officially gave up on existence while I finally got some fresh air.}
\end{humormethodbox}
\par\medskip

\begin{humormethodbox}{SFT}{1.7/5}
\humorsftsegment{2}{I had to drive 15 minutes. I drove 15 minutes and I get to this party. \textcolor{red!75!black}{I'm in this car, there's like 200 people in there}. I'm like: this is what a car party looks like? It's like we're on a bus.}

\segmentdivider

\humorsftsegment{3}{I said, well, this is what we're going to do. We're going to do this: I'm going to be at the front of the party. You'll have to get up, but this is how it's going to work: I'll take my turn at the front of the party and you'll take your turn at the front.}

\segmentdivider

\humorsftsegment{4}{and if we keep doing that we won't have to get up. but my cousin's like what you talking about? we're not supposed to do that. but I'm like, well, why don't we do this?}

\segmentdivider

\humorsftsegment{5}{and my cousin's like no, we're supposed to be on the dance floor the whole time. so I said all right, so this is the plan: \textcolor{red!75!black}{if you want the dance floor, you got to go get the dance floor}.}

\segmentdivider

\humorsftsegment{6}{so I got up in my cousin's face and I was like if you really want the dance floor, you got to get up in my face and go for the dance floor. and then he got up in my face and then I said: oh, that's what I thought.}
\end{humormethodbox}
\par\medskip

\captionof{figure}{Generated story examples for off-the-shelf Qwen, our method (Qwen writer), and direct SFT Qwen, using ``The car party'' story from the humor dataset.
SFT shifts from a car holding 200 people to instructions about taking turns at the front and then a dance floor, without establishing a stable place for the scene.}
\label{fig:appendix-humor-sft-car-party}
\par\endgroup\medskip

\else

\begin{figure*}[!htbp]
\centering

\begin{humormethodbox}{Shared Story Opening}{N/A}
\humorsftsegment{1}{There's not a married man in this audience right now that can go out and feed pigeons. and his wife call and she say, babe, where you at, and you go, I'm feeding pigeons.}
\segmentdivider

\humorsftsegment{2}{and she believed that she was feeding some f**king pigeons and it never happened. it's an argument off the back. where you at, babe, I'm out here feeding pigeons. you ain't feeding no damn pigeons.}

\end{humormethodbox}

\begin{subfigure}[t]{0.32\textwidth}
\vspace{0pt}
\begin{humormethodbox}{Off-the-shelf}{4.3/5}
\humorsftsegment{3}{You try to tell her, "Honey, it's spiritual." She's already on the phone with the GPS. She says, "Your location says you're at the bus stop." I'm like, "It's not the bus stop, it's the pigeon sanctuary." She hangs up. You're not feeding pigeons, you're hiding from the truth.}

\segmentdivider

\humorsftsegment{4}{So now I'm just standing there with a bag of stale bread, looking at a bird like it's my therapist. "I understand, Charlie, but she thinks you're at the bus stop." The pigeon looks at me, drops a crumb, and flies away. That bird judged me harder than my wife just did. It was silent, but the message was loud: get your life together.}

\segmentdivider

\humorsftsegment{5}{I drop the bag. What am I doing? I'm thirty-five years old standing next to a sewer grate talking to a pigeon about my marriage problems. That bird has seen worse, it's a rat. I go home, I say, "I was at the sanctuary." She says, "The sanctuary? The only thing you sanctified was the bus stop." I'm done. No more pigeons.}
\end{humormethodbox}
\end{subfigure}\hfill
\begin{subfigure}[t]{0.32\textwidth}
\vspace{0pt}
\begin{humormethodbox}{Ours}{4.3/5}
\humorsftsegment{3}{I'm just trying to get a little bit of bird food in my hand before I have to explain why I'm standing in a public square. It's a simple transaction, a guy and a pigeon, and somehow I become the villain in a story about trust. I just want the crumb, I don't want the drama.}

\segmentdivider

\humorsftsegment{4}{So the wife arrives, scanning the square like she's defusing a bomb. She spots me with a bag of seeds and a very judgmental pigeon. She doesn't yell about lying; she just sighs like I've missed a funeral. That silence is louder than any screaming. I'm the man who lied about birds, but really, I just wanted to feel something that wouldn't ask me to change the oil.}

\segmentdivider

\humorsftsegment{5}{So we walk back in that heavy silence, me clutching the half-eaten bird seed like contraband. She just looks at the pigeon following us, shaking its head, and I realize that's the only relationship in my life that doesn't require a contract. Maybe I should just move to the park and live with the birds. They don't care about the oil.}
\end{humormethodbox}
\end{subfigure}\hfill
\begin{subfigure}[t]{0.32\textwidth}
\vspace{0pt}
\begin{humormethodbox}{SFT}{1.7/5}
\humorsftsegment{3}{where you at, you know you at. you at the house with a couple. you ain't feeding no pigeons. \textcolor{red!75!black}{I was at the park, I was at the park}. I didn't have no park. you at the park, I was at the park.}

\segmentdivider

\humorsftsegment{4}{\textcolor{red!75!black}{I was at the park, I was at the park}. you at the park, I was at the park. it is not the park, you at the park. you at the park with a couple.}

\segmentdivider

\humorsftsegment{5}{\textcolor{red!75!black}{you at the park, you at the park with a couple}, you at the park. it is not the park, you at the park with a couple. you at the park, you at the park with a couple.}
\end{humormethodbox}
\end{subfigure}
\caption{Generated story examples for off-the-shelf Qwen, our method (Qwen writer), and direct SFT Qwen, using ``Feeding pigeons'' story from humor dataset.
SFT repeats variations of being at the park across all three segments without developing the wife's disbelief or the pigeon premise.}
\label{fig:appendix-humor-sft-pigeons}
\end{figure*}

\begin{CompactStoryFigure}
\centering

\begin{humormethodbox}{Shared Story Opening}{N/A}
\humorsftsegment{1}{You see, Gane, a haircut as a man, it's like wearing makeup as a woman. Yes, it makes you ten times more attractive. You see, once I get a trim, my face feels different. yeah, I look different.}
\end{humormethodbox}

\begin{subfigure}[t]{0.32\textwidth}
\vspace{0pt}
\begin{humormethodbox}{Off-the-shelf}{4.3/5}
\humorsftsegment{2}{But the moment I walk out that door, the magic vanishes. It's like I'm back to being the same guy who just paid forty bucks to look like he has his life together for about ten minutes. The barber gives me the "you look fresh" nod, but I know by tomorrow the wind has already ruined it completely.}

\segmentdivider

\humorsftsegment{3}{Then I get home and look in the mirror, and suddenly I'm thinking, "Why didn't I ask for a little more off the sides?" That's the trick with the mirror. It lies. It tells you that forty-dollar cut looks amazing, but the moment you step outside into the real world, gravity takes over and you're just a balding mess again.}

\segmentdivider

\humorsftsegment{4}{So now I'm standing there, staring at the reflection, questioning my life choices. I start imagining different scenarios where I have a full head of hair, just to see if I like the guy in that version of reality. But then I remember, that guy doesn't exist, and neither does the money I just spent on a temporary disguise that the wind is already trying to steal back.}

\segmentdivider

\humorsftsegment{5}{And the worst part is the wind, right? It knows exactly where your bald spot is hiding. You walk outside, confident, feeling sharp, and suddenly a gust just reaches through your new style, finds the weak spot, and announces your age to everyone on the block. It's like nature is just screaming, "Hey, we see you!"}

\segmentdivider

\humorsftsegment{6}{So I just give up now. I leave the house with a hat, but not even a stylish one, just a beanie pulled down so far my eyebrows are doing the work. I'm walking around like a suspicious character who just robbed a bakery, pretending that's the look. At least with a hat, the wind can't steal the hair I don't have, and honestly, that's the only victory left in this battle.}
\end{humormethodbox}
\end{subfigure}\hfill
\begin{subfigure}[t]{0.32\textwidth}
\vspace{0pt}
\begin{humormethodbox}{Ours}{4.0/5}
\humorsftsegment{2}{But the problem is, I walk out of that shop feeling like a Hollywood star, and my wife looks at me like I committed a felony. "Why are you wearing that helmet?" she asks. I say, "It's a style!" She says, "It looks like a bad decision with scissors." I'm just trying to be handsome here, people!}

\segmentdivider

\humorsftsegment{3}{Anyway, I tried to explain to her that the helmet look has a certain edge. She just laughed and said, "The only edge here is the one you're walking on." I swear, if I had known marriage was this much work, I would have stayed single. Well, maybe not single, but at least I wouldn't have picked the guy with the bad haircut.}

\segmentdivider

\humorsftsegment{4}{So I decided to just live with the helmet style until it grows out. It's a bold choice, really. My wife says it's tragic, but I call it "low maintenance." Mostly because I'm too scared to touch the thing with my own hands. I'm basically a human construction site waiting for a miracle.}

\segmentdivider

\humorsftsegment{5}{So now I'm just walking around looking like a helmet malfunctioned on a construction worker. I tried to blowdry it once, and the mirror looked back at me with pure judgment. It's not hair anymore; it's a structural integrity issue. I feel like if I sneeze too hard, the whole thing collapses onto my nose.}

\segmentdivider

\humorsftsegment{6}{Finally, I walked into the barbershop and said, "Take it all off." He asked if I was sure, and I screamed, "Yes! Just shave it down to the scalp!" Now I look like a bald potato, but at least my wife isn't asking where I put her safety helmet. Freedom is just a bald head away, people.}
\end{humormethodbox}
\end{subfigure}\hfill
\begin{subfigure}[t]{0.32\textwidth}
\vspace{0pt}
\begin{humormethodbox}{SFT}{1.3/5}
\humorsftsegment{2}{I feel like you know I'm walking down the street and if I didn't get a haircut I don't know. I feel like I would have to walk a little slower, a little more like I didn't have my confidence and then, as soon as I get a trim, I feel like.}

\segmentdivider

\humorsftsegment{3}{\textcolor{red!75!black}{I'm like, I'm like I'm like I'm like}. I'm walking down the street, I'm like, you know, I'm like, oh, and I'll look in a mirror, in a store, and I'm like, I'm like, you know, I'm like, I look at my face, I'm like.}

\segmentdivider

\humorsftsegment{4}{and then, as soon as I get a trim, I can walk through the streets and not feel like someone's about to hit me in the face with a brick. I'm like. I'm like. I'm like I don't want to get hit with a brick in the face.}

\segmentdivider

\humorsftsegment{5}{\textcolor{red!75!black}{that's why I wear makeup}. I'm like, well, if I don't wear makeup, you're going to hit me in the face with a brick. I'm like, I'm like, okay, I'll put on some makeup.}

\segmentdivider

\humorsftsegment{6}{if that means that you don't want to hit me in the face with a brick, then that's okay. \textcolor{red!75!black}{I'll put on some makeup, You hit me in the face with a brick}. I'll put on some makeup.}
\end{humormethodbox}
\end{subfigure}

\caption{Generated story examples for off-the-shelf Qwen, our method (Qwen writer), and direct SFT Qwen, using ``The haircut'' story from the humor dataset.
SFT stalls in repeated first-person fragments, then leaves the haircut premise for an unintroduced makeup-and-brick exchange.}
\label{fig:appendix-humor-sft-haircut}
\end{CompactStoryFigure}

\begin{CompactStoryFigure}
\centering

\begin{humormethodbox}{Shared Story Opening}{N/A}
\humorsftsegment{1}{you guys meet Craig's list. why are we clapping? why are we clapping? you should be murdered. was it for dating, specifically? or was it like, oh, you had a couch and I came to pick it up and our eyes locked.}
\end{humormethodbox}

\begin{subfigure}[t]{0.32\textwidth}
\vspace{0pt}
\begin{humormethodbox}{Off-the-shelf}{4.3/5}
\humorsftsegment{2}{It was mostly just the desperation that killed people. You'd post a listing, "guy looking for fun," and within minutes some guy named "Sven" replies saying he's a Viking who just needs a warm place to nap. You meet him at the coffee shop, he's not Swedish, he's just really confused by technology.}

\segmentdivider

\humorsftsegment{3}{So he shows up in a leather jacket that's definitely a bedsheet pinned with safety pins. He orders the most expensive item on the menu, stares at it, and whispers, "The beans. They smell like my ancestors' sweat." I'm thinking, this is either a documentary or a hostage situation.}

\segmentdivider

\humorsftsegment{4}{Finally, he realizes he can't pay. He asks if I can hold the bill for "later." I say sure, but only if you leave the coffee shop. He tries to run, trips over his own boot, and lands face-first in the pumpkin spice latte. Now I'm the one in the hostage situation, holding a guy who smells like wet dog and old history.}

\segmentdivider

\humorsftsegment{5}{So I'm wiping pumpkin spice off his leather bedsheet jacket when he finally whispers, "The ancestors... they are thirsty." I just wanted a date, not to hydrate a dead warrior. He's staring at the sugar packet like it's the map to Valhalla, convinced this is a ritual.}

\segmentdivider

\humorsftsegment{6}{So I hand him the sugar and he chokes, claiming it's poison from the gods. I tell him it's granulated sucrose from Nebraska, but he's already backing out the door, screaming for his longship. I go back home to check my listings and realize I'm just happy to be alone with my cat, who actually knows what a coffee bean is.}
\end{humormethodbox}
\end{subfigure}\hfill
\begin{subfigure}[t]{0.32\textwidth}
\vspace{0pt}
\begin{humormethodbox}{Ours}{4.3/5}
\humorsftsegment{2}{The dating thing, though, it felt like a job interview where the only requirement was having a weird hobby. You'd meet someone, and they'd immediately start talking about their collection of vintage bottle caps while you're just thinking about whether they have a pulse. It's terrifying, honestly.}

\segmentdivider

\humorsftsegment{3}{So you're sitting there listening to this person explain the aerodynamic properties of bottle caps, and you're just trying to keep it polite. You realize you're not looking for a partner anymore, you're looking for a therapist who doesn't charge per hour. It gets exhausting trying to find a normal person in a sea of specific weirdness.}

\segmentdivider

\humorsftsegment{4}{Eventually, you just give up on finding that specific weirdness and decide normal is overrated. So you end up talking to someone who seems perfectly average, and that's when the panic sets in. Who is this person, and why do they look like they could be my neighbor? It's too comfortable, too safe, and honestly, a little boring compared to the bottle cap guy.}

\segmentdivider

\humorsftsegment{5}{So I'm sitting there with the neighbor guy, and the silence is so loud I can hear my own thoughts judging him. He asks if I want to check out his garage sale, which is the worst date proposal in history. At least the bottle cap guy had a passion; this guy just has a lot of used lawn chairs and a desperate need for validation.}

\segmentdivider

\humorsftsegment{6}{I tell him I'm not interested in his furniture, but he insists, saying the chairs have "character." So I leave, and now I'm back to being single, surrounded by people with hobbies I don't understand and neighbors who want me to flip their lawn chairs. At this point, I think the real love of my life is just my couch, and it never asks me for a garage sale.}
\end{humormethodbox}
\end{subfigure}\hfill
\begin{subfigure}[t]{0.32\textwidth}
\vspace{0pt}
\begin{humormethodbox}{SFT}{1.3/5}
\humorsftsegment{2}{Oh, you don't have a couch. I'm not going to say anything if you don't have a couch. you don't have a couch, but you still found someone on Craig's list. okay, I'm not going to judge you, but you're just a liar.}

\segmentdivider

\humorsftsegment{3}{That's okay. So how did you meet? \textcolor{red!75!black}{You met Craig's list, but you also went to the same school. So which one is it?} Did you find each other on Craig's list? Did you find each other on Craig's list?}

\segmentdivider

\humorsftsegment{4}{Did you post a Craig's list ad looking for a girl from the same school? Is that what happened? No, you guys found each other. Okay, so, okay, so, okay, so, okay, So you guys met in college?}

\segmentdivider

\humorsftsegment{5}{Did you guys meet in college? You guys met in college. you guys are lying. We don't want you in our show. I hate liars. I don't want liars here. You guys met Craig's List. How long have you been together?}

\segmentdivider

\humorsftsegment{6}{You guys have been together for two years and you didn't realize it. What the f***? What is happening? I want to understand how \textcolor{red!75!black}{I'm in love with two girls here}. What is happening?}
\end{humormethodbox}
\end{subfigure}

\caption{Generated story examples for off-the-shelf Qwen, our method (Qwen writer), and direct SFT Qwen, using ``Meeting on Craigslist'' story from the humor dataset.
SFT keeps asking whether the couple met on Craigslist or at school as if the two facts conflict, then introduces two women without setup.}
\label{fig:appendix-humor-sft-craigslist}
\end{CompactStoryFigure}

\begin{CompactStoryFigure}
\centering

\begin{humormethodbox}{Shared Story Opening}{N/A}
\humorsftsegment{1}{My cousin said: man, we're having a car party, Bernie, come on, man. I said man, I don't know. man. he said: please come, Bernie, I'm telling everybody you're coming. I said okay, man. so I came Boston.}
\end{humormethodbox}

\begin{subfigure}[t]{0.32\textwidth}
\vspace{0pt}
\begin{humormethodbox}{Off-the-shelf}{4.7/5}
\humorsftsegment{2}{So I'm driving into this house in the suburbs, right? And it's packed. People are drinking, music's blasting, but there's no car. There isn't even a garage. Just a sedan parked on the lawn looking terrified. I'm like, "Where's the car?" He's like, "Oh, it's metaphorical. It's a vibe."}

\segmentdivider

\humorsftsegment{3}{So I'm standing there sweating, and a guy hands me a plastic bag with a single hubcap in it. He says, "This is your ride. Feel the torque." I tried to spin it, but it just fell on my foot. Then the DJ starts playing the sound of a V8 engine, which was just a guy in a basement yelling, "Vroom, vroom."}

\segmentdivider

\humorsftsegment{4}{Anyway, I'm limping around with this hubcap on my shoe, trying to blend in. Someone yells, "Everyone grab a steering wheel!" so I find a plastic one someone brought from their kid's toy store. Now I'm just driving circles in the kitchen while everyone else is doing laps around the coffee table, screaming about horsepower that doesn't exist.}

\segmentdivider

\humorsftsegment{5}{So I'm doing laps around the coffee table, screaming about horsepower, when the host yells, "Time for the pit stop!" I think they mean food, but no, they all just grab spoons and start polishing their shoes like they're cleaning off tire tracks. I tried to use my hubcap as a mirror, but I just saw a very worried man staring back at me, sweating through his shirt.}

\segmentdivider

\humorsftsegment{6}{So I'm just staring at my reflection in this hubcap, sweating through my shirt, when the host screams, "We're out of gas!" Everyone panics and starts running in circles holding invisible fuel cans. I finally snap, drop the steering wheel, and yell, "I'm leaving!" I walked out that door, got to my real car, and drove home in silence. That was the best car party I've ever been to.}
\end{humormethodbox}
\end{subfigure}\hfill
\begin{subfigure}[t]{0.32\textwidth}
\vspace{0pt}
\begin{humormethodbox}{Ours}{4.7/5}
\humorsftsegment{2}{So I show up in Boston, and the "party" is just three guys staring at a rusted sedan in a parking lot. No music, no snacks, just the car. He goes, "Bernie, look, it's a vehicle." I said, "It's a paperweight! This is the worst party since the time my toaster tried to join the union!"}

\segmentdivider

\humorsftsegment{3}{So I stood there looking at this rusted heap while my cousin tried to explain the features. He pointed at the flat tire and said, "See? That's low profile." I just wanted to go back to my apartment and pretend this never happened. My whole day is ruined because of a car that looks like it's melting.}

\segmentdivider

\humorsftsegment{4}{Then he pops the hood and this cloud of smoke hits me like a slap in the face. "Oh, look," he says, wiping grease on his jeans, "that's the engine breathing." I said, "It's not breathing, it's choking! That thing sounds like it's fighting a bear in there. I think it just threw a rod at me."}

\segmentdivider

\humorsftsegment{5}{At this point, the smoke cleared enough for me to see the actual damage, or what was left of it. My cousin just shrugged, saying, "Well, it's character building." I told him the only thing this is building is a fine with the fire department. We were standing there in a parking lot, watching a car literally give up on life.}

\segmentdivider

\humorsftsegment{6}{So finally, I told him, "You know what? This car has too much character. I'm done." He looked at me like I was crazy. I said, "Next time you want a party, buy a cake, not a death trap!" Then I walked away, leaving them with a vehicle that officially gave up on existence while I finally got some fresh air.}
\end{humormethodbox}
\end{subfigure}\hfill
\begin{subfigure}[t]{0.32\textwidth}
\vspace{0pt}
\begin{humormethodbox}{SFT}{1.7/5}
\humorsftsegment{2}{I had to drive 15 minutes. I drove 15 minutes and I get to this party. \textcolor{red!75!black}{I'm in this car, there's like 200 people in there}. I'm like: this is what a car party looks like? It's like we're on a bus.}

\segmentdivider

\humorsftsegment{3}{I said, well, this is what we're going to do. We're going to do this: I'm going to be at the front of the party. You'll have to get up, but this is how it's going to work: I'll take my turn at the front of the party and you'll take your turn at the front.}

\segmentdivider

\humorsftsegment{4}{and if we keep doing that we won't have to get up. but my cousin's like what you talking about? we're not supposed to do that. but I'm like, well, why don't we do this?}

\segmentdivider

\humorsftsegment{5}{and my cousin's like no, we're supposed to be on the dance floor the whole time. so I said all right, so this is the plan: \textcolor{red!75!black}{if you want the dance floor, you got to go get the dance floor}.}

\segmentdivider

\humorsftsegment{6}{so I got up in my cousin's face and I was like if you really want the dance floor, you got to get up in my face and go for the dance floor. and then he got up in my face and then I said: oh, that's what I thought.}
\end{humormethodbox}
\end{subfigure}

\caption{Generated story examples for off-the-shelf Qwen, our method (Qwen writer), and direct SFT Qwen, using ``The car party'' story from the humor dataset.
SFT shifts from a car holding 200 people to instructions about taking turns at the front and then a dance floor, without establishing a stable place for the scene.}
\label{fig:appendix-humor-sft-car-party}
\end{CompactStoryFigure}

\fi

\end{document}